\documentclass[journal]{IEEEtran}

\usepackage{moreverb,url}
\usepackage{xcolor}
\usepackage{amsmath}
\usepackage{amssymb}
\usepackage{amsthm}
\usepackage{graphicx}
\usepackage{booktabs}
\usepackage{siunitx}
\usepackage{standalone}
\usepackage[ruled,vlined,linesnumbered]{algorithm2e}
\usepackage[bookmarks=true]{hyperref}
\usepackage{mdframed}
\usepackage{threeparttable}
\usepackage{fancyvrb}
\usepackage{soul}
\usepackage{dsfont,mathabx}
\usepackage{floatrow}
\usepackage{adjustbox}
\usepackage{multirow, makecell, tabularx}
\usepackage{cite}
\usepackage[caption=false,font=normalsize,labelfont=sf,textfont=sf]{subfig}

\def\BibTeX{{\rm B\kern-.05em{\sc i\kern-.025em b}\kern-.08em
    T\kern-.1667em\lower.7ex\hbox{E}\kern-.125emX}}
\usepackage{balance}

\theoremstyle{definition}

\newtheorem{problem}{Problem}
\newtheorem{definition}{Definition}
\newtheorem{remark}{Remark}

\theoremstyle{theorem}
\newtheorem{lemma}{Lemma}

\newtheorem{theorem}{Theorem}

\begin{document}


\title{MoRoCo: An Online Topology-Adaptive Framework
  for Multi-Operator Multi-Robot Coordination under Restricted Communication
}

\author{Zhuoli Tian, Yanze Bao, Yuyang Zhang, and Meng Guo$^*$}

\maketitle
\begin{abstract}
  Fleets of autonomous robots are increasingly deployed with multiple human operators in communication-restricted environments for exploration and intervention tasks such as subterranean inspection, reconnaissance, and search-and-rescue.
  In these settings, communication is often limited to short-range ad-hoc links, making it difficult to coordinate exploration while supporting online human-fleet interactions.
  Existing work on multi-robot exploration largely focuses on information gathering itself, but pays limited attention to the fact that operators and robots issue time-critical requests during execution.
  These requests may require different communication structures, ranging from intermittent status delivery to sustained video streaming and teleoperation.
  To address this challenge, this paper presents MoRoCo, an online topology-adaptive framework for multi-operator multi-robot coordination under restricted communication.
  MoRoCo is built on a latency-bounded intermittent communication backbone that guarantees a prescribed delay for information collected by any robot to reach an operator, together with a detach-and-rejoin mechanism that enables online team resizing and topology reconfiguration.
  On top of this backbone, the framework instantiates request-consistent communication subgraphs to realize different modes of operator-robot interaction by jointly assigning robot roles, positions, and communication topology.
  It further supports the online decomposition and composition of these subgraphs using only local communication,
  allowing multiple requests to be serviced during exploration. The framework extends to heterogeneous fleets, multiple teams, and robot failures.
  Extensive human-in-the-loop simulations and hardware experiments demonstrate effective and reliable coordination under restricted communication.
\end{abstract}

\section{Introduction}\label{sec:intro}

Autonomous robots such as unmanned aerial vehicles (UAVs)
and ground vehicles (UGVs) are becoming common companions
of human operators in various scenarios,
especially to explore unknown and hazardous sites
before entry,
or search and rescue after earthquakes,
see~\cite{couceiro2017overview,klaesson2020planning}.
The robots become extended eyes and arms of the operators
by fusing their perception results to the operators
and performing actions as instructed by the operators.
Existing strategies for collaborative
exploration in~\cite{hussein2014multi,colares2016next,
zhou2023racer,patil2023graph}
often assume all-to-all communication
among the robots.
In other words, any local observation by one robot
is immediately available to others,
which is equivalent to one robot with multiple moving sensors.
This is however impractical in aforementioned scenes
where the communication facilities are unavailable or severely degraded.
In such cases, the robot-robot and robot-operator communication
is often limited to ad-hoc networks in close proximity.
This imposes great challenges for the operators
to coordinate, supervise and interact with the fleet,
as communications would not be feasible unless actively planned.

\begin{figure}[t!]
  \centering
  \includegraphics[width=1.0\linewidth]{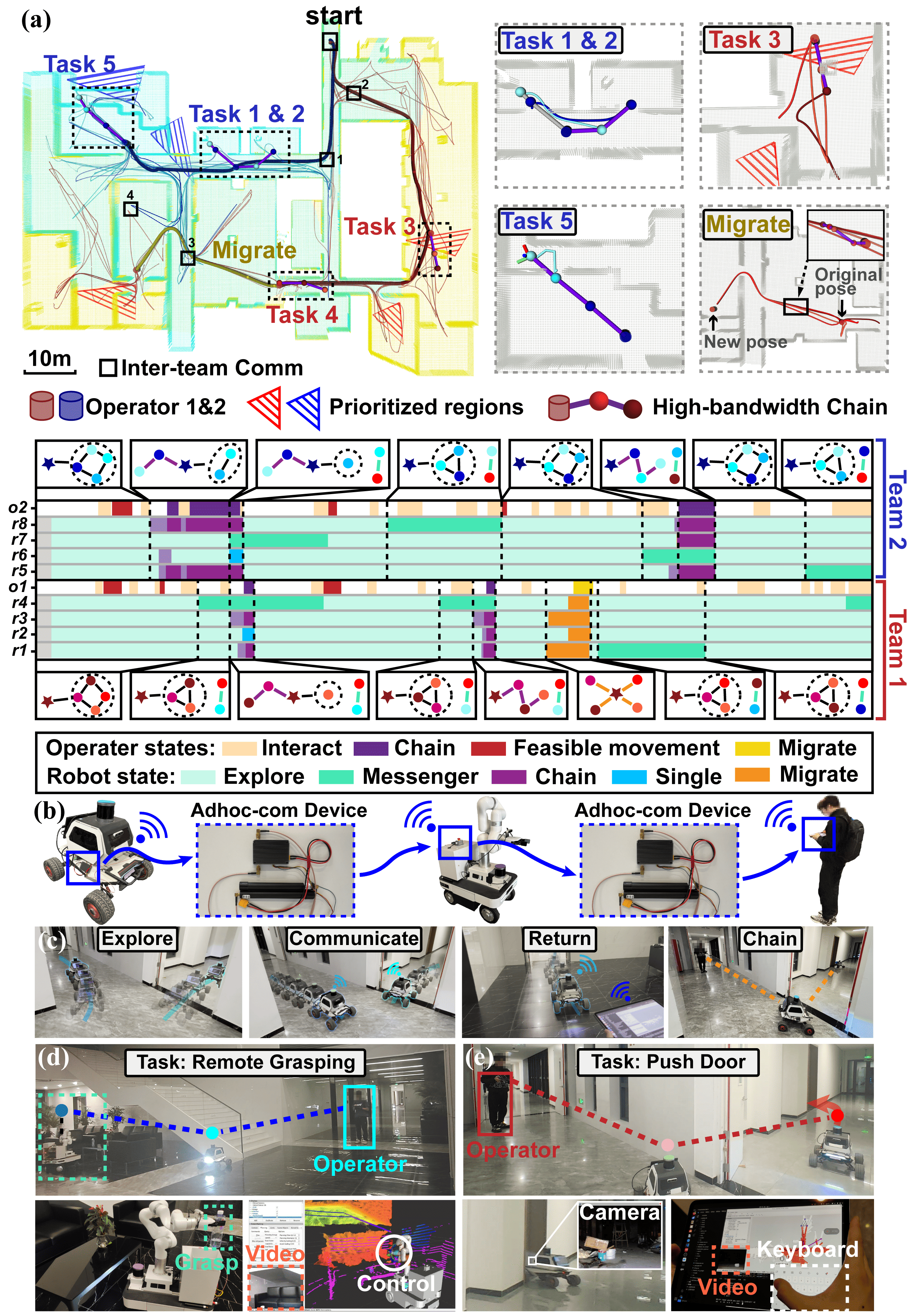}
  \vspace{-4.5mm}
  \caption{Illustration of the considered scenario, where multiple operators and robots
    are deployed in an unknown and communication-restricted environment for exploration and task execution.
    \textbf{(a)} Simulation of 2 operators and 8 robots collaboratively exploring a large-scale environment,
    while accomplishing 5 tasks via splitting and merging embedded communication graphs;
    \textbf{(b)} Hardware experiments of 2 operators and 4 robots,
    with ad-hoc network devices for local communication in close proximity;
    \textbf{(c)} Intermittent communication protocol
    that ensures bounded-latency information flow between operators and robots;
    \textbf{(d)-(e)}  Online request fulfillment via changing the communication topology
    for video streaming and teleoperation.}
  \label{fig:scenario}
\end{figure}

\begin{figure*}[t!]
  \centering
  \includegraphics[width=0.95\linewidth]{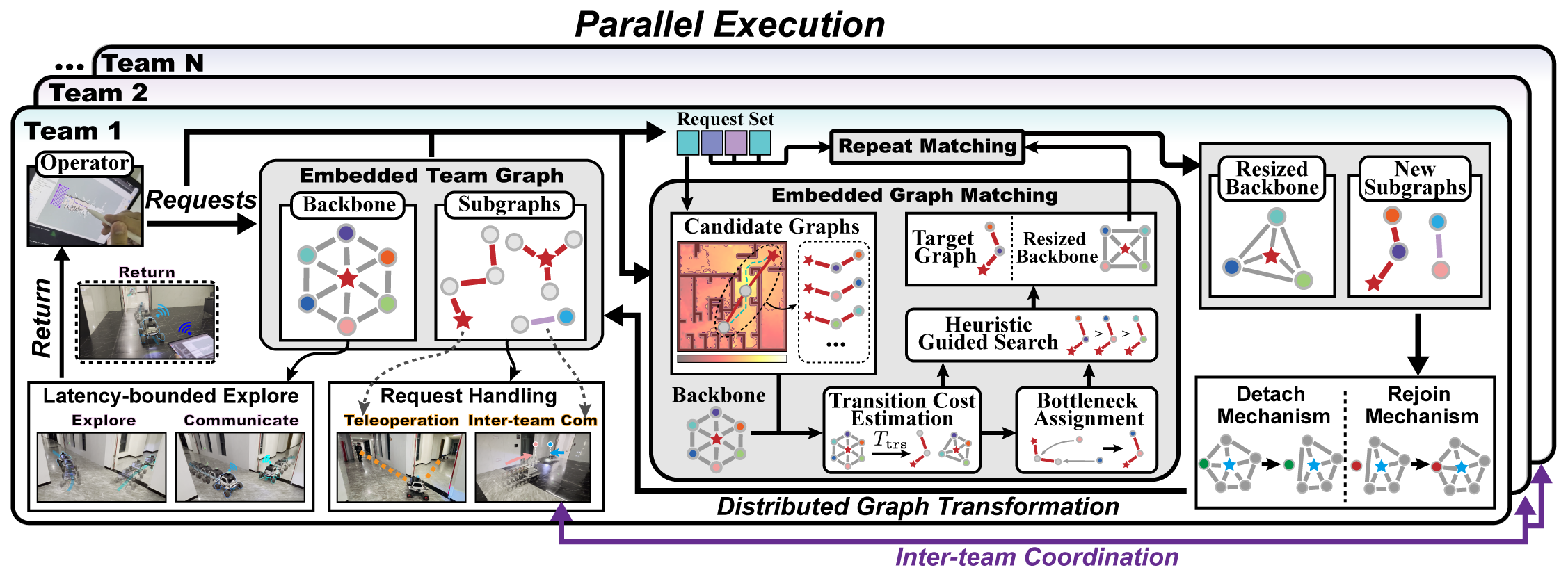}
  \vspace{-2mm}
  \caption{Overall framework of MOROCO for multi-team human--robot coordination
    under restricted communication.
    Multiple teams of operator and robots simultaneously explore the environment and
    fulfill several online requests, following the embedded communication graphs
    under the intermittent communication protocol (\textbf{Left});
    The embedded graph is optimized and matched against different targeted graphs
    given the current graph and the set of active requests (\textbf{Middle});
    The resulting subgraphs are realized by the proposed detach-and-rejoin
    mechanism via only local coordination (\textbf{Right}).}
  \label{fig:method}
\end{figure*}

Furthermore, the role of the human operator and
the online interaction with the fleet during exploration
are less studied and mostly replaced by a static base station for visualization.
However, under restricted communication, the operator might be \emph{completely
unaware} of the progress and results of the ongoing exploration
if none of the robots actively return to the operator to relay such information.
Consequently, the operator can not supervise the exploration task in real-time,
e.g., checking the system status such as battery and liveness;
and dynamically adjusting the priorities of the areas to be explored.
In addition, as shown in Fig.~\ref{fig:scenario},
the operator often needs {direct} access to
the video stream and control of a particular robot, e.g.,
to inspect certain regions close-up for important semantic features
that are otherwise overlooked during exploration;
or assist a robot through tough terrain via manual control.
In another direction, the robots often need confirmation from the operator
regarding actions or detected features, or immediate assistance.
For these interactions, a \emph{reliable and consistent information flow}
is often required
from the operator to the robot even during motion,
which is particularly challenging in unknown environments
without global communication.

To address these challenges, this work introduces \textbf{MoRoCo}, an
online topology-adaptive framework for coordination, supervision, and
interaction in multi-operator multi-robot systems operating in unknown
environments under restricted communication. Seven types of bilateral
and multi-modal interactions that arise during exploration are
explicitly modeled. As outlined in Fig.~\ref{fig:method}, MoRoCo is
built on a latency-bounded intermittent communication backbone that
mediates all request delivery using only local communication. This
backbone ensures that information gathered by any robot reaches an
operator within a prescribed latency bound by coordinating successive
communication events in time, location, and message content. A
detach-and-rejoin procedure enables dynamic resizing of the fleet while
preserving this communication objective. Each request type is further
associated with a desired communication structure, such as a jointed
wheel, line, or star. Building on the backbone, MoRoCo instantiates
request-consistent embedded graphs that specify robot roles, positions,
and communication topology for different interaction modes. To realize
these graphs online, the framework employs a graph-matching-based
request-instantiation procedure that maps the current embedded graph to
an executable target graph. In the presence of multiple simultaneous
requests, an online feasible coordination policy incrementally
decomposes the fleet into subgraphs that service individual requests,
and later recomposes these subgraphs into larger exploration graphs
upon completion. In this way, MoRoCo unifies collaborative
exploration, online request handling, and communication-topology
adaptation under intermittent connectivity. Moreover, the framework
generalizes to heterogeneous fleets, and is demonstrated to handle
exploration uncertainty and robot failures. Extensive human-in-the-loop
simulations are conducted with more than $10$ robots and $3$ operators
in typical office environments and subterranean caves of varying
scales. In addition, hardware experiments are performed with a fleet
of $4$ ground vehicles and $2$ operators in indoor environments.

The main contributions are threefold:
(I) a structured formulation for multi-operator multi-robot systems
operating in unknown environments under restricted communication,
which captures seven explicit bilateral and multi-modal request types
arising between robots and operators during exploration;
(II) the proposed {MoRoCo} framework, which combines
latency-bounded intermittent communication, collaborative exploration,
and online request fulfillment through topology-adaptive coordination;
and
(III) a graph-based realization mechanism based on embedded-graph
instantiation, matching, decomposition, and composition, together with
framework-level guarantees on executable online request servicing and
restoration of the exploration backbone.

\section{Related Work}\label{sec:related}
\subsection{Collaborative Exploration}
The problem of collaborative exploration has a long history in robotics,
as partially summarized in Table~\ref{tab:feature_comparison}.
The frontier-based method from~\cite{yamauchi1997frontier}
introduces an intuitive yet powerful metric for guiding the exploration.
Originally for a single robot, it was extended to multi-robot teams
by assigning frontiers for concurrent exploration:
e.g., greedily to the nearest robot by distance~\cite{yamauchi1999decentralized},
centralized assignment via Voronoi partitioning~\cite{bi2023cure},
distributed and sequential auction~\cite{hussein2014multi},
multi-vehicle routing for optimal sequences~\cite{zhou2023racer},
or based on expected information gain~\cite{burgard2005coordinated,
  baek2025pipe, patil2023graph}.
More recent learning-based methods~\cite{calzolari2024reinforcement, sygkounas2022multi}
leverage learned heuristics to prioritize frontiers or predict the unexplored map directly.
Inter-robot communication is crucial for collaboration, enabling coordinated
planning and avoiding redundant exploration.
With known relative coordinates, map merge can be
efficiently and accurately done~\cite{horner2016map}; otherwise, advanced
methods like feature matching or geometric optimization are needed~\cite{yu2020review}.
Work like ROAM~\cite{asgharivaskasi2025riemannian} proposes a decentralized optimization
scheme for consistency, and SlideSLAM~\cite{liu2024slideslam} enables map fusion via
semantic labelling and matching.
However, these works all assume that robots can exchange information via
wireless communication \emph{perfectly and instantly} at all time.
Thus, all robots always have access to {the same global map}
 and the same set of frontiers.
This is impractical or infeasible without pre-installed communication
infrastructures. As discussed in~\cite{esposito2006maintaining},
due to obstructions in subterranean environments~\cite{ohradzansky2022multi,
  dahlquist2025deployment}
or indoor structures~\cite{al2015enhanced},
inter-robot communication is severely limited both in range and reliability.
Consequently, these approaches are not suitable
for communication-constrained and unknown scenarios.

\subsection{Limited Communication}
To overcome the challenge of limited communication, many recent works
focus on the joint planning of inter-robot communication and autonomous
exploration. A straightforward solution is to establish a communication
network covering the entire environment by pre-deploying communication
devices~\cite{bi2023cure, yu2021smmr}. However, this approach is limited
to known or small-scale environments and cannot be applied to large-scale
unknown scenarios. A more practical approach designs exploration
strategies that explicitly account for communication
constraints~\cite{pei2013connectivity}.
Systematic techniques for mobility-communication networks have been
developed, such as formulating a mixed integer program for the DARPA
subterranean challenge~\cite{klaesson2020planning}. Other methods include
a Riemannian optimization approach for efficient exploration~\cite{asgharivaskasi2025riemannian},
a reinforcement learning strategy with one-hop communication~\cite{wang2025multirobot,tan20254cnetdiffusionapproachmap},
and distributed optimization for semantic mapping and its representation
to limit communication overhead~\cite{asgharivaskasi2025riemannian}.

Moreover, as partially summarized in Table~\ref{tab:feature_comparison},
different intermittent communication strategies address various performance
metrics, such as the ``Zonal and Snearkernet'' relay algorithm~\cite{vaquero2018approach};
the four-state strategy including explore, meet, sacrifice, relay~\cite{cesare2015multi};
distributed relay task assignment~\cite{marchukov2019fast}; centralized
optimization of rendezvous points~\cite{gao2022meeting}; and predicting robot
positions via communication unavailability~\cite{schack2024sound}. Fully
distributed protocols for intermittent communication between robot pairs have
also been proposed~\cite{kantaros2019temporal}. Conversely,
some work imposes fully-connected communication networks at all times
via collaborative motion planning~\cite{zavlanos2011graph}
or by placing front and relay nodes to ensure a lower-bounded
bandwidth~\cite{marcotte2020optimizing,xia2023relink}.
Droppable radios are used as communication relays between robots and
a static base station~\cite{saboia2022achord,riley2023fielded}. Recent work also
proposes efficient map representations to reduce communication overhead~\cite{zhang2022mr},
while others tackle the hard constraint of lower-bounded information
latency for all robots~\cite{tian2024ihero}.
It is important to note that the aforementioned methods typically assume a
static base station and consider \emph{only one communication mode}, either
intermittent communication or full connectivity. Thus, these
approaches are limited when different modes are required for
varying bandwidth, latency and topology. More importantly, they mostly
neglect the crucial role of human operators and the potential online
interactions between the operators and the fleet.

\newcommand{\cmark}{$\checkmark$}
\newcommand{\xmark}{$\times$}

\begin{table}[t]
  \centering
  \caption{Comparison with Related Work.}
  \label{tab:feature_comparison}
  \vspace{-2mm}
  \setlength{\tabcolsep}{3pt}
  \begin{tabular*}{0.98\columnwidth}{@{\extracolsep{\fill}}lcccccc@{}}
    \toprule
    \multirow{2}{*}{\textbf{Method}} &
    {\footnotesize \textbf{Pre-inst.}} &
    {\footnotesize \textbf{Interm.}} &
    {\footnotesize \textbf{Human}} &
    {\footnotesize \textbf{Bound}} &
    {\footnotesize \textbf{Diff.}} &
    {\footnotesize \textbf{Multi}} \\[-0.8ex]
    & {\footnotesize \textbf{infra.}} &
      {\footnotesize \textbf{comm.}} &
      {\footnotesize \textbf{int.}} &
      {\footnotesize \textbf{Lat.}} &
      {\footnotesize \textbf{Top.}} &
      {\footnotesize \textbf{teams}} \\[-0.2ex]
      \midrule
Racer~\cite{zhou2023racer}     & \xmark & \xmark & \xmark & \xmark & \xmark & \xmark \\
CURE~\cite{bi2023cure}         & \cmark & \xmark & \xmark & \xmark & \xmark & \xmark \\
Achord~\cite{saboia2022achord} & \cmark$^{\dagger}$ & \cmark & \xmark & \xmark & \cmark & \xmark \\
Madcat~\cite{riley2023fielded} & \cmark$^{\dagger}$ & \cmark & \cmark & \xmark & \cmark & \xmark \\
    RELINK~\cite{xia2023relink}    & \xmark & \xmark & \xmark & \cmark & \cmark & \xmark \\
iHERO~\cite{tian2024ihero}     & \xmark & \cmark & \cmark & \cmark & \xmark & \xmark \\
    JSSP~\cite{Ani2024iros}        & \xmark & \cmark & \xmark & \xmark & \xmark & \xmark \\
    Corah~\cite{8633953}           & \xmark & \cmark & \xmark & \xmark & \xmark & \xmark \\
    Realm~\cite{11128211}          & \xmark & \xmark & \xmark & \cmark & \cmark & \xmark \\
    \midrule
    \textbf{MoRoCo (Ours)}         & \xmark & \cmark & \cmark & \cmark & \cmark & \cmark \\
    \bottomrule
  \end{tabular*}

  \vspace{2pt}
  \raggedright
  \footnotesize $^{\dagger}$Uses droppable beacons during online execution.
\end{table}

\subsection{Human-Fleet Interaction}
Indeed, the human operator plays \emph{an indispensable role}
in the operation of robotic fleets, even when these systems exhibit
significant autonomous capabilities~\cite{dahiya2023survey}.
In practice, operators must not only monitor the fleet status online
but also intervene when critical decisions or risky actions arise,
while robots may require confirmation or direct assistance in unforeseen situations.  
Most existing work overlooks such bilateral interaction and instead assumes
a static base station for visualization~\cite{klaesson2020planning, gao2022meeting,
saboia2022achord, schack2024sound, dahlquist2025deployment},
resulting in rather uniform fleet behaviors until exploration is completed.  
An interactive exploration strategy was proposed in~\cite{tian2024ihero},
but it is limited to simple exploration tasks and a single communication mode.  
Recent studies investigate richer interaction paradigms,
including multi-modal inputs such as voice, gesture, gaze, and touch~\cite{riley2023fielded},
as well as AR-based interfaces for immersive real-time supervision and drag-and-drop commands~\cite{chen20243d}.  
Joysticks and wristbands have also been used to support multi-robot supervision
while monitoring operator workload~\cite{villani2020humans}.  
However, these approaches typically rely on \emph{consistent and reliable} communication between the operators and the fleet.  
These bilateral and online interactions bring numerous challenges for communication-constrained scenarios,
which are not yet addressed formally in the literature: e.g., how to ensure a timely exchange
between the operators and the fleet including newly explored map and requests;
how to choose different modes of communication such as intermittent or rigid,
to fulfill different requests and switch between modes.
\section{Problem Description}\label{sec:problem}

\subsection{Multi-operator-robot Teams}\label{subsec:ws}
\begin{figure}[t!]
  \centering
  \includegraphics[width=0.9\linewidth]{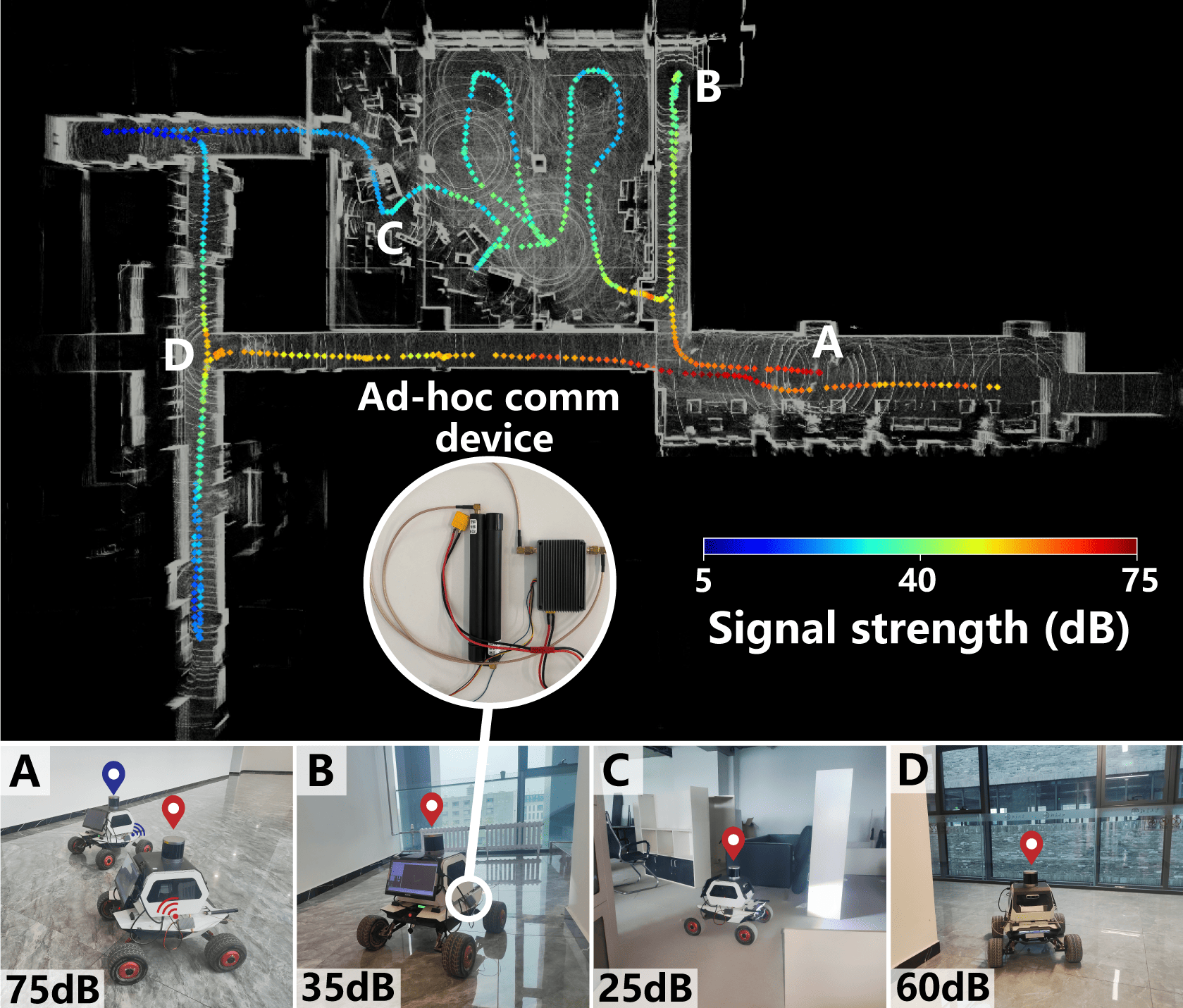}
  \vspace{-2mm}
  \caption{
    The measured signal strength in an office environment
    via the ad-hoc communication network (\textbf{Top})
    at different robot positions (\textbf{Bottom}).
  }
  \label{fig:comm_model}
\end{figure}

Consider a 3D workspace~$\mathcal{A}\subset \mathbb{R}^3$,
of which its {map} including the boundary, freespace and obstacles are all unknown.
To obtain a complete map and features within the workspace,
in total $K>0$ operator-robot
teams~$\mathcal{T}\triangleq \{\mathcal{T}_1,\cdots, \mathcal{T}_K\}$
are deployed into the workspace.
Each team~$\mathcal{T}_k$ consists of one human operator~$\texttt{h}_k$
and a small fleet of autonomous robots~$\mathcal{N}_k$
such as UAVs and UGVs,
i.e., $\mathcal{T}_k\triangleq \{\texttt{h}_k\}\bigcup \mathcal{N}_k$.
In other words, there are~$K$ operators
and~$N$ robots in total, i.e.,~$N\triangleq \sum_{k\in \mathcal{K}}
N_k$ with~$\mathcal{N}\triangleq
\bigcup_{k\in \mathcal{K}} \mathcal{N}_k$
and~$\mathcal{K}\triangleq \{1,\cdots,K\}$.
As shown in Fig.~\ref{fig:scenario} and~\ref{fig:comm_model},
each robot~$i\in \mathcal{N}$ is equipped with various sensors such
as IMU, Lidar and RGBD cameras;
and actuators to move around the workspace,
which allows for simultaneous localization and mapping (SLAM) as follows.

\begin{definition}\label{def:slam}
Each robot is capable of simultaneous
localization and mapping (SLAM) with collision avoidance, i.e.,
\begin{equation}\label{eq:slam}
\big(M^+_i,\, p^+_i,\, \boldsymbol{\tau}_i\big) \triangleq
\texttt{SLAM}\big(p_i,\, p_{\texttt{g}},\, M_i,\,\mathcal{A}\big),
\end{equation}
where~$\texttt{SLAM}(\cdot)$ is a generic SLAM module;
$M_i(t)\subset \mathcal{A}$ is the \emph{local} map of
robot~$i\in \mathcal{N}$ at time~$t\geq 0$;
$p_i(t)\in M_i$ is the pose of robot~$i$;
$p_{\texttt{g}} \in M_i$ is a goal pose within~$M_i$;
$\boldsymbol{\tau}_i \subset M_i$ is the collision-free path from~$p_i(t)$ to~$p_{\texttt{g}}$;
$M^+_i \subset \mathcal{A}$ is the updated map after traversing the path at time~$t'>t$;
and~$p^+_i(t')$ is its updated pose, close to~$p_{\texttt{g}}$
if the navigation is successful. \hfill $\blacksquare$
\end{definition}

On the other hand, the operators~$\texttt{h}_k\in \mathcal{T}_k$
can \emph{not} build a map,
but can localize themselves with pose~$p^k_{\texttt{h}}\in \mathcal{A}$
and move within their local maps~$M^k_{\texttt{h}}$,
i.e., $p^k_{\texttt{h}}(t)\in M^k_{\texttt{h}}(t) \subset \mathcal{A}$
for all~$t\geq 0$ and~$k\in \mathcal{K}$.
Thus, it is essential that the operators to communicate with the robots
to obtain their local maps.

\begin{remark}\label{rm:map-model}
The exact representation of the map depends on the underlying algorithm
and software used for SLAM,
e.g., the occupancy map in~\cite{thrun2002probabilistic},
or  the semantic map in~\cite{tian2022kimera}.
More details can be found in the numerical experiments.
\hfill $\blacksquare$
\end{remark}

\subsection{Limited Inter-operator-robot Communication}\label{subsec:com}
Furthermore, as shown in Fig.~\ref{fig:comm_model}, each robot and operator is equipped
with a wireless communication unit for robot-robot-operator information exchange,
e.g., ad-hoc networks, UWB, radio or 5G network~\cite{klaesson2020planning,xia2023relink}.
The following model of communication is applied
to the robotic fleet and operators.

\begin{definition}\label{def:com}
Robot~$i\in \mathcal{N}$ can communicate with robot~$j\in \mathcal{N}$ at
time~$t>0$, if the communication quality such as SNR is above
a certain threshold, i.e.,
\begin{equation}\label{eq:com}
  (D^+_i, D^+_j) \triangleq \texttt{Com}\big(D_i, D_j\big),
  \text{\, if \,}\texttt{Qual}(p_i,p_j,\mathcal{A}) > \underline{\delta};
\end{equation}
where~$\texttt{Com}(\cdot)$ is the communication module; $D_i(t)\in \mathcal{D}$ is the
{local} data at robot~$i$ at~$t\geq 0$, similar for robot~$j$;
$\mathcal{D}$ includes map data, video stream, coordination messages
and requests; $\texttt{Qual}(\cdot)$ evaluates the quality given positions
and workspace; $\underline{\delta}>0$ is a lower bound on
the quality; and~$D^+_i(t),\, D^+_j(t)$ are the updated data
after communication.
\hfill $\blacksquare$
\end{definition}

By default, $(p_i,\,M_i)\in D_i$ holds for all robots and operators.
Denote by~$\mathcal{C}_i(t)\subset \mathcal{N}$ the set of robots that robot~$i$
can communicate with at time~$t$ under Def.~\ref{def:com}.
The operator-robot and operator-operator communications are defined
similarly and omitted for brevity. Moreover, several communicating
robots or operators can form a communication {chain} to relay
information reliably and consistently.

\begin{definition}\label{def:chain}
A \emph{communication chain} from robot~$i_0\in \mathcal{N}$ to the
operator~$\texttt{h}$ at time~$t\geq 0$ is given by:
\begin{equation}\label{eq:chain}
\overline{i\texttt{h}} \triangleq i_0 i_1 \cdots i_{\ell} \cdots i_L \texttt{h},
\end{equation}
where neighboring nodes~$i_{\ell} \in \mathcal{C}_{i_{\ell+1}}(t)$ satisfy
the condition in~\eqref{eq:com},~$\forall \ell\in \{0,\cdots,L-1\}$;
and~$i_{L}\in \mathcal{C}_{\texttt{h}}(t)$.
\hfill $\blacksquare$
\end{definition}

\begin{remark}\label{rm:requests}
The above two communication modes are useful for different purposes:
the pairwise communication by~\eqref{eq:com} is suitable for
intermittent data exchange, while the communication
chain by~\eqref{eq:chain} is essential for an operator to have
a consistent and reliable access to a robot for a duration.
  \hfill $\blacksquare$
\end{remark}

\subsection{Operator-fleet Online Interaction}\label{subsec:robot-human}
Both the operators and the robots are capable of sending and receiving
online \emph{requests} to each other via the communication module above.
More specifically, as summarized in Table~\ref{tab:requests},
seven types of requests are considered across three cases.

\textbf{Operator-robot requests}. The operators can send {four} types of
requests~$\xi^k_{1,2,3,4}$.
(I) {Intra-team latency constraint}~$\{\xi^k_1\}$. The operator~$\texttt{h}_k$ must
obtain the local maps of all robots in its team~$\mathcal{T}_k$ with a
latency smaller than a given threshold~$T^k_{\texttt{h}}>0$ at all times,
i.e., $\bigcup_{i\in \mathcal{N}_k} M_i(t) \subset M^k_{\texttt{h}}(t+T^k_{\texttt{h}})$, $\forall t \geq 0$.
This means the explored map by any robot must be known to the
operator~$\texttt{h}_k$ latest by time~$t+T^k_{\texttt{h}}$. This request remains
active once issued;
(II) {Prioritized or avoided region}~$\{\xi^k_2\}$. A certain region
$\xi^k_2\subset \mathcal{A}$ should be prioritized or avoided during
exploration by team~$\mathcal{N}_k$. Fulfilled once the locations are
all explored;
(III) {Operator movement}~$\{\xi^k_3\}$. The operator~$\texttt{h}_k$ requests
to move to a desired location $p^{k,\star}_{\texttt{h}}$. Fulfilled when the
operator reaches $p^{k,\star}_{\texttt{h}}$;
(IV) {Remote access or tele-operation}~$\{\xi^k_4\}$. The operator~$\texttt{h}_k$
requests remote access to robot~$i^\star\in \mathcal{N}_k$ at location~$p^\star$
for duration~$T_{i^{\star}}>0$ as $\xi^k_4\triangleq (i^\star,\, p^\star,\, T_{i^{\star}})$.
Fulfilled if a communication chain by~\eqref{eq:chain} is established from
the operator to robot~$i^\star$ via relays, and uninterrupted for duration~$T_{i^{\star}}$
.

\begin{table}[t!]
\centering
\caption{Summary of Online Operator-fleet Requests}
\label{tab:requests}
\small
\begin{tabularx}{0.95\linewidth}{@{}c c X@{}}
\toprule
\textbf{Request} & \textbf{Type} & \textbf{Description and Notation} \\
\midrule
\multirow{2}{*}{\centering $\{\xi^k_1\}$} & \multirow{2}{*}{\centering Operator-robot} & Intra-team latency of team $\mathcal{T}_k$ should be bounded by $T^k_{\texttt{h}}$.\\
\multirow{2}{*}{\centering $\{\xi^k_2\}$} & \multirow{2}{*}{\centering Operator-robot} & Regions $\mathcal{P}^+$ and $\mathcal{P}^-$ should be prioritized/explored and avoided. \\
\multirow{2}{*}{\centering $\{\xi^k_3\}$} & \multirow{2}{*}{\centering Operator-robot} & Dynamic movement of operator $\texttt{h}_k$ to the new location as $p^{k,\star}_{\texttt{h}}$.\\
\multirow{2}{*}{\centering $\{\xi^k_4\}$} & \multirow{2}{*}{\centering Operator-robot} & Communication chain to robot~$i^\star$ at~$p^\star$ for time $T_{i^{\star}}$.\\
\midrule
\multirow{2}{*}{\centering $\{\xi^i_5\}$} & \multirow{2}{*}{\centering Robot-operator} & Data $D_i$ should be modified by operator and returned as~$D^+_i$. \\
\multirow{2}{*}{\centering $\{\xi^i_6\}$} & \multirow{2}{*}{\centering Robot-operator} & Direct assistance to robot~$i$ at position $p_i$ for duration~$T_i$. \\
\midrule
\multirow{2}{*}{\centering $\{\xi^{k_1k_2}_7\}$} & \multirow{2}{*}{\centering Op-operator} & Inter-team latency
of team $\mathcal{T}_{k_1}$ and $\mathcal{T}_{k_2}$ should be bounded by $T_{\texttt{c}}$. \\
\bottomrule
\end{tabularx}
\end{table}

\textbf{Robot-operator requests}. Any robot can send {two} types of
requests~$\xi^i_{5,6}$ to the operators.
(I) {Confirmation and modification}~$\{\xi^i_5\}$. Robot~$i$ requests an
immediate confirmation from any operator regarding actions and features,
denoted by~$\xi^i_5\triangleq D_i$. Fulfilled if $D_i$ reaches any operator
and is returned as modified data~$D^+_i$;
(II) {Direct assistance} $\{\xi^i_6\}$. Robot~$i$ requests assistance
in unexpected situations~$\xi^i_6\triangleq (p_i,\, T_i)$. Fulfilled if a
communication chain is established from \emph{any} operator to robot~$i$ at
$p_i\in M_i$, uninterrupted for duration~$T_i>0$.

\textbf{Operator-operator requests}. Each operator~$\texttt{h}_{k_1}$ can send
{one} type of request to another operator~$\texttt{h}_{k_2}$,
as the {inter-team latency constraint}~$\{\xi^{k_1k_2}_{7}\}$. One robot
in team~$\mathcal{N}_{k_1}$ must communicate with one robot in team~$\mathcal{N}_{k_2}$
at least once every period of~$T_{\texttt{c}}>0$. This request remains active
once issued.

Thus, the set of all requests received by~$t\geq 0$ is denoted by
$\Xi(t)\triangleq \{\{\xi^k_{1,2,3,4}\},\{\xi^i_{5,6}\},\{\xi^{k_1k_2}_7\}\}$,
which are unknown beforehand. Each request is removed from~$\Xi(t)$
once fulfilled.

\begin{remark}\label{rm:requests_online}
Both the intra-team latency~$\{\xi^k_1\}$ and the inter-team latency~$\{\xi^{k_1k_2}_7\}$
are treated as online requests despite being perpetually active once issued.
This is because both bounds as $T^k_{\texttt{h}}$ and~$T_{\texttt{c}}$
can be modified online according to exploration progress,
e.g., increased as subgroups move further away and the newly-explored
area decreases.
  \hfill $\blacksquare$
\end{remark}

\subsection{Problem Statement}\label{subsec:problem}
Consequently, the behavior of each robot or an operator can be described
by a timed sequence of navigation and communication events as their local
plans.
\begin{definition}\label{def:plan}
  The \emph{local plan} of robot~$i\in \mathcal{N}$ is denoted by:
\begin{equation}\label{eq:plan}
\Gamma_i \triangleq \boldsymbol{\tau}^0_i \mathbf{c}^0_i\, \boldsymbol{\tau}^1_i\, \mathbf{c}^1_i
\, \boldsymbol{\tau}^2_i\,\mathbf{c}^2_i \, \cdots,
\end{equation}
where~$\mathbf{c}^\ell_i \triangleq (p^\ell_i,\, i^\ell_i,\, D^\ell_i)$
is a communication event defined by the communication location, the
other robot or the operator, and the exchanged data from~\eqref{eq:com};
the navigation path~$\boldsymbol{\tau}^\ell_i\subset \mathcal{A}$
is the timed sequence of waypoints that robot~$i$ has visited between the
consecutive communication events; and~$\ell>0$ records the round of
communication events.
\hfill $\blacksquare$
\end{definition}

Note that the behavior of an operator~$\texttt{h}_k$ is defined similarly
to~$\Gamma_k$, for all $k \in \mathcal{K}$.
For brevity, let
$\widehat{\Gamma} \triangleq \{\Gamma_i\} \cup \{\Gamma_k\}$
denote the joint plan of all robots and operators, and write
$\widehat{\Gamma} \models \Xi(t)$ if the behaviors of the
multi-operator multi-robot system collectively fulfill the
requirements of the requests in~$\Xi(t)$.
Accordingly, the overall mission-level objective can be stated as
follows:
\begin{subequations} \label{eq:problem}
 \begin{align}
  &\mathop{\mathbf{min}}\limits_{\widehat{\Gamma},\, \overline{T}}\; \overline{T} \notag\\
  \textbf{s.t.}\quad
  & \mathcal{A} \subseteq M^k_{\texttt{h}}(\overline{T}),\;\forall k, \label{subeq:terminal}\\
  & \eqref{eq:slam}-\eqref{eq:plan},\;\forall i,\forall k,\quad
    \widehat{\Gamma} \models \xi,\;\forall \xi \in \Xi,
    \label{subeq:system_requests}
 \end{align}
\end{subequations}
where~$\widehat{\Gamma}$ is the joint plan of all robots and operators,
and $\overline{T}>0$ is the time instant when the environment is
completely mapped and the resulting map is available to all
operators, as required by~\eqref{subeq:terminal}. Constraint
\eqref{subeq:system_requests} enforces both the operational
constraints and the fulfillment of all requests in~$\Xi$.

Problem~\eqref{eq:problem} defines the overall mission-level objective.
In the online setting considered here, however, the request set is
revealed incrementally during execution, and future operator motions,
robot encounters, and communication opportunities are not known
a priori. Therefore, MoRoCo does not solve~\eqref{eq:problem}
directly as a globally optimal planning problem. Instead, it realizes
this mission objective through a sequence of online feasible
coordination decisions on the current communication backbone, as
detailed in Sec.~\ref{subsec:online}.

\begin{remark}\label{rm:prob}
The consideration of multiple teams, each with one operator and a few
robots, is motivated by several practical concerns:
(I) The workspace is large and complex, making it suitable for several
medium-size fleets to explore in parallel;
(II) One operator has limited attention to supervise and interact with
too many robots, as revealed by several recent studies in~\cite{dahiya2023survey,villani2020humans};
and (III) The response to online requests within one team is more efficient
compared with a global team.
More numerical comparisons can be found in the experiments.
\hfill $\blacksquare$
\end{remark}

\section{Proposed Solution}\label{sec:solution}
As illustrated in Fig.~\ref{fig:method}, {MoRoCo} consists of
three coupled components. Sec.~\ref{subsec:spread} presents the
latency-bounded intermittent communication backbone and the
detach-and-rejoin procedure for dynamic resizing. Then,
Sec.~\ref{subsec:requests} introduces the request-consistent
embedded-graph instantiation layer. Sec.~\ref{subsec:online} presents
the online coordination strategy for handling multiple simultaneous
requests. Generalizations and limitations are discussed in
Sec.~\ref{subsec:discuss}.

\subsection{Latency-bounded Intermittent Communication Protocol with
Dynamic Resizing}\label{subsec:spread}
Timely information sharing between the operators and the fleet is
essential for online request handling. MoRoCo therefore first
establishes a latency-bounded intermittent communication backbone on a
joint-wheel graph, and augments it with a distributed
detach-and-rejoin mechanism for dynamic resizing. Together, these
components provide the backbone for the
request-instantiation and online coordination layers.

\subsubsection{Latency-bounded Intermittent Communication Protocol}
\label{subsubsec:spread-problem}
Consider a single team $\mathcal{T}$ with the operator~$\texttt{h}$ and the robots~$\mathcal{N}$,
tasked with the information gathering tasks
$\mathcal{F}\triangleq \{f_m\}$, where each task $f_m$ specifies a
location to be visited for information, such as exploration.
The information collected by \emph{any} robot~$i$ at time~$t$
must reach the operator within a predefined latency bound~$T_\texttt{h}>0$.
Let $\delta_\texttt{h}(t)>0$ denote the maximum latency over all the robots at
time~$t$.

\begin{problem}\label{prob:latency}
Given the tasks~$\mathcal{F}$ and the bound $T_\texttt{h}$, synthesize
the local plans $\{\Gamma_i\}$ such that the completion time of $\mathcal{F}$ is
minimized and the latency~$\delta_\texttt{h}(t)\leq T_\texttt{h}$ holds
for all $t>0$.
\hfill $\blacksquare$
\end{problem}

The core idea is to enforce bounded latency through scheduled pairwise communication.
The robots exchange the information at the communication events and replan the next
event so that a timely return to the operator remains feasible within
$T_\texttt{h}$.
This logic is captured conveniently by an embedded joint-wheel coordination
graph defined as follows.

\begin{definition}\label{def:embedded_graph}
An \emph{embedded graph} is a tuple
$\mathcal{G}\triangleq (\mathcal{V},\, \mathcal{E},\, \sigma,\, x)$, where
$\mathcal{V}$ is the vertex set;
$\mathcal{E}\subseteq \mathcal{V}\times\mathcal{V}$ is the edge set;
$\sigma\triangleq (\sigma_\mathcal{V},\sigma_\mathcal{E})$ embeds vertex and edge
into attributes~$(\Sigma_\mathcal{V},\Sigma_\mathcal{E})$;
and $x:\mathcal{V}\rightarrow\boldsymbol{X}_\mathcal{V}$ is the geometric embedding.
\hfill $\blacksquare$
\end{definition}

\begin{definition}\label{def:ejw}
An \emph{embedded joint-wheel coordination graph} is an embedded graph
$\mathcal{G}^{\texttt{w}}\triangleq (\mathcal{V},\, \mathcal{E},\, \sigma,\, x)$ such that:
(I) $\mathcal{V}\triangleq \mathcal{N}\cup\{\texttt{h}\}$;
(II) $\mathcal{E}\triangleq \mathcal{E}_\texttt{r}\cup\mathcal{E}_\texttt{h}$, where
$\mathcal{E}_\texttt{r}$ forms a ring over $\mathcal{N}$, i.e.,
$(k_\ell,k_{\ell+1})\in\mathcal{E}_\texttt{r}$ for $\ell=1,\cdots,N$ with
$k_{N+1}\triangleq k_1$, and
$\mathcal{E}_\texttt{h}\triangleq \{(\texttt{h},i)\mid i\in\mathcal{N}\}$.
\hfill $\blacksquare$
\end{definition}

\begin{figure}[t!]
  \centering
  \includegraphics[width=0.9\linewidth]{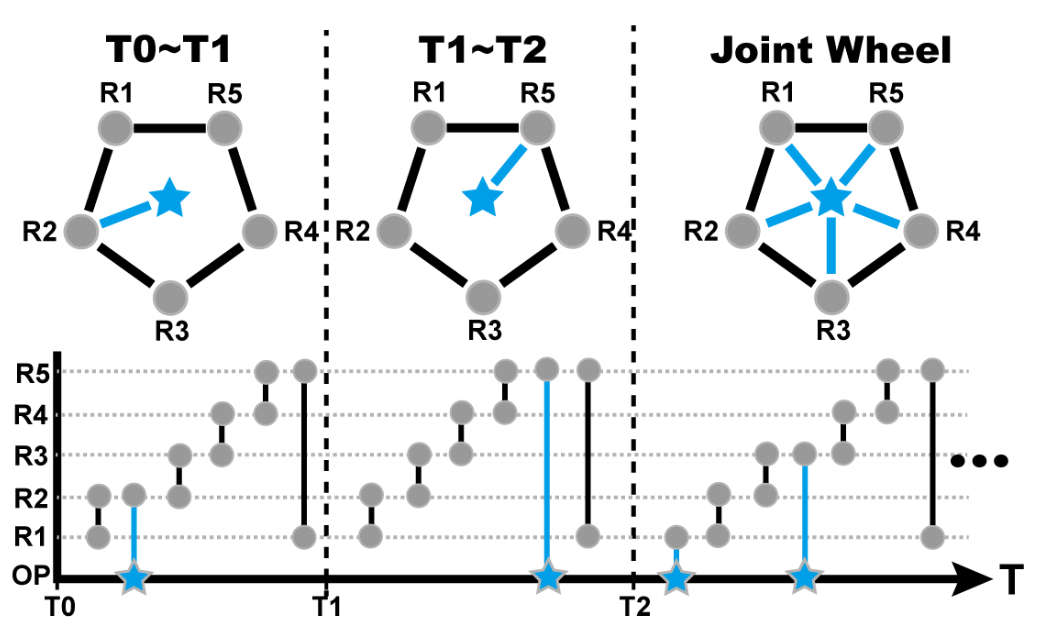}
  \vspace{-2mm}
  \caption{
  \textbf{Top:} the joint-wheel communication graph over time;
  \textbf{Bottom:} the pairwise meeting events (black vertical lines)
  and return events (blue lines).
  }\label{fig:joint-wheel}
\end{figure}

\begin{remark}\label{rem:joint-wheel}
As illustrated in Fig.~\ref{fig:joint-wheel}, the term \emph{joint-wheel} emphasizes a joint connectivity
rather than a static graph. At any time instant, the communication protocol activates
at most one rim edge in $\mathcal{E}_\texttt{r}$, hence only one robot pair
communicates.
Over a period longer than~$T_{\texttt{h}}$,
the union of all rim edges in $\mathcal{E}_\texttt{r}$ and hub edges in $\mathcal{E}_\texttt{h}$ induces a
wheel-like connectivity pattern. This should be distinguished from the
static line and star graphs introduced later.
\hfill $\blacksquare$
\end{remark}

The embedded graph $\mathcal{G}^{\texttt{w}}$ above is constructed and maintained
in a fully distributed way via local coordination
at intermittent communication events as described below.

\textbf{Initialization.}
To begin with, the vertex embedding encodes the local coordination state, defined as:
\begin{equation}\label{eq:def-vertex-embed}
\sigma_\mathcal{V}(i)\triangleq (\Gamma_i,\, \chi_i),\quad
\sigma_\mathcal{V}(\texttt{h})\triangleq (\widehat{p}_\texttt{h},\, \chi_\texttt{h}),
\end{equation}
where the local plan $\Gamma_i$ specifies the sequence of
visiting locations and communication events for robot $i$ as in~\eqref{eq:plan};
the latency variable~$\chi_i\in \mathbb{R}^{N}$ estimates the minimum time stamp of
information from each robot that is guaranteed to reach the operator;
and $\widehat{p}_\texttt{h}\in \mathbb{R}^2$ is the current target location of the operator,
while~$\chi_{\texttt{h}}\in \mathbb{R}^{N}$ is the confirmed time stamp of each robot.
The initialization assigns $\Gamma_i=\varnothing$ and $\chi_i=\mathbf{0}$ for all
$i\in\mathcal{N}$, with $\widehat{p}_\texttt{h}=\varnothing$ and $\chi_\texttt{h}=\mathbf{0}$.
Moreover, the edge embedding encodes the scheduled communication events as:
\begin{equation}\label{eq:def-edge-embed}
\sigma_\mathcal{E}(i,j)\triangleq (t^+_{ij},\, p^+_{ij}),\quad \forall (i,j)\in\mathcal{E}_\texttt{r};
\end{equation}
and $\sigma_\mathcal{E}(\texttt{h},i)\triangleq (t^+_{\texttt{h}i}, \, p^+_{\texttt{h}i})$
are similar for each~$i\in \mathcal{N}$.
The initial communications are set
to~$\sigma_\mathcal{E}(i,j)=(0,\, p_{\texttt{h}})$,
as all robots start around the operator location and communicate immediately.
The value~$\sigma_\mathcal{E}(\texttt{h},i)=\varnothing$ indicates that no return is
scheduled initially.
Lastly, the geometric embedding
assigns each vertex its physical state and its local data, written as
$x(i)\triangleq (p_i(t),\, D_i(t))$ and $x(\texttt{h})\triangleq (p_\texttt{h}(t),\, D_\texttt{h}(t))$.

\begin{figure}[t!]
  \centering
  \includegraphics[width=0.9\linewidth]{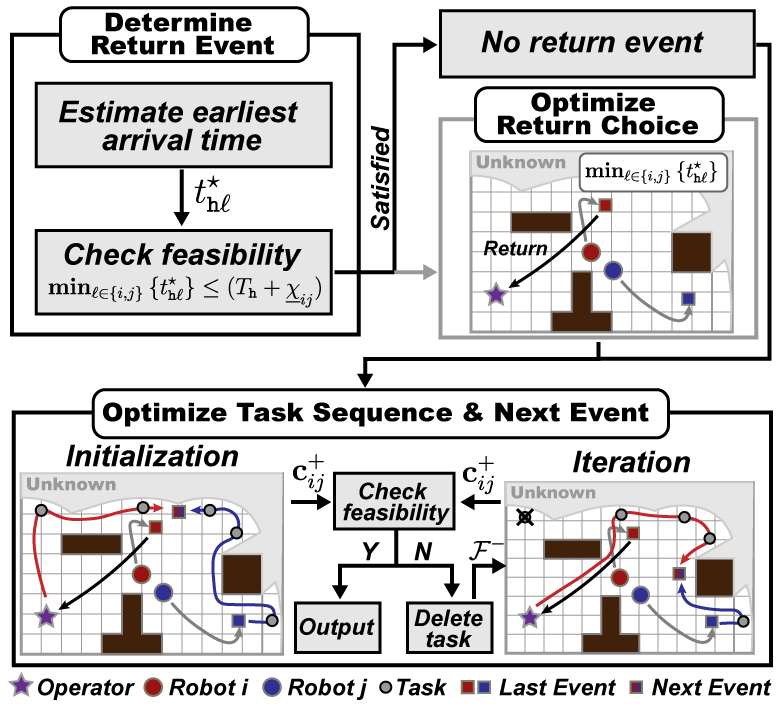}
  \vspace{-2mm}
  \caption{The proposed pairwise coordination scheme
    via intermittent communication.
    \textbf{Top}: return event to the operator is determined first;
    \textbf{Bottom}: the next meeting event is optimized given
    the current exploration tasks.
  }\label{fig:pairopt}
\end{figure}

\textbf{Pairwise Coordination.}
At a communication event $\mathbf{c}_{ij}$, the robots read the current geometric
states $x(i)$ and $x(j)$, exchange the incident embeddings, and compute the fused
map~$\mathcal{M}_{ij}$. The exchange yields the merged latency vector:
\begin{equation}\label{eq:chi-merge}
\chi_{ij}\triangleq \textbf{max}\{\chi_i,\, \chi_j\},
\end{equation}
where the maximum operation is taken component-wise. Then, the coordination computation
proceeds in three stages, as summarized in Alg.~\ref{alg:coordination} and Fig.~\ref{fig:pairopt}.

\emph{Stage-I: Determine Return Event.}
The quantity~$\underline{\chi}_{ij}\triangleq \textbf{min}_{k\in\mathcal{N}} \{\chi_{ij}(k)\}$,
is the minimum team time stamp that is guaranteed to reach the
operator, as inferred at $\mathbf{c}_{ij}$.
Let~$t^{L_\ell}_\ell$ and~$p^{L_\ell}_\ell$ denote the terminal time and location
in the local plan~$\Gamma_\ell$ of robot~$\ell\in\{i,j\}$.
The earliest arrival time at the
operator for a candidate return robot $\ell\in\{i,j\}$ is estimated as:
\begin{equation}\label{eq:estret}
t^\star_{\texttt{h}\ell}\triangleq t^{L_\ell}_{\ell}+T^{\texttt{nav}}_\ell(p^{L_\ell}_\ell,\widehat{p}_\texttt{h}),
\end{equation}
where $T^{\texttt{nav}}_\ell(\cdot,\cdot)$ is the navigation-time model and the
detailed construction is provided in the Supplementary file.
A return to the operator is {unnecessary} if:
\begin{equation}\label{eq:return_cond}
\textbf{min}_{\ell\in\{i,j\}} \, \{t^\star_{\texttt{h}\ell}\}\leq (T_\texttt{h}+\underline{\chi}_{ij}),
\end{equation}
and otherwise the return robot is selected as
$n\triangleq \textbf{argmin}_{\ell\in\{i,j\}}\, \{t^\star_{\texttt{h}\ell}\}$,
where $n$ denotes the robot that returns to the operator immediately after its last event.

\emph{Stage-II: Optimize Task Sequence and Next Communication Event.}
Conditioned on the return decision, the optimization produces the tailing plans
for both robots:
$\Gamma^+_i\triangleq \tau^+_i\mathbf{c}^+_{ij}$ and $\Gamma^+_j\triangleq \tau^+_j\mathbf{c}^+_{ij}$,
where $\tau^+_i$ and $\tau^+_j$ are the optimized task sequences and
$\mathbf{c}^+_{ij}$ is the next communication event shared by the two robots.
The schedule of $\mathbf{c}^+_{ij}$ is set to~$(t^+_{ij},p^+_{ij})$,
which is constrained locally due to the latency bound:
\begin{equation}\label{eq:pairing_cond}
t^+_{ij}+T^{\texttt{nav}}_\ell(p^+_{ij},\widehat{p}_\texttt{h})
\leq T_\texttt{h}+\chi^\star,
\end{equation}
where $\chi^\star\triangleq \underline{\chi}_{ij}$ if no return is scheduled in
Stage-I, and $\chi^\star$ is computed using the scheduled return otherwise.
Feasible outputs $(\tau^+_i,\tau^+_j,\mathbf{c}^+_{ij})$ are obtained by an
iterative TSP-based procedure. At each iteration, a TSP is constructed over the
remaining task locations in $\mathcal{F}^-$ together with the current terminal
locations of robots $i$ and $j$. The edge weights are given by the navigation
time model~$T^{\texttt{nav}}(\cdot)$. The TSP solution yields a joint
visiting order $\mathbf{p}_{ij}$, from which $\tau^+_i$ and $\tau^+_j$ are
extracted, and a candidate $\mathbf{c}^+_{ij}$ is selected along $\mathbf{p}_{ij}$.
Feasibility of the meeting event is checked by~\eqref{eq:pairing_cond}.
If violated, a task in~$\tau^+_i$ or $\tau^+_j$ is removed and the procedure repeats.

\emph{Stage-III: Update Embeddings.}
Consequently, the current event $\mathbf{c}_{ij}$ is removed from the plans to obtain the
prefixes $\Gamma^-_i$ and $\Gamma^-_j$. The updated plans satisfy
$\widehat{\Gamma}_i=\Gamma^-_i\Gamma^+_i$ and $\widehat{\Gamma}_j=\Gamma^-_j\Gamma^+_j$.
The latency vectors are updated by the information exchange as
$\widehat{\chi}_i\triangleq \widehat{\chi}_j\triangleq \chi_{ij}$,
where $\chi_{ij}$ is defined in \eqref{eq:chi-merge}. The local embedding update
is then given by:
\begin{equation}\label{eq:embed-update}
\begin{aligned}
\sigma_\mathcal{V}(i) &\leftarrow (\widehat{\Gamma}_i,\,\widehat{\chi}_i),\quad
\sigma_\mathcal{V}(j) \leftarrow (\widehat{\Gamma}_j,\,\widehat{\chi}_j);\\
\sigma_\mathcal{E}(i,j) &\leftarrow (t^+_{ij},\,p^+_{ij}),\,
\sigma_\mathcal{E}(\texttt{h},n) \leftarrow (t^\star_{\texttt{h}n},\,\widehat{p}_\texttt{h});\\
\end{aligned}
\end{equation}
where the last line is applied only if Stage-I schedules a return, and the
hub-edge embedding remains $\varnothing$ otherwise.

Combined with the joint-wheel communication topology,
the above procedure ensures exploration with bounded latency.
as shown in the following lemmas and theorem.
The detailed proofs are provided in the Appendix.

\begin{lemma}\label{prop:latency}
  The latency constraint~$\delta_\texttt{h}(t)\leq T_\texttt{h}, \forall t>0$ is satisfied if
  for every return event $\mathbf{c}_{h,n_i}\in \Gamma_{\texttt{h}}$, it holds that:
  (I) $t_{\texttt{h}}(n_i) \leq T_{\texttt{h}} + \chi_{n_{i-1}}$ and (II) $\chi_{n_i} > \chi_{n_{i-1}}$.
\end{lemma}

\begin{lemma}\label{prop:exist-solution}
  The iteration process in Alg.~\ref{alg:coordination} ends in finite steps,
  and always returns a feasible plan that satisfies
  the meeting event constraint~\eqref{eq:pairing_cond} during each pairwise coordination.
\end{lemma}

\begin{theorem}\label{prop:Ts-constraint}
  Under the joint-wheel communication topology
  and the coordination protocol in Alg.~\ref{alg:coordination},
  the intra-team latency constraint is satisfied,
  i.e.,~$\delta_\texttt{h}(t)\leq T_\texttt{h}$ holds for all $t>0$.
\end{theorem}

\begin{algorithm}[t]
  \caption{Latency-aware Intermittent Pair-wise Coordination:
  $\texttt{PairCoord}(\cdot)$}
  \label{alg:coordination}
	\LinesNumbered
        \SetKwInOut{Input}{Input}
        \SetKwInOut{Output}{Output}
\Input{Local plans~$\Gamma_i$,\, $\Gamma_j$,\, Stamp $\chi_{ij}$,\, Map $\mathcal{M}_{ij}$, Unassigned tasks~$\mathcal{F}^-$.}
\Output{Revised local plans~$\widehat{\Gamma}_i$,\, $\widehat{\Gamma}_j$; next event~$\mathbf{c}_{ij}$}
\tcc{\textbf{Determine Return Event.}}
  Compute~$t^\star_{\texttt{h}\ell}$ by~\eqref{eq:estret} for $\ell\in\{i,j\}$\;
  \If{condition \eqref{eq:return_cond} not satisfied}{
    Determin return robot by~$n\triangleq \texttt{argmin}_{\ell\in\{i,j\}}\, \{t^\star_{\texttt{h}\ell}\}$\;
    Append a return event to~$\Gamma_n$\;
  }
  \tcc{\textbf{Optimize Task Sequence and Next Pairing Event.}}
  $\mathbf{p}_{ij}\leftarrow \texttt{TSP-Path}(p^{L_i}_i, \mathcal{F}^-, p^{L_j}_j)$\;
  $\mathbf{c}_{ij}\leftarrow \texttt{SelComm}(\mathbf{c}^{L_i}_i, \mathbf{c}^{L_j}_j, \mathbf{p}_{ij})$\;
   \While{condition~\eqref{eq:pairing_cond} not satisfied}{\label{alg_line:continue_itr}
   Remomve $f^\star$ with maximum cost from $\mathcal{F}^-$\;\label{alg_line:remove_frontier}
   Recompute $\mathbf{p}_{ij}$ and $\mathbf{c}_{ij}$\;
  }
  $\boldsymbol{\tau}^{+}_i,\, \boldsymbol{\tau}^{+}_j \leftarrow
  \texttt{Split}(\mathbf{p}_{ij},\, \mathbf{c}_{ij})$\;\label{alg_line:split}
  \tcc{\textbf{Update Local Plans.}}
  $\Gamma^-_\ell \leftarrow \texttt{Remove}(\Gamma_\ell,\, \mathbf{c}_{ij})$, for~$\ell\in\{i, j\}$\;
  $\widehat{\Gamma}_\ell \leftarrow \Gamma^-_\ell + \boldsymbol{\tau}^{+}_\ell + \mathbf{c}_{ij}$, for~$\ell\in\{i, j\}$\;
  Update embeddings via~\eqref{eq:embed-update}\;
\end{algorithm}

\begin{remark}\label{rem:corner-cases}
The joint-wheel graph above typically consists of at least three robots.
For $|\mathcal{N}|=1$, the robot explores independently and returns to the operator
periodically by $T^k_{\texttt{h}}$. For $|\mathcal{N}|=2$, the same protocol is
applied on a line topology and the local plan reduces to $\Gamma_i\triangleq \tau_i\mathbf{c}_i$.
\hfill $\blacksquare$
\end{remark}

\subsubsection{Distributed Detach-and-Rejoin Mechanism}\label{sec:detach-join}
\begin{figure}[t!]
  \centering
  \includegraphics[width=0.95\linewidth]{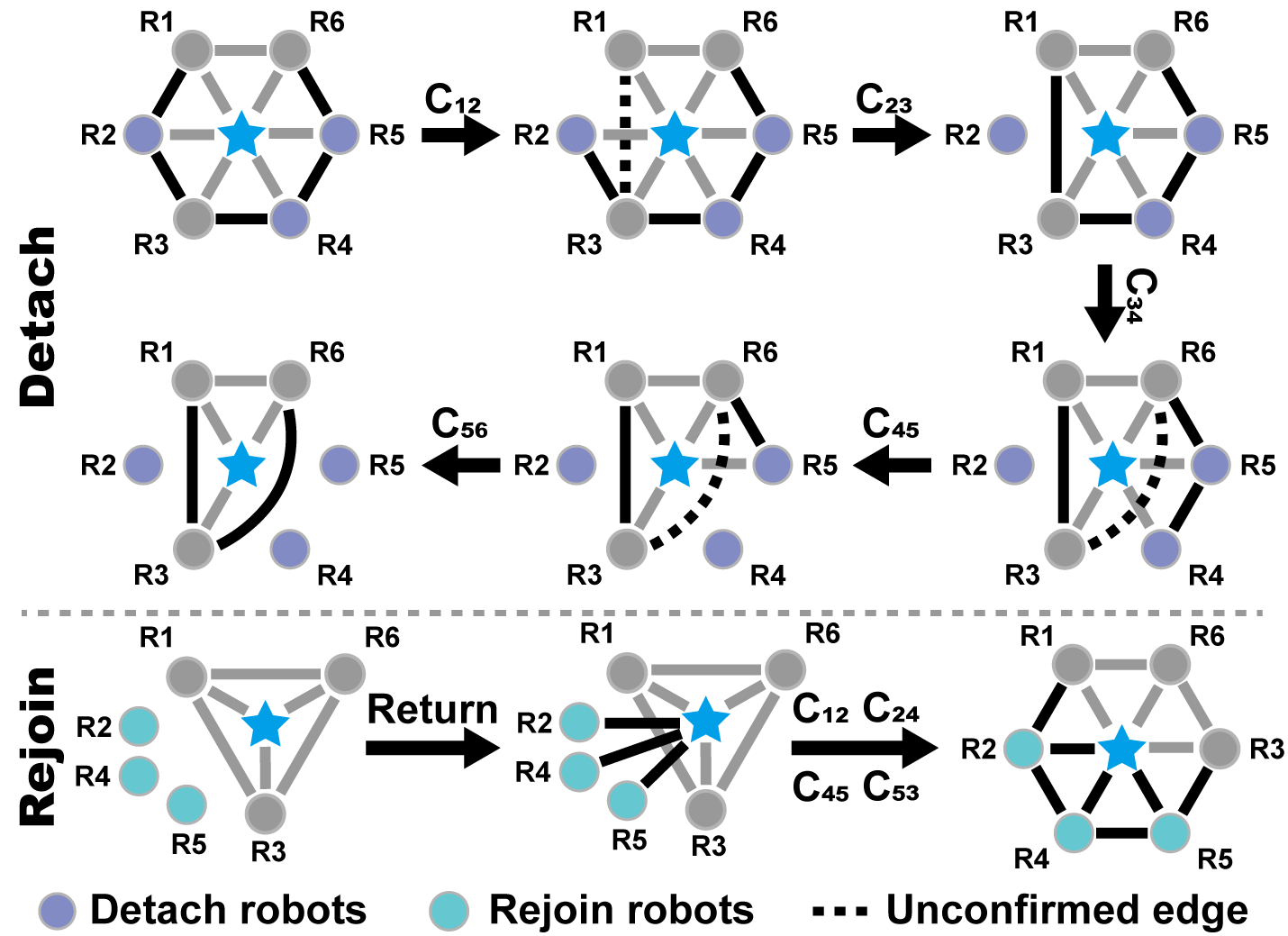}
  \vspace{-2mm}
  \caption{The distributed schemes of
    detach (\textbf{Top}) and rejoin (\textbf{Bottom}),
    where the edges can be involved (black), unrelated (gray),
    and unconfirmed (dashed).
  }\label{fig:detach-rejoin}
\end{figure}

Flexible resizing under intermittent communication is enabled through a
distributed detach-and-rejoin procedure that preserves the wheel topology and
respects latency constraints. A central design choice is to complete rewiring
through \emph{two consecutive meeting events}, ensuring that successor robots
receive consistent plan and adjacency updates before a robot leaves or rejoins
the wheel, which is illustrated in Fig.~\ref{fig:detach-rejoin}.

(I) \textbf{To detach},
consider three consecutive robots~$(i,j,k)$ along the wheel, where robot~$j$
intends to detach. Let $\Gamma_\ell$ denote robot~$\ell$'s nominal local plan and
$\widehat{\Gamma}_\ell$ the optimized plan maintained online.
At the meeting between robots~$i$ and~$j$, robot~$j$ does not depart immediately.
Instead, using robot~$i$'s most recent knowledge of robot~$k$'s plan $\Gamma_k$,
robot~$i$ computes a bypass schedule that directly connects $(i,k)$,
via
$\widehat{\Gamma}_i, \widehat{\Gamma}_k, \mathbf{c}_{ik}
= \texttt{PairCoord}(\Gamma_i,\Gamma_k)$;
and updates the edge embeddings by removing $(i,j)$ and inserting $(i,k)$, i.e.,
\[
\sigma_{\mathcal{E}}(i,j) \leftarrow \emptyset,\qquad
\sigma_{\mathcal{E}}(i,k) \leftarrow (t_{ik},p_{ik}),
\]
where $(t_{ik},p_{ik})$ is the time--location pair of the next event
$\mathbf{c}_{ik}$. Robot~$j$ then follows its original schedule to meet robot~$k$
and forwards the bypass message so that robot~$k$ adopts $\widehat{\Gamma}_k$ and
records the updated predecessor implied by $\mathbf{c}_{ik}$. Only after this
second meeting does robot~$j$ finalize detachment and depart to remote location.
Thus, the detach event is denoted by:
\begin{equation}
  \pi_j \triangleq (j,\, t_{jk},\, p_{jk}),
\end{equation}
where $(t_{jk},p_{jk})$ is the confirmed meeting between $(j,k)$
as the time and location for robot~$j$ to detach from~$\mathcal{G}_k$.

(II) \textbf{To rejoin}, robot~$j$ first returns to the operator~$\texttt{h}$ and waits until a
team robot~$m$ also arrives at~$\texttt{h}$. Reinsertion is completed by two
consecutive meetings. Let~$\ell$ denote robot~$m$'s current successor on the
wheel. First, robots~$m$ and~$j$ execute pair-wise coordination at~$\texttt{h}$ to
synchronize plans and schedule the next event as
$
\widehat{\Gamma}_m, \widehat{\Gamma}_j, \mathbf{c}_{mj}
= \texttt{PairCoord}(\Gamma_m,\Gamma_j)$.
Then, the embeddings are updated by replacing the successor edge $(m,\ell)$ with
$(m,j)$, i.e.,
\[
\sigma_{\mathcal{E}}(m,\ell) \leftarrow \emptyset,\qquad
\sigma_{\mathcal{E}}(m,j) \leftarrow (t_{mj},p_{mj}),
\]
where~$(t_{mj},p_{mj})$ is from event~$\mathbf{c}_{mj}$. Second, robot~$j$ meets
robot~$\ell$ to establish the new successor relation via
$\texttt{PairCoord}(\Gamma_j,\Gamma_\ell)$, and the edge embedding
$\sigma_{\mathcal{E}}(j,\ell) \leftarrow (t_{j\ell},p_{j\ell})$ is added.
This two-step reinsertion avoids inconsistent topology and restores a valid
wheel graph.

Lastly, for \emph{multiple} detaching or rejoining robots,
the above scheme can be extended by adjusting the number of required
forwarding meetings. For detachment, nonconsecutive robots may detach
sequentially using the same two-meeting handoff. If a consecutive block detaches,
a single bypass plan is computed by the predecessor of the block and is forwarded
across the block until reaching the first non-detaching successor; each
intermediate robot forwards once and then exits. For rejoining, a block waiting
at the operator~$\texttt{h}$ can be inserted between a returning robot~$m$ and its successor
through consecutive pairwise coordination; the last inserted
robot then performs the successor take-over meeting to close the wheel.

\subsection{Request-consistent Embedded Graph Instantiation}
\label{subsec:requests}
Built on the latency-bounded communication, this section
presents the request-instantiation layer of {MoRoCo}. Given the
wheel backbone and an incoming request, it constructs a
request-consistent embedded graph together with the resized backbone and the detach schedule. It then introduces the
procedure to compute this transition, followed by the
request-specific graph constructions and detach--rejoin executions.

\subsubsection{Problem of Request-consistent Graph Instantiation}

Within {MoRoCo}, each request~$\xi$ in
Table~\ref{tab:requests} is realized through a family of candidate
embedded target graphs that encode the desired robot roles and
communication topology, denoted by~$\widehat{\mathcal{G}}_k^{\xi}$.
Given the current embedded wheel backbone graph
$\mathcal{G}_k^{\texttt{w}}$, the request-instantiation problem seeks: a resized wheel backbone $\mathcal{G}_k^{\texttt{w'}}$,
a selected target graph
$\mathcal{G}_k^{\xi}\in\widehat{\mathcal{G}}_k^{\xi}$,
and the corresponding detach events
$\widehat{\pi}_k^{\xi}\triangleq\{\pi_r^{\xi}\}$ that realize the
required topology transition. The performance measure is given by:
\begin{equation}\label{eq:match_objective}
  \underset{\mathcal{G}_k^{\texttt{w'}},\,\mathcal{G}_k^{\xi},\,
  \widehat{\pi}_k^{\xi},\,\Pi_k^{\xi}}{\textbf{min}}
\Big(
T_k^{\texttt{trs}}(\mathcal{G}_k^{\texttt{w}},\,
\mathcal{G}_k^{\texttt{w'}})
+ w_{\texttt{n}} \,
\underset{r \in \mathcal{N}_k^{\xi}}{\textbf{max}}
\left\{ T_r^{\texttt{nav}}(\pi_r^{\xi},\Pi_k^{\xi}) \right\} \Big),
\end{equation}
where
$T_k^{\texttt{trs}}(\mathcal{G}_k^{\texttt{w}},\,
\mathcal{G}_k^{\texttt{w'}})$ is the transition time induced by
resizing the wheel backbone;
$T_r^{\texttt{nav}}(\pi_r^{\xi},\Pi_k^{\xi})$ is the navigation time for robot~$r$
to execute its detach event and reach the role specified by the
assignment $\Pi_k^{\xi}$;
$\mathcal{N}_k^{\xi}$ is the set of robots that switch from
$\mathcal{G}_k^{\texttt{w}}$ to~$\mathcal{G}_k^{\xi}$; and
$w_{\texttt{n}}>0$ trades off the backbone-transition time and the
request-realization time.

\begin{problem}\label{prob:graph_matching}
Given the current wheel backbone graph~$\mathcal{G}_k^{\texttt{w}}$
and the candidate target graphs~$\widehat{\mathcal{G}}_k^{\xi}$ for
request~$\xi$, define a graph instantiation procedure:
\begin{equation}\label{eq:single_graphmatch}
  \mathcal{G}_k^{\texttt{w'}},\,\mathcal{G}_k^{\xi},\,
  \widehat{\pi}_k^{\xi},\,\Pi_k^{\xi}
  = \texttt{GraphMatch}\!\left(\mathcal{G}_k^{\texttt{w}},\,
  \widehat{\mathcal{G}}_k^{\xi},\,\xi\right),
\end{equation}
which realizes a request-consistent topology transition by minimizing
the objective in~\eqref{eq:match_objective}.
\hfill $\blacksquare$
\end{problem}

\begin{algorithm}[t!]
\caption{Request-consistent graph instantiation for a given
request $\texttt{GraphMatch}(\cdot)$}
\label{alg:graphmatch}
\LinesNumbered
\SetKwInOut{Input}{Input}
\SetKwInOut{Output}{Output}

\Input{$\mathcal{G}_k^{\texttt{w}}$, request $\xi$, candidate set
$\widehat{\mathcal{G}}_k^{\xi}$, weight $w_{\texttt{n}}$}
\Output{$\mathcal{G}_k^{\xi}$, $\mathcal{G}_k^{\texttt{w'}}$,
$\widehat{\pi}_r^{\xi}$, $\Pi_k^{\xi}$}

Extract $\mathcal{C}^+_k$ from $\mathcal{G}_k^{\texttt{w}}$\;
\While{candidates in $\widehat{\mathcal{G}}_k^{\xi}$ remain}{
  Select candidate graph $\mathcal{G}_k^{\xi}$ given
  $\mathcal{C}^+_k, \mathbf{p}_0^{\xi}$ by
  \eqref{eq:graphmatch_next_candidate}\;
  Extract $\mathcal{N}_k^{\xi}, \mathbf{p}_0^{\xi}$ from
  $\mathcal{G}_k^{\xi}$\;
  $\mathcal{G}_k^{\texttt{w'}} \leftarrow
  \mathcal{G}_k^{\texttt{w}}[\mathcal{N}_k\setminus
  \mathcal{N}_k^{\xi}]$\;

  $(T_{\texttt{trs}},\,\widehat{\pi}_r^{\xi})
  \leftarrow
  \texttt{MultiDetach}(\mathcal{G}_k^{\texttt{w}},\,
  \mathcal{N}_k^{\xi})$ by \eqref{eq:graphmatch_trs}\;

  $\Pi_k^{\xi} \leftarrow
  \texttt{LBAP}(\mathcal{N}_k^{\xi},\,\mathbf{p}_0^{\xi},\,
  \widehat{\pi}_r^{\xi})$ by \eqref{eq:role_assignment}\;

  Update best $\widehat{\mathcal{G}}_k^{\xi}$ by
  \eqref{eq:match_objective}\;
}
\Return $\mathcal{G}_k^{\xi}$, $\mathcal{G}_k^{\texttt{w'}}$,
$\widehat{\pi}_r^{\xi}$, $\Pi_k^{\xi}$ with the best candidate\;
\end{algorithm}

\begin{figure*}[t!]
  \centering
  \includegraphics[width=0.93\linewidth]{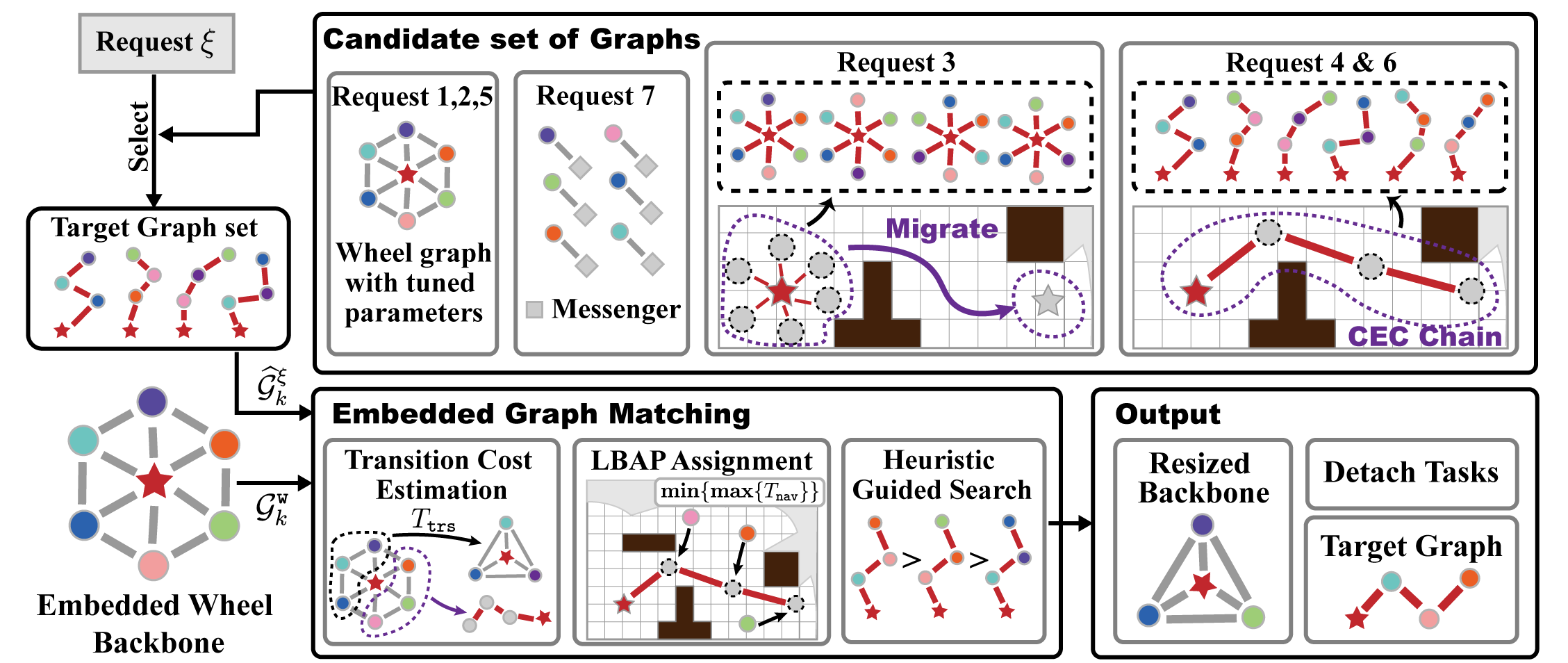}
  \vspace{-2mm}
  \caption{Illustration of the request-consistent graph
  instantiation procedure. Given a request $\xi$,
  \texttt{GraphMatch} selects a request-specific target graph from a
  candidate set and realizes it from the embedded wheel backbone
  $\mathcal{G}_k^{\texttt{w}}$, yielding a resized backbone, the
  instantiated target graph, and the detach events.}
  \label{fig:graph-matching}
\end{figure*}

\subsubsection{Optimization Procedure}
As illustrated in
Fig.~\ref{fig:graph-matching}, given a request-induced candidate set
$\widehat{\mathcal{G}}_k^{\xi}$, it first computes a feasible
transition schedule to resize the wheel backbone, then assigns
detached robots to target roles through an unbalanced linear
bottleneck assignment, and finally traverses the candidate set to
return the solution that minimizes
\eqref{eq:match_objective}.

The first step evaluates the transition time required to resize the
current wheel backbone into a feasible smaller wheel. For a candidate
$\mathcal{G}_k^{\xi}\in\widehat{\mathcal{G}}_k^{\xi}$, let
$\mathcal{N}_k^{\xi}$ denote the robots allocated to
$\mathcal{G}_k^{\xi}$. Removing $\mathcal{N}_k^{\xi}$ from the wheel
backbone $\mathcal{G}_k^{\texttt{w}}$ yields the resized wheel
$\mathcal{G}_k^{\texttt{w'}}$. Conditioned on the return robot that
conveys the request, the detachment routine in
Sec.~\ref{sec:detach-join} is invoked to produce a feasible timed
schedule and its transition time:
\begin{equation}
\Big(T_{\texttt{trs}}(\mathcal{G}_k^{\texttt{w}},\,
\mathcal{G}_k^{\texttt{w'}}),\,
\widehat{\pi}_r^{\xi}\Big)
\triangleq
\texttt{MultiDetach}\!\left(\mathcal{G}_k^{\texttt{w}},\,
\mathcal{N}_k^{\xi}\right),
\label{eq:graphmatch_trs}
\end{equation}
where each $\widehat{\pi}_r^{\xi}\triangleq (r,\,t_r,\,p_r)$
specifies the detach time and detach location of robot~$r$. The
transition completion time $T_{\texttt{trs}}$ is the earliest time
when $\mathcal{G}_k^{\texttt{w'}}$ is fully formed and its embeddings
are updated by~\eqref{eq:embed-update}.

In the second step, the detached robots are assigned to the role
locations of the selected target graph $\mathcal{G}_k^{\xi}$. Let
$\mathbf{p}_0^{\xi}\triangleq
\{p^\xi_{r}\}_{r\in \mathcal{N}_k^{\xi}}$ be the initial role
locations in $\mathcal{G}_k^{\xi}$. An assignment
$\Pi_k^{\xi}:\mathcal{N}_k^{\xi}\to\mathbf{p}_0^{\xi}$ induces the
bottleneck arrival time
$\textbf{max}_{r\in\mathcal{N}_k^{\xi}}
\{t_r+T_\texttt{nav}^{r}(p_r,\Pi_k^{\xi}(r))\}$,
where $T_\texttt{nav}^{r}(p_r,\Pi_k^{\xi}(r))$ is the navigation time
from the detach location to the assigned role. Selecting
$\Pi_k^{\xi}$ to minimize this term yields an unbalanced linear
bottleneck assignment problem (LBAP)~\cite{burkard2012assignment}:
\begin{equation}\label{eq:role_assignment}
  \Pi_k^{\xi} \leftarrow \texttt{LBAP}\!\left(\mathcal{N}_k^{\xi},\,
  \mathbf{p}_0^{\xi},\, \widehat{\pi}_r^{\xi}\right),
\end{equation}
which can be solved efficiently by off-the-shelf tools such as
OR-Tools~\cite{gor}, and returns a concrete role assignment
and the navigation plans for all detached robots.

The final step traverses the candidate set
$\widehat{\mathcal{G}}_k^{\xi}$ using a best-first heuristic. After
evaluating the current candidate $\mathcal{G}_{k,q}^{\xi}$, the next
candidate is selected by ranking the remaining graphs via the
surrogate score below, derived from \eqref{eq:match_objective}:
\begin{equation}\label{eq:graphmatch_next_candidate}
\mathcal{G}_{k,q+1}^{\xi}
=
\underset{\mathcal{G}\in\widehat{\mathcal{G}}_k^{\xi}\setminus
\{\mathcal{G}_{k,q}^{\xi}\}}{\textbf{argmin}}
\big(
\widetilde{T}_k^{\texttt{trs}}(\mathcal{G},\mathcal{C}_k^{+})
+
w_{\texttt{n}}\,
\widetilde{T}_k^{\texttt{nav}}(\mathbf{p}_0^{\xi},\mathcal{C}_k^{+})
\big),
\end{equation}
where $\mathcal{C}_k^{+}$ is the set of subsequent meeting events to
be executed as implied by the local plans $\{\Gamma_i\}$. Here
$\widetilde{T}_k^{\texttt{trs}}(\mathcal{G},\mathcal{C}_k^{+})$ and
$\widetilde{T}_k^{\texttt{nav}}(\mathbf{p}_0^{\xi},\mathcal{C}_k^{+})$
are fast predictors of the transition completion time and bottleneck
arrival time, respectively, conditioned on the confirmed meeting
events $\mathcal{C}_k^{+}$. This heuristic prioritizes candidates
that are likely to admit early feasible detachment and short
worst-case travel, while avoiding repeated evaluation of exact
durations during the search. The overall realization procedure is
summarized in Alg.~\ref{alg:graphmatch}. Its role within MoRoCo is to
return an executable request-consistent transition for a given
candidate family, as formalized by the following lemma and its proof
provided in the Appendix.

\begin{lemma}\label{thm:graphmatch-exec-short}
For any request $\xi$ for team $\mathcal{T}_k$, suppose there exists at
least one candidate $\mathcal{G}_k^{\xi}\in\widehat{\mathcal{G}}_k^{\xi}$
such that \eqref{eq:graphmatch_trs} returns a detach schedule for the
associated robot set $\mathcal{N}_k^{\xi}$. Then
Alg.~\ref{alg:graphmatch} terminates in finite time and returns
$(\mathcal{G}_k^{\texttt{w'}},\,\mathcal{G}_k^{\xi},\,
\widehat{\pi}_k^{\xi})$, together with $\Pi_k^{\xi}$, such that the
resulting transition is executable under the detach mechanism and the
objective in~\eqref{eq:match_objective} is minimized over all
candidates in $\widehat{\mathcal{G}}_k^{\xi}$.
\end{lemma}

\subsubsection{Request-Specific Execution Modes}
\label{subsubsec:request}
Requests are executed either by updating the
embeddings of the wheel backbone or by instantiating a
request-specific embedded subgraph through the detach--execute--rejoin
mechanism. When a request requires a topology transition,
Alg.~\ref{alg:graphmatch} is invoked to instantiate a feasible target
subgraph and return role assignments together with timed detach events
for the involved robots, while the remaining robots preserve the wheel
backbone to maintain latency-bounded intermittent communication.

(I) \textbf{Intra-team communication with bounded latency.}
For intra-team exploration requests $\xi^k_1$, $\xi^k_2$, and
$\xi^k_5$, the joint-wheel backbone $\mathcal{G}^{\texttt{w}}_k$ is
retained and no subgraph instantiation is required. The base request
$\xi^k_1$ is satisfied directly by the backbone, which enforces the
latency-bounded information flow. Requests $\xi^k_2$ and $\xi^k_5$ are
realized by modifying the embeddings in~\eqref{eq:embed-update} under
the same topology. Specifically, $\xi^k_2$ excludes frontiers inside
the prohibited region~$\mathcal{P}^-$ and prioritizes other frontiers
toward $\mathcal{P}^+$ via the embedded weights:
\begin{equation}\label{eq:frontier_weight}
J(f) \triangleq \sum_{p\in \mathcal{P}^+}\Vert x(f)-p\Vert, \quad
\forall f\in \mathcal{F}_k,
\end{equation}
such that larger $J(f)$ is deprioritized. Request $\xi^k_5$ appends an
immediate return opportunity for timely delivery of $D_i$, updating the
relevant time-stamps in the embeddings without changing
$\mathcal{G}^{\texttt{w}}_k$ for the remaining period.

(II) \textbf{Inter-team exchange with bounded latency.}
For the inter-team request $\xi^{k_1k_2}_7$, neighboring teams
$\mathcal{T}_{k_1}$ and $\mathcal{T}_{k_2}$ periodically exchange
compressed mission data under the inter-team bound $T_{\texttt{c}}$.
This request is realized as a singleton messenger subgraph, where a
messenger robot detaches from its backbone, reaches an external meeting
event $\mathbf{c}_{\texttt{e}}=(p_{\texttt{e}},t_{\texttt{e}})$,
exchanges data, and rejoins. Feasibility with respect to the intra-team
meeting sequence is enforced by selecting a messenger robot from the
most recent intra-team meeting pair and scheduling the external event
according to:
\begin{equation}\label{eq:interteam_req}
\begin{aligned}
  & t_{ij}^{+} + T_{\ell}^{\texttt{nav}} \ge t_{\texttt{e}},\;
  \forall \ell\in\{i,j\}, \\
  & p_{\texttt{e}}^{+}=\mathop{\mathbf{argmin}}\limits_{p \in
    \boldsymbol{\tau}_{k_1k_2}}
    \left\{\mathop{\mathbf{max}}\limits_{\ell=k_1,k_2}
    \Big\{T_{\texttt{m}_\ell}^{\texttt{nav}}
    (p_{\texttt{h}_\ell},\,p)\Big\}\right\},
\end{aligned}
\end{equation}
where $\mathbf{c}_{ij}^{+}=(p_{ij}^{+},t_{ij}^{+})$ is the next
intra-team meeting event and $\boldsymbol{\tau}_{k_1k_2}$ is the
shortest path between operator locations $p_{\texttt{h}_{k_1}}$ and
$p_{\texttt{h}_{k_2}}$. The inter-team meeting time is advanced by
$t_{\texttt{e}}^{+}\triangleq t_{\texttt{e}}+T_{\texttt{c}}^{k_1k_2}$,
after which the messenger rejoins the backbone.

(III) \textbf{Operator relocation and group migration.}
Request $\xi^k_3$ relocates the operator $\texttt{h}_k$ to a target
position $p_{\texttt{h}_k}^+$ while preserving latency feasibility. If
$p_{\texttt{h}_k}^+$ satisfies:
\begin{equation}\label{eq:base-location}
T_i^{\texttt{nav}}(p^+_{\texttt{h}_k},\, p^{\ell_i}_i)\leq
T_i^{\texttt{nav}}(p_{\texttt{h}_k},\, p^{\ell_i}_i),\,
\forall i\in \mathcal{N}_k,\, \ell_i\leq L_i;
\end{equation}
then the embeddings are updated in place and the backbone topology
remains unchanged. Otherwise, the relocation is realized as a
connectivity-preserving migration. In this case, \texttt{GraphMatch}
instantiates an embedded star subgraph centered at $\texttt{h}_k$ and
assigns robots to relative placements that maintain connectivity during
motion, after which the wheel backbone is restored for continued
exploration.

(IV) \textbf{Consistent link via a communication-ensured chain.}
Requests $\xi^k_4$ and $\xi^k_6$ require a reliable and consistent
chain between the operator $\texttt{h}_k$ and a designated robot at
$p_i^\star$, as required in Def.~\ref{def:chain}. Accordingly, MoRoCo
realizes these requests through a communication-ensured chain (CEC) by
selecting anchor points along the shortest path from $p_{\texttt{h}_k}$
to $p_i^\star$ such that the hop-by-hop link quality satisfies:
\begin{equation}\label{eq:cec}
\mathbf{p}_k^{\xi}\triangleq p_0p_1\cdots p_C,\quad
\texttt{Qual}(p_c,p_{c+1},\mathcal{M}_{\texttt{h}_k})
\ge \underline{\delta}^{\texttt{c}};
\end{equation}
allocating robots to these anchor points yields an embedded line
subgraph that enforces multi-hop connectivity based on estimated
communication quality. Unassigned robots continue exploration within
the reduced wheel backbone.


\subsection{Online Coordination for Multiple Simultaneous Requests}
\label{subsec:online}
This section presents the online coordination layer of
{MoRoCo} for handling multiple simultaneous requests.
Starting from the single-request instantiation mechanism introduced
above, MoRoCo handles multiple requests online by checking request
admissibility on the current wheel backbone and sequentially
instantiating feasible request subgraphs.

\begin{figure}[t!]
    \centering
    \includegraphics[width=0.95\linewidth]{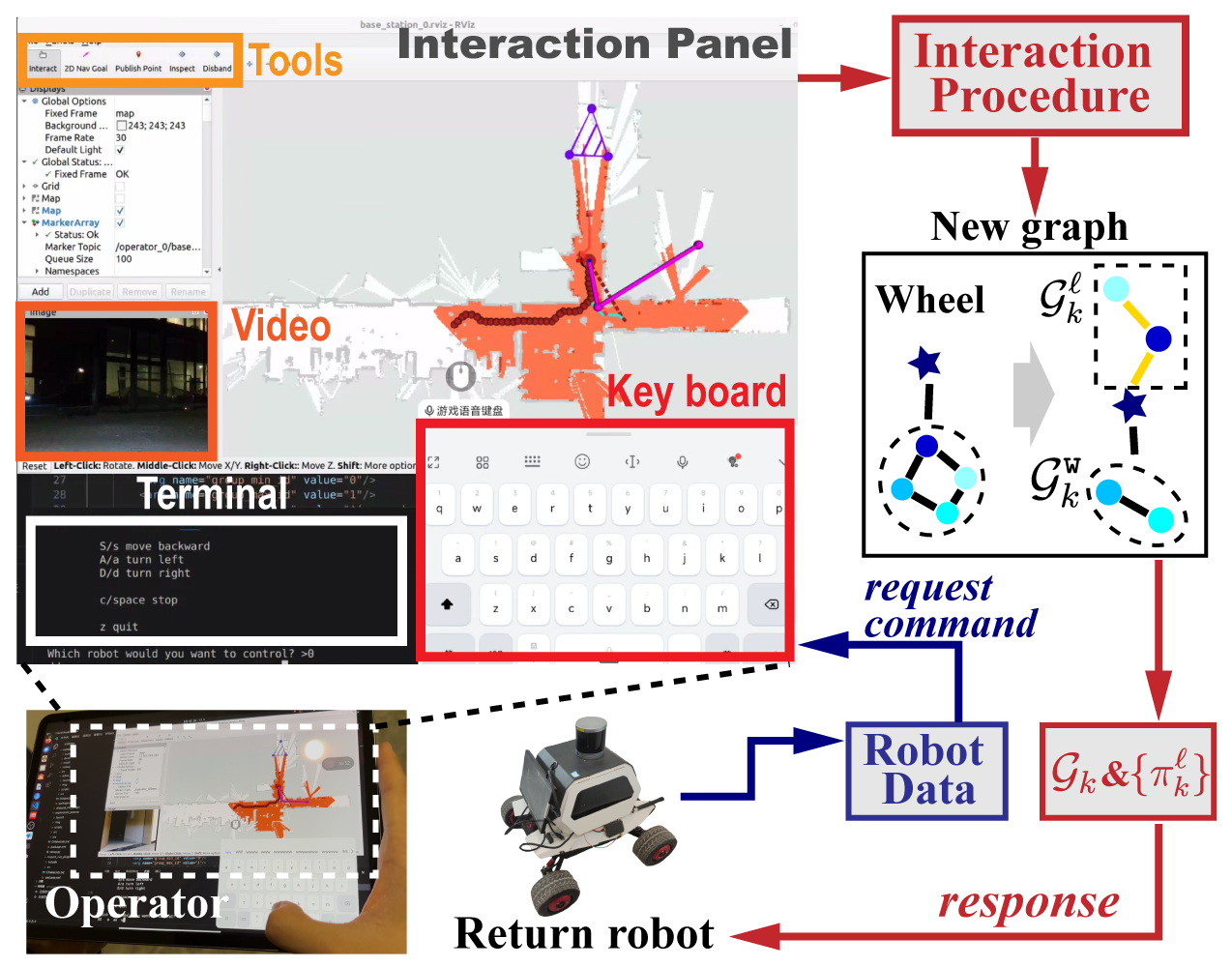}
    \vspace{-2mm}
    \caption{Illustration of the procedure on how the operator can
      interact with its team via the panels within the GUI, and the
      type of data exchanged during the return event including the
      gathered information and human commands.}
    \label{fig:interaction}
\end{figure}

\subsubsection{Compatibility and Sequential Request Instantiation}
\label{subsec:multi_req}
At time~$t$, the embedded graph of team~$\mathcal{T}_k$ is given by:
\begin{equation}\label{eq:decomposition}
  \mathcal{G}_k(t) \triangleq \mathcal{G}^{\texttt{w}}_k \cup
  \bigcup_{j=1}^{L_k}\mathcal{G}^{j}_k,
\end{equation}
where $\mathcal{G}^{\texttt{w}}_k$ is the wheel backbone used for
exploration, and each $\mathcal{G}^{j}_k$ is an instantiated request
subgraph that fulfills request~$\xi^{j}_k$, with $L_k$ denoting the
number of currently active requests.

Suppose robot~$\texttt{r}\in\mathcal{N}^{\texttt{w}}_k$ receives a
bundle of $L>0$ new requests
$\Xi_k\triangleq\{\xi^\ell_k\}_{\ell=1}^{L}$.
Each request~$\xi$ is associated with a candidate family of embedded
target graphs $\widehat{\mathcal{G}}_k^{\xi}$, including
\eqref{eq:frontier_weight} for $\{\xi^k_2\}$,
\eqref{eq:base-location} for $\{\xi^k_3\}$,
\eqref{eq:cec} for $\{\xi^k_4\}$ and $\{\xi^i_6\}$,
and \eqref{eq:interteam_req} for $\{\xi^{k_1k_2}_7\}$.
A subset $\Xi\subseteq\Xi_k$ is said to be \emph{compatible} on
$\mathcal{G}^{\texttt{w}}_k$ if the current wheel backbone can provide
disjoint robot subgroups for instantiating all requests in~$\Xi$, i.e.,
\begin{equation}\label{eq:compat}
  \mathcal{N}^{\xi}_k \cap \mathcal{N}^{\xi'}_k = \emptyset,\
  \forall\,\xi \neq\xi'\in\Xi;\;
  \bigcup_{\xi\in\Xi}\mathcal{N}^{\xi}_k \subseteq
  \mathcal{N}^{\texttt{w}}_k,
\end{equation}
where $\mathcal{N}^{\xi}_k$ denotes the set of robots allocated to
fulfill~$\xi$.
However, the operator-relocation request $\{\xi^k_3\}$, including both
motion within the feasible region and migration, is only compatible
with the complete wheel backbone. If the fleet is already split,
robots in detached subgraphs cannot be guaranteed to
receive the new operator location after migration. Therefore,
$\{\xi^k_3\}$ is deferred until all other active requests are completed
and the complete wheel backbone is restored.

Determining a maximum compatible subset of $\Xi_k$ under
\eqref{eq:compat} is generally combinatorial,
requiring an additional subset-selection problem at each activation event.
MoRoCo does not pursue such a globally optimal selection.
Instead, it adopts a tractable sequential policy: requests are examined according to a priority order,
and a request is retained only if it is admissible on the current wheel backbone
and $\texttt{GraphMatch}(\cdot)$ returns an executable instantiation.
The priority order is determined either by operator specification or by the default FIFO setting.
Specifically, let $\mathcal{G}^{\texttt{w},0}_k$ denote the wheel
graph before processing the new request bundle.
For $\ell=1,\cdots,L$, if request $\xi^\ell_k$ is admissible,
the remaining wheel graph $\mathcal{G}^{\texttt{w},\ell-1}_k$ is split by
instantiating a target graph from
$\widehat{\mathcal{G}}_k^{\xi^\ell_k}$ via
Alg.~\ref{alg:graphmatch}, i.e.,
\begin{equation}\label{eq:combinatorial}
  \mathcal{G}^{\texttt{w},\ell}_k,\ {\mathcal{G}}^{\ell}_k,\
  \widehat{\pi}^{\ell}_k,\ \Pi^{\ell}_k
  \leftarrow
  \texttt{GraphMatch}\!\left(\mathcal{G}^{\texttt{w},\ell-1}_k,\,
  \widehat{\mathcal{G}}^{\ell}_k,\, \xi^{\ell}_k\right),
\end{equation}
where $\widehat{\mathcal{G}}^{\ell}_k$ denotes the target-graph family
for $\xi^\ell_k$, ${\mathcal{G}}^{\ell}_k$ is the instantiated request
subgraph, and $\widehat{\pi}^{\ell}_k$ is the associated set of detach
tasks, consistent with \eqref{eq:single_graphmatch}.
If no executable instantiation exists for the current request on the
remaining wheel backbone, that request is deferred and the procedure
moves on to the next one in the priority order.
After all newly arrived requests have been processed, the embedded graph
$\mathcal{G}_k(t)$ in~\eqref{eq:decomposition} is augmented by the
newly instantiated subgraphs that were found executable.

\begin{algorithm}[t]
  \caption{Overall online execution of the MoRoCo framework $\texttt{MoRoco}(\cdot)$}
  \label{alg:moroco}
  \LinesNumbered
  \SetKwInOut{Input}{Input}
  \SetKwInOut{Output}{Output}
  \Input{$\{\mathcal{G}^{\texttt{w}}_{k,0}\}$, $\{\Gamma_i\}$.}
  \Output{Local plans~$\{\Gamma_i\}$ and evolving graphs
  $\{\mathcal{G}_k\}$.}
  \While{not terminated and all teams~$\{\mathcal{T}_k\}$ in parallel}{
    \tcc{\textbf{Within the wheel graph}}
    Robot~$i\in\mathcal{N}^{\texttt{w}}_k$ executes~$\Gamma_i$ in
    parallel\;
    \uIf{meeting event $\mathbf{c}_{ij}$ occurs}{
      $\Gamma_i,\Gamma_j,\mathbf{c}^{+}_{ij}
      \leftarrow \texttt{PairCoord}(\Gamma_i,\,\Gamma_j,\,\chi_{ij})$\;
      Update embeddings of $\mathcal{G}^{\texttt{w}}_k$\;
    }

    \tcc{\textbf{Request instantiation}}
    \uIf{request bundle $\Xi_k$ received}{
      Order requests in $\Xi_k$ by priority\;
      \ForEach{request $\xi^\ell_k \in \Xi_k$}{
        \uIf{$\xi^\ell_k$ admissible on $\mathcal{G}^{\texttt{w}}_k$}{
          $\widetilde{\mathcal{G}}^{\texttt{w}}_k,\,
          \widetilde{\mathcal{G}}^\ell_k,\,
          \widetilde{\pi}^\ell_k,\,
          \widetilde{\Pi}^\ell_k
          \leftarrow
          \texttt{GraphMatch}(\mathcal{G}^{\texttt{w}}_k,\,
          \widehat{\mathcal{G}}^\ell_k,\,\xi^\ell_k)$\;
          \uIf{an executable instantiation}{
            $\mathcal{G}^{\texttt{w}}_k \leftarrow \widetilde{\mathcal{G}}^{\texttt{w}}_k$, add $\widetilde{\mathcal{G}}^\ell_k$ to $\mathcal{G}_k$\;
            Store $\widetilde{\pi}^\ell_k$ for detachment\;
          }
        }
      }
      Follow detach schedules and update~$\mathcal{G}_k$\;
    }

    \tcc{\textbf{Request execution}}
    \ForEach{active request~$\xi^\ell_k$ in parallel}{
      All robots in~$\mathcal{N}^\ell_k$ perform request~$\xi^\ell_k$\;
      \uIf{request $\xi^\ell_k$ is fulfilled}{
        Trigger rejoin procedure and update $\mathcal{G}_k$\;
      }
    }
  }
\end{algorithm}

\subsubsection{Online Interaction and Execution}
\label{subsec:interaction}
As shown in Fig.~\ref{fig:interaction}, each operator interacts with
the fleet through a graphical panel that supports the online requests
listed in Table~\ref{tab:requests}. At time~$t$, a command from
operator~$\texttt{h}_k$ is parsed into a request bundle
$\Xi_k \triangleq \{\xi_k^\ell\}$ with associated parameters and
transmitted to team~$\mathcal{T}_k$ via intermittent communication.
Execution is fully distributed and driven by local pairwise
coordination in Alg.~\ref{alg:coordination}. Each team maintains an
embedded communication graph $\mathcal{G}_k$, initialized as a
joint-wheel backbone
$\mathcal{G}_k=\mathcal{G}^{\texttt{w}}_{k,0}$ to satisfy the
intra-team latency bound $T^k_{\texttt{h}}$ specified by $\xi^k_1$,
while inter-team exchanges under $\{\xi^{k_1k_2}_7\}$ are constrained
by the bound $T_{\texttt{c}}$. During robot meetings, local embeddings
and data are updated and fused through $\texttt{PairCoord}(\cdot)$,
enabling latency-bounded information propagation without centralized
synchronization.

Upon receiving a new bundle $\Xi_k$, the online
coordination layer of {MoRoCo} processes the requests sequentially
under the admissibility condition~\eqref{eq:compat}, and instantiates
executable request subgraphs within the overall runtime summarized in
Alg.~\ref{alg:moroco}.
Consequently, $\mathcal{G}_k$ evolves into
the union of the wheel backbone and the subgraphs required by active
requests, such as a singleton for $\{\xi^{k_1k_2}_7\}$, a line for
$\{\xi^k_4\}$ and $\{\xi^k_6\}$, and a star for $\{\xi^k_3\}$. Robots
detach to execute within the instantiated subgraphs, while the
remaining wheel backbone continues exploration and preserves
latency-bounded intermittent connectivity. After a request is
completed, the corresponding subgraph is removed and the involved
robots rejoin the backbone, restoring $\mathcal{G}_k$ for continued
exploration and subsequent online requests.
The following theorem states the framework-level guarantee provided by
the online execution policy in
Alg.~\ref{alg:moroco}, with the proof provided in the Appendix.

\begin{theorem}
\label{thm:moroco-complete-short}
For any team $\mathcal{T}_k$ and any finite-time arrival sequence of
request bundles, suppose that whenever a request~$\xi_k^\ell$ becomes
admissible under the compatibility condition in~\eqref{eq:compat},
there exists at least one candidate
$\mathcal{G}_k^{\xi^\ell_k}\in\widehat{\mathcal{G}}_k^{\xi^\ell_k}$
such that \eqref{eq:graphmatch_trs} returns a detach schedule for the
associated robots~$\mathcal{N}_k^{\xi^\ell_k}$.
Assume further that every instantiated request subgraph completes its
assigned request and rejoins the wheel backbone in finite time.
Then Alg.~\ref{alg:moroco} satisfies the following properties:
\begin{enumerate}
    \item[(I)] after each request activation or completion, the
    embedded graph $\mathcal{G}_k(t)$ remains an executable decomposition
    of the form~\eqref{eq:decomposition}, consisting of a remaining wheel
    backbone and finitely many subgraphs for active requests;
    \item[(II)] every request that becomes admissible and is instantiated by
    Alg.~\ref{alg:moroco} is serviced in finite time;
    \item[(III)] there exists a finite time $t_f$ such that
    $\mathcal{G}_k(t)=\mathcal{G}_k^{\texttt{w}}$ for all $t\ge t_f$,
    i.e., all request subgraphs have been removed and the complete wheel
    backbone is restored.
\end{enumerate}
\end{theorem}


\subsection{Discussion}\label{subsec:discuss}

\begin{figure}[t!]
  \centering
  \includegraphics[width=0.95\linewidth]{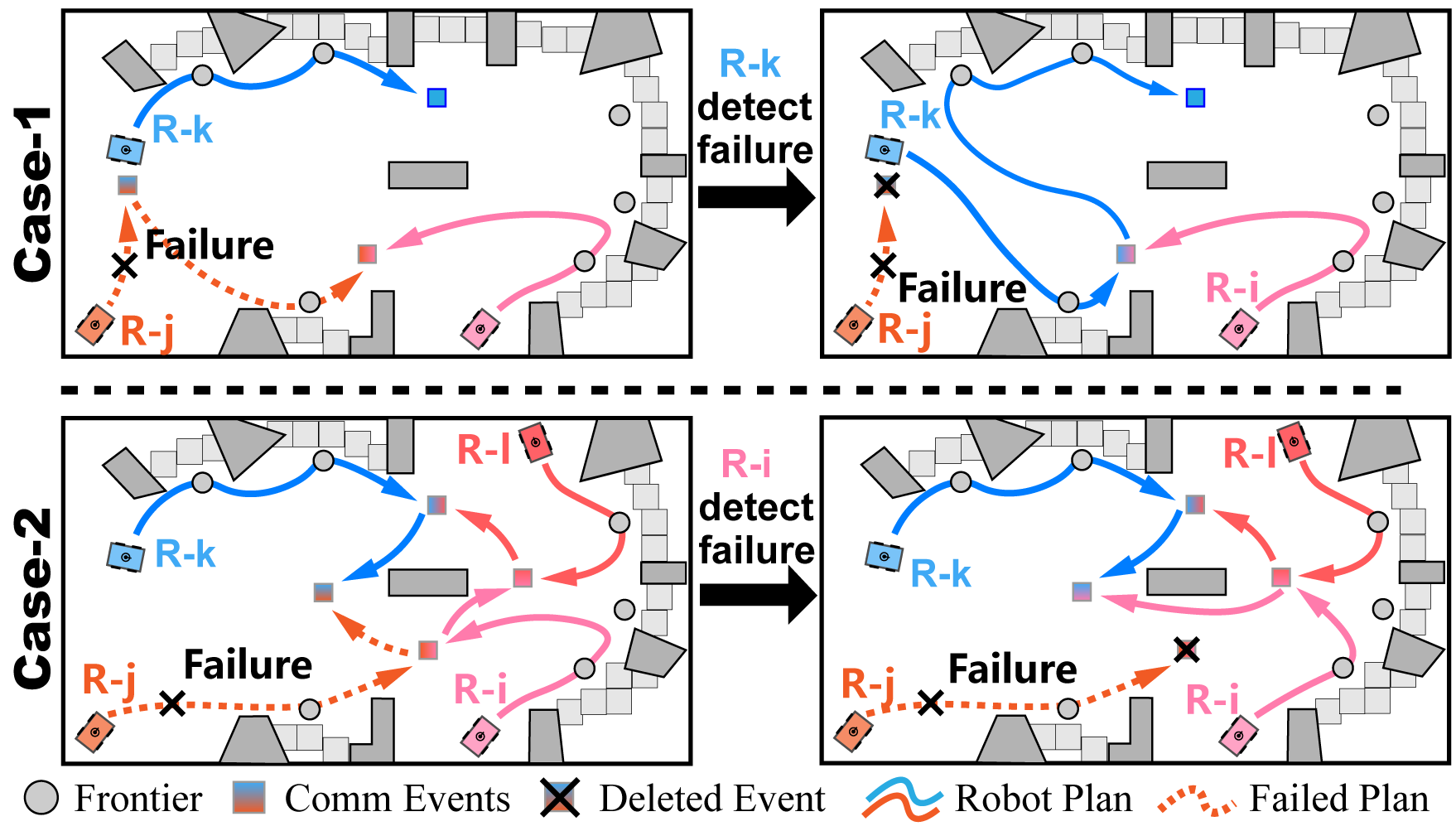}
  \vspace{-2mm}
  \caption{The proposed failure detection and recovery scheme.
  \textbf{Top}: if robot $j$ fails before meeting its predecessor $k$,
  robot $k$ bypasses $j$ after waiting $T^-_{\texttt{max}}$;
  \textbf{Bottom}: if $j$ fails before meeting its successor $i$,
  robot $i$ triggers recovery after waiting $T^+_{\texttt{max}}$, and
  the joint wheel topology is restored after two meetings.}
  \label{fig:failure-cases}
\end{figure}

\subsubsection{Complexity Analysis}
The computational cost of {MoRoCo} is event-driven. At robot meeting
events, the dominant routine is $\texttt{PairCoord}(\cdot)$ in
Alg.~\ref{alg:coordination}, whose cost is dominated by a TSP-path
computation over the merged frontier set
$\mathcal{F}_{ij}$ with complexity $\mathcal{O}(|\mathcal{F}_{ij}|^3)$
and one shortest-path query on the explored map with complexity
$\mathcal{O}(N_m \log N_m)$, yielding
$\mathcal{O}(|\mathcal{F}_{ij}|^3(|\mathcal{F}_{ij}| + N_m \log N_m))$
overall, where $\mathcal{F}_{ij}$ is the merged frontier set,
and $N_m$ is the number of map nodes.
When topology instantiation is required, MoRoCo invokes
$\texttt{GraphMatch}(\cdot)$, whose dominant step is the bottleneck
assignment with complexity $\mathcal{O}(N_k^3)$, so the overall cost is
$\mathcal{O}(N_cN_k^3)$, where $N_c$ is the number of candidate graphs
in $\widehat{\mathcal{G}}_k^{\xi}$ and $N_k$ is the team size. Hence,
for one request bundle
$\Xi_k=\{\xi_k^\ell\}_{\ell=1}^{L}$, the online framework in
Alg.~\ref{alg:moroco} incurs the same cost as
$\texttt{PairCoord}(\cdot)$ at meeting events, and otherwise processes
admissible requests sequentially with complexity
$
\mathcal{O}\!\left(\sum_{\ell=1}^{L_a}
N_{c,\ell}\,(N_k^{\texttt{w},\ell-1})^3\right),
$
where $L_a\leq L$ is the number of admissible requests;
$N_{c,\ell}$ is the number of candidate graphs for~$\xi_k^\ell$; and
$N_k^{\texttt{w},\ell-1}$ is the size of the remaining wheel backbone
before the $\ell$-th instantiation. Since
$N_k^{\texttt{w},\ell-1}\leq N_k$, this yields the worst-case bound
$\mathcal{O}(L_a N_{c,\max} N_k^3)$, where
$N_{c,\max}=\textbf{max}_\ell\{ N_{c,\ell}\}$. Hence, the online cost
of MoRoCo scales linearly with the number of admissible requests in a
bundle and cubically with the team size in the worst case.

\subsubsection{Generalization}
Several nontrivial extensions arise naturally from the MoRoCo
framework, as validated in the sequel.
\emph{(I) Heterogeneous fleets.} Heterogeneity is handled by
incorporating mobility and resource feasibility into navigation-time estimation,
restricting goal selections to reachable options:
\begin{equation}\label{eq:new-nav}
T_i(p_s,\,p_g)=T_{i}^{\texttt{Nav}}(p_s,\,p_g,M_i),\,
\text{ if }\, R_i(p_s,\,p_g) < r_s,
\end{equation}
which prevents unreachable assignments while enabling capability-aware
request realization and task allocation;
\emph{(II) Failure detection and recovery.}  As illustrated in Fig.~\ref{fig:failure-cases},
Failures are detected locally via bounded waiting times $T^-_{\texttt{max}}$ and $T^+_{\texttt{max}}$
for predecessor- and successor-side checks, respectively;
upon detection, the failed robot is removed and the wheel topology is restored through a small number of
subsequent meetings, with $T^+_{\texttt{max}}>T^-_{\texttt{max}}$ ensuring conservative successor-side detection.
\emph{(III) Spontaneous meeting events.} Unexpected encounters are accommodated by fusing local data while
preserving the current topology; only inter-team meetings between neighboring teams trigger an update of the
external-event schedule, whereas non-neighboring teams exchange data without additional coordination.

\begin{figure}[t!]
  \centering
  \includegraphics[width=0.95\linewidth]{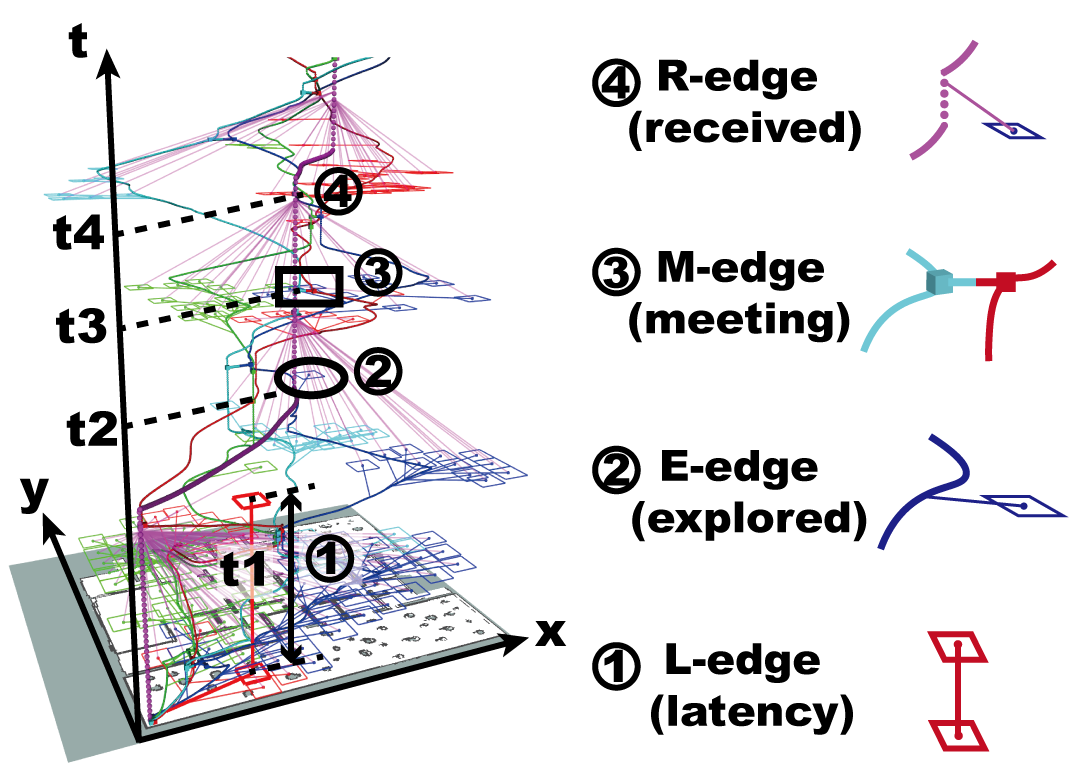}
  \vspace{-2mm}\caption{\textbf{Left}: the proposed data structure
    as the spatial-temporal tree of data flow (\emph{SDF-Tree}),
    to encapsulate information gathered during exploration and data-propagation;
  \textbf{Right}: the spatial-temporal trajectories of each robot and the operator,
  and four types of edges in the tree.
  Note that $t_1$ is the maximum latency, $t_2$ is is the time of exploration for the grid area,
  $t_3$ is meeting time, and $t_4$ the time that this area is first received by operator.}
  \label{fig:sdf_tree}
\end{figure}

\subsubsection{Limitations}\label{subsec:limitation}
Several limitations motivate future extensions.
First, accurate map merging is essential for intermittent coordination, yet high merge reliability may require increased overlap when initial poses are uncertain, which conflicts with the objective of minimizing redundant exploration; this trade-off is not explicitly modeled.
Then, the current system largely ignores semantic cues, which could improve region prioritization, automate request generation, and enhance map merging through semantic anchors, and is critical for missions such as search and rescue.
For similar reasons, interaction is currently limited to touch or typing,
which can be burdensome for non-expert users; multimodal interfaces, LLM-based command translation, and reusable request templates may improve usability and task coverage.
Lastly, exploration efficiency remains sensitive to map topology; predictive mapping and learned heuristics could further improve frontier selection and communication-event placement~\cite{calzolari2024reinforcement, sygkounas2022multi}, accelerating both exploration and contingent task fulfillment.


\begin{figure*}[t!]
  \centering
  \includegraphics[width=0.95\linewidth]{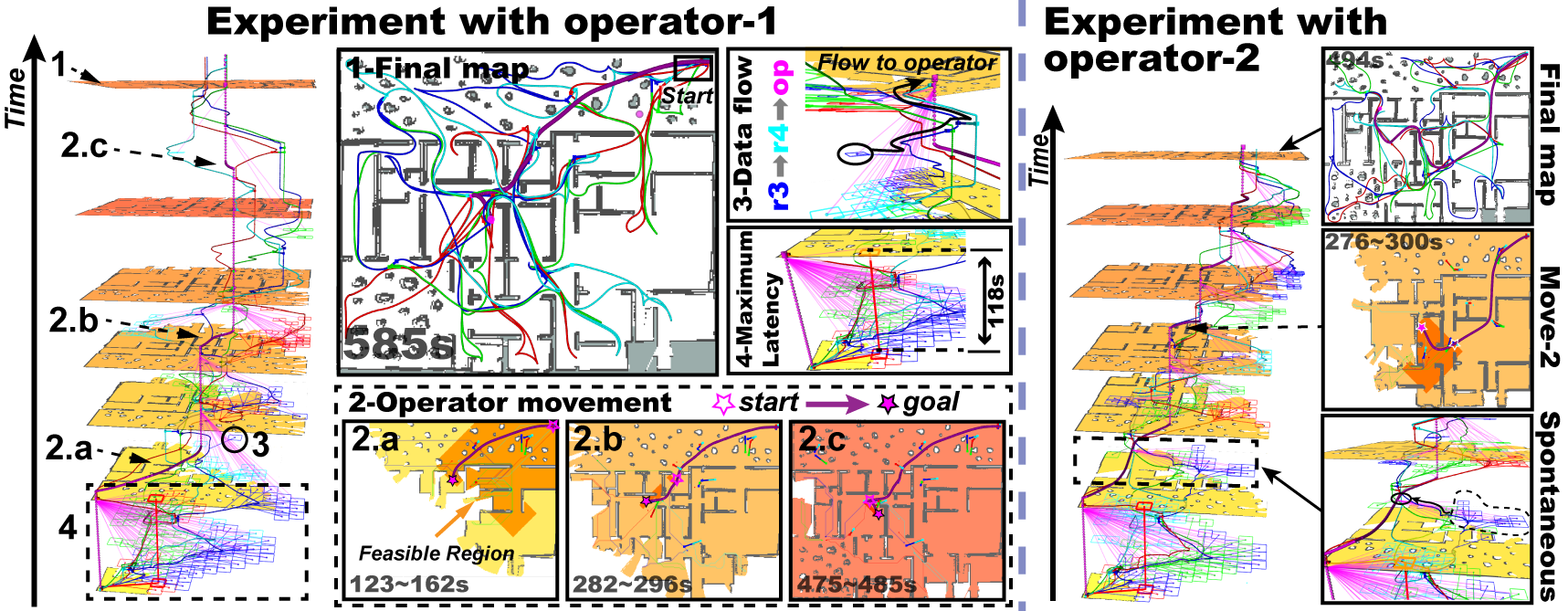}
  \vspace{-2mm}\caption{Comparisons of the same team within Scenario-1
    under two different behaviors of operators
    (\textbf{Left} and \textbf{Right}),
    visualized by the SDF-tree in addition to the local map of the operator
    at several return events.
    Notice the differences in the operator movement (thick lines)
    and the associated feasible region.
  }
  \label{fig:exp_explore_1}
\end{figure*}

\section{Numerical Experiments} \label{sec:experiments}
For validation,
extensive numerical simulations and hardware experiments
are presented in this section.
All components are implemented in \texttt{Python} and \texttt{ROS} and tested on a
laptop with an Intel Core i7-1280P CPU.
Simulation and experiment videos can be found in the supplementary files.

\subsection{System Description}\label{subsec:description}
\subsubsection{System Setup}
Each robot $i\in\mathcal{N}$ builds a local SLAM map $M_i$ using a 2D occupancy grid representation~\cite{moravec1985high}.
All simulations are conducted in \texttt{ROS-Stage} with \texttt{gmapping}, where robots use differential-drive navigation with the default ROS stack,
with a maximum linear/angular speed of $1.0\,\mathrm{m/s}$ and $1.0\,\mathrm{rad/s}$, and a sensing range of $15\,\mathrm{m}$.
Communication is intermittent and only enabled when the measured link quality exceeds $50\,\mathrm{dB}$;
the ground-truth quality follows an analytical propagation model~\cite{marcotte2020optimizing} that is unknown to the planner,
while a learned estimator predicts it from sampled measurements.
Local maps are merged upon connectivity using \texttt{multirobot\_map\_merge}~\cite{horner2016map}.
Operators interact with the fleet through a customized \texttt{Rviz} GUI as in Figure~\ref{fig:interaction} to issue online requests
and update the latency thresholds $T^k_\texttt{h}$ and $T_\texttt{c}$ at return events.

\begin{table}[t!]
  \centering
  \vspace{-2mm}\caption{Evaluation of Key Parameters under Different Team Configuration.}\label{table:exp1-data}
  \vspace{-0.01in}

  \footnotesize
  \setlength{\tabcolsep}{0.06cm}
  \renewcommand{\arraystretch}{0.82}

  \begin{threeparttable}
    \resizebox{0.95\linewidth}{!}{
    \begin{tabular}{l c c c c c}
      \toprule[1pt]
      \multirow{2}{*}{\textbf{Metrics}} &
      \multirow{2}{*}{\textbf{op-1+4ro}} &
      \multirow{2}{*}{\textbf{op-2+4ro}} &
      \multicolumn{3}{c}{\textbf{op-1+2ro \& op-2+2ro}} \\
      \cmidrule(lr){4-6}
      & & & {$T_{\texttt{c}}=240$} & {$T_{\texttt{c}}=300$} & {$T_{\texttt{c}}=360$} \\
      \midrule
      $T_\texttt{op1}$ [s] \tnote{1}        & 585            & -              & 553            & \textbf{413} & 466 \\
      $T_\texttt{op2}$ [s] \tnote{1}        & -              & 494            & 545            & 494          & \textbf{452} \\
      $L_\texttt{op1}$ [m] \tnote{2}        & \textbf{50.0}  & -              & 52.8           & 52.7         & 50.7 \\
      $L_\texttt{op2}$ [m] \tnote{2}        & -              & 79.7           & 41.1           & 41.6         & \textbf{40.6} \\
      $\delta_{\texttt{in}}$ [s] \tnote{3}  & 118            & 117            & 93             & \textbf{85}  & 104 \\
      $\delta_{\texttt{ext}}$ [s] \tnote{4} & -              & -              & \textbf{276}   & 335          & 371 \\
      $M_{\texttt{o}}$ [\%] \tnote{5}       & -              & -              & 71.9           & \textbf{40.7}& 44.5 \\
      \bottomrule[1pt]
    \end{tabular}
    }

    \begin{tablenotes}[para]
      \footnotesize
      \item[1] Time when operator 1 or 2 receives the complete map.
      \item[2] Distance travelled by the operator 1 or 2.
      \item[3] Maximum intra-team latency.
      \item[4] Maximum inter-team latency.
      \item[5] Percentage of overlapping area explored by both teams.
    \end{tablenotes}
  \end{threeparttable}

  \vspace{-2mm}
\end{table}

\subsubsection{Spatial-temporal Tree of Data Flow}
For concise visualization, we use a spatial-temporal tree of data flow
(\textbf{SDF-tree}) in the $x$-$y$-$t$ space, as shown in
Fig.~\ref{fig:sdf_tree}. Each explored grid cell is represented by a
leaf node attached to the corresponding robot trajectory through an
\emph{E-edge}; once that information reaches the operator, the same
node is linked to the operator trajectory by an \emph{R-edge}, whose
time-axis projection gives the latency. Communication events are shown
by \emph{M-edges}, while an \emph{L-edge} marks the maximum latency
within or between teams. In this way, the SDF-tree jointly records
where a region is explored, when it is explored, and when it becomes
available to the operator, so both exploration progress and information
propagation can be read from the same structure. 

\begin{figure}[t!]
  \centering
  \includegraphics[width=0.95\linewidth]{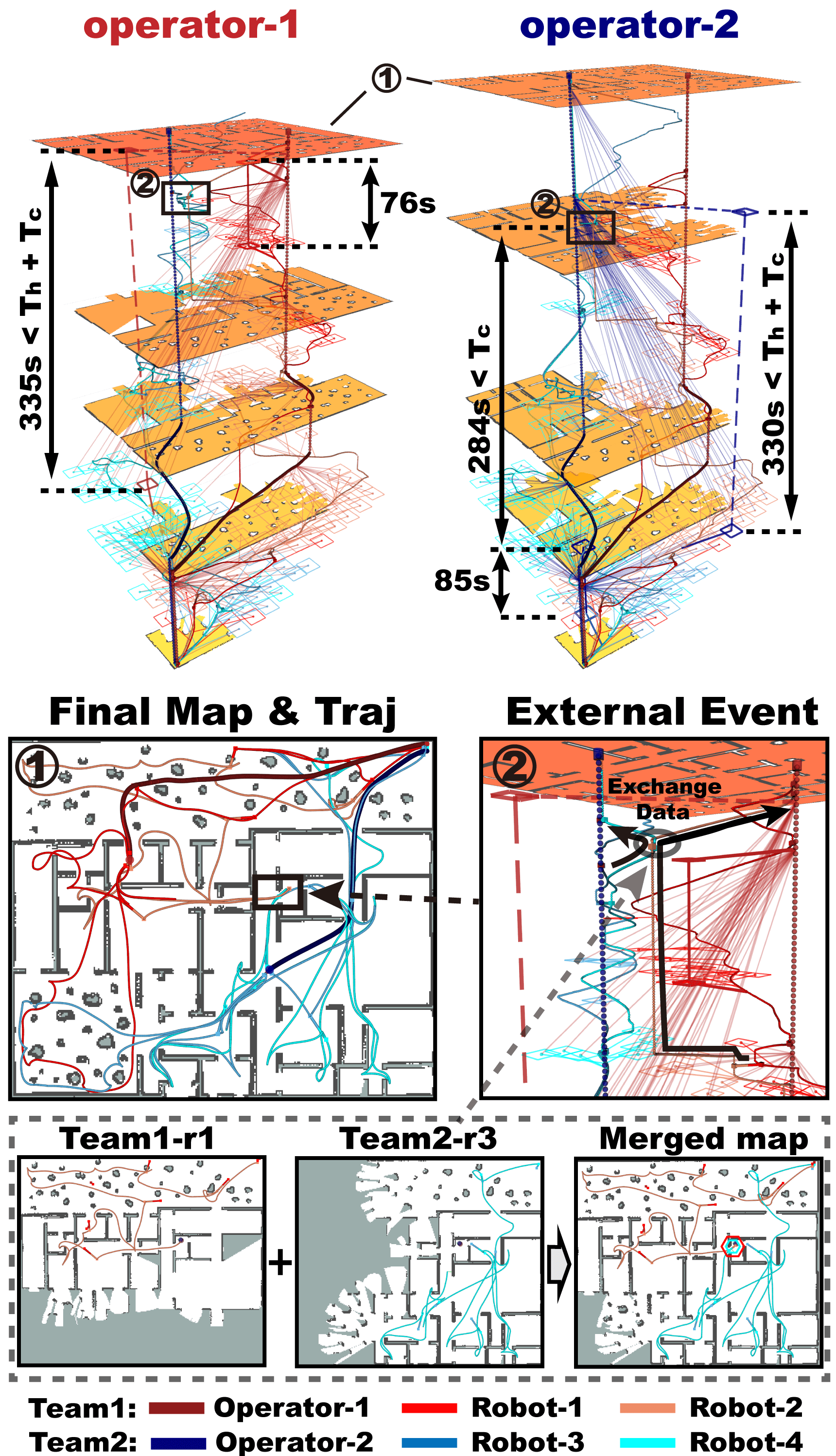}
  \vspace{-2mm}\caption{Simulation results of two teams deployed in Scenario-1,
    where the latency constraints~$T_\texttt{c}=300s$.
  \textbf{Top}: the SDF-Trees of Team-1 and Team-2.
  Both contain all the robot trajectories, M-edges and E-edges,
  while the R-edges, L-edges and maps are linked to the corresponding operator.
  The L-edges with (without) dashed line represent maximum inter (intra)-team latency;
  \textbf{Middle}: the final map with trajectories of all robots and operators
  (Team-1 in red and Team-2 in blue) and one snapshot of an external
  communication event;
  \textbf{Bottom}: the local maps of Robot-1 in Team-1
  and Robot-3 in Team-2 are merged during the external event.}
  \label{fig:exp_explore_2}
\end{figure}

\begin{figure}[t!]
  \centering
  \includegraphics[width=0.95\linewidth]{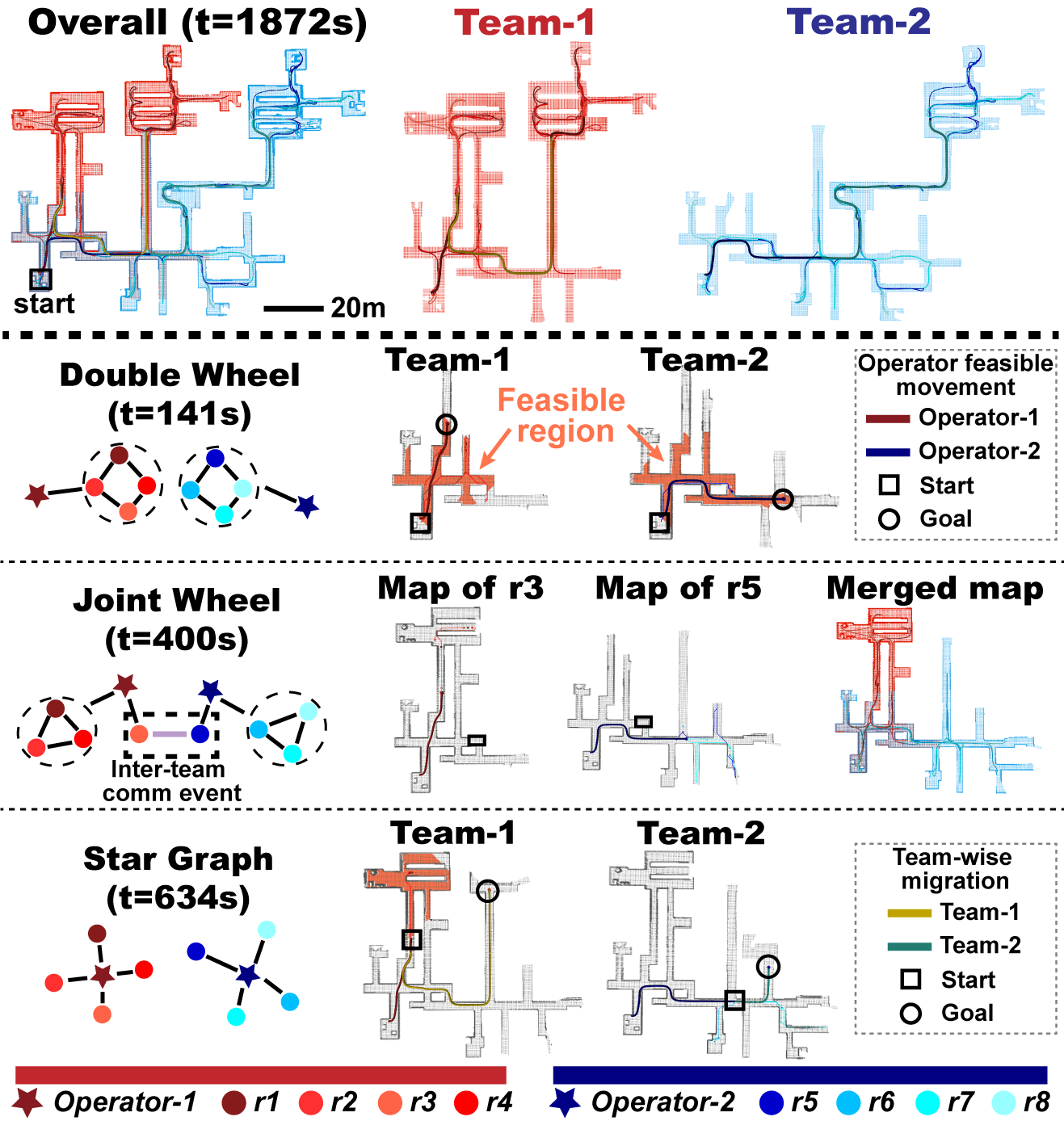}
  \vspace{-2mm}\caption{Simulation results of two teams deployed in a large subterranean environment
    for Scenario-2,
  where the latency $T_\texttt{h}=120s$ and $T_\texttt{c}=300s$.
  \textbf{Top:} the explored area of each team within the final map for Team-1 (red)
  and Team-2 (blue),
  including the trajectories for robots and operators;
  \textbf{Bottom:} the evolution of embedded graphs for both teams over time.
  }\label{fig:exp-indoor}
\end{figure}

\begin{figure}[t!]
  \centering
  \includegraphics[width=0.95\linewidth]{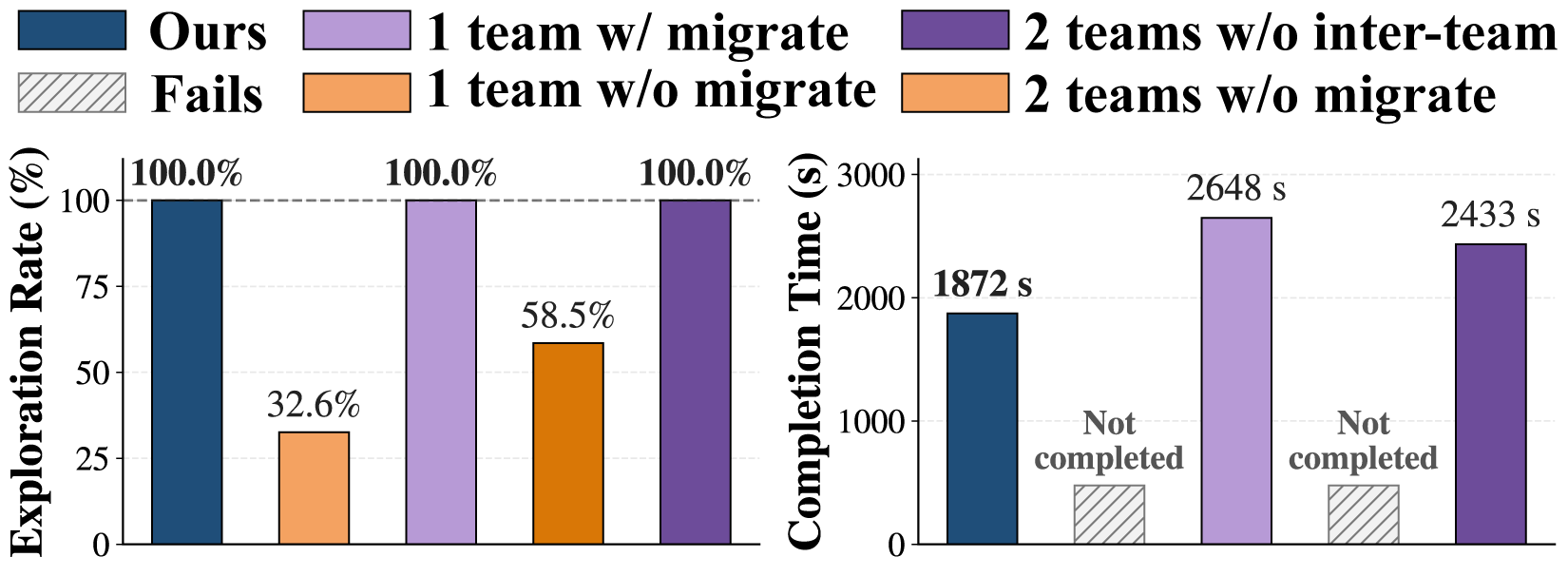}
  \vspace{-2mm}\caption{
    Comparison of exploration rate (\textbf{Left}) and completion time (\textbf{Right})
    for the fleet of single and two teams with and without ``migrate'' requests.
}\label{fig:exp-indoor-compare}
\end{figure}

\subsection{Operator-in-the-loop Collaborative Exploration}\label{subsec:explore}
To begin with, the requests $\xi_1$, $\xi_3$, and $\xi_7$
are investigated within two different scenarios
for a single team and two teams,
to validate the human-robot collaborative exploration framework
under latency constraints.

\subsubsection{Simulated Scenarios}
The following two scenarios of varying complexity and features
are considered:

\textbf{Scenario-1}:
as shown in Figure \ref{fig:exp_explore_1}, the task is to explore a large building complex
surrounded by forest area, which has a size of $60m\times 50m$.
The operator is allowed to move only within the feasible region,
and does not send any other request.
The intra-team latency bound $T^k_\texttt{h}$ is set to $120s$ for each team,
and inter-team latency $T_\texttt{c}$ is set to $300s$
for each pair of neighboring teams.
A single team of four robots and one operator is tested first,
and then extended to two teams with two robots per team.

\textbf{Scenario-2}:
as shown in Figure~\ref{fig:exp_explore_2}, a $120m\times 90m$ large-scale subterranean environment,
similar to the DARPA SubT setting in~\cite{saboia2022achord}, is considered.
It features lobby rooms connected by long, narrow passages with frequent sharp turns,
and multiple branching structures with dead ends, which significantly degrades connectivity and poses challenges for communication-constrained exploration.
Therefore, numerical validations are conducted under different team configurations, communication strategies, and operator motions.
Specifically, two teams with $4$ robots and $1$ operator are tested and compared against a single team with $8$ robots and $1$ operator under different operator commands.

\begin{figure*}[t!]
  \centering
  \includegraphics[width=0.95\linewidth]{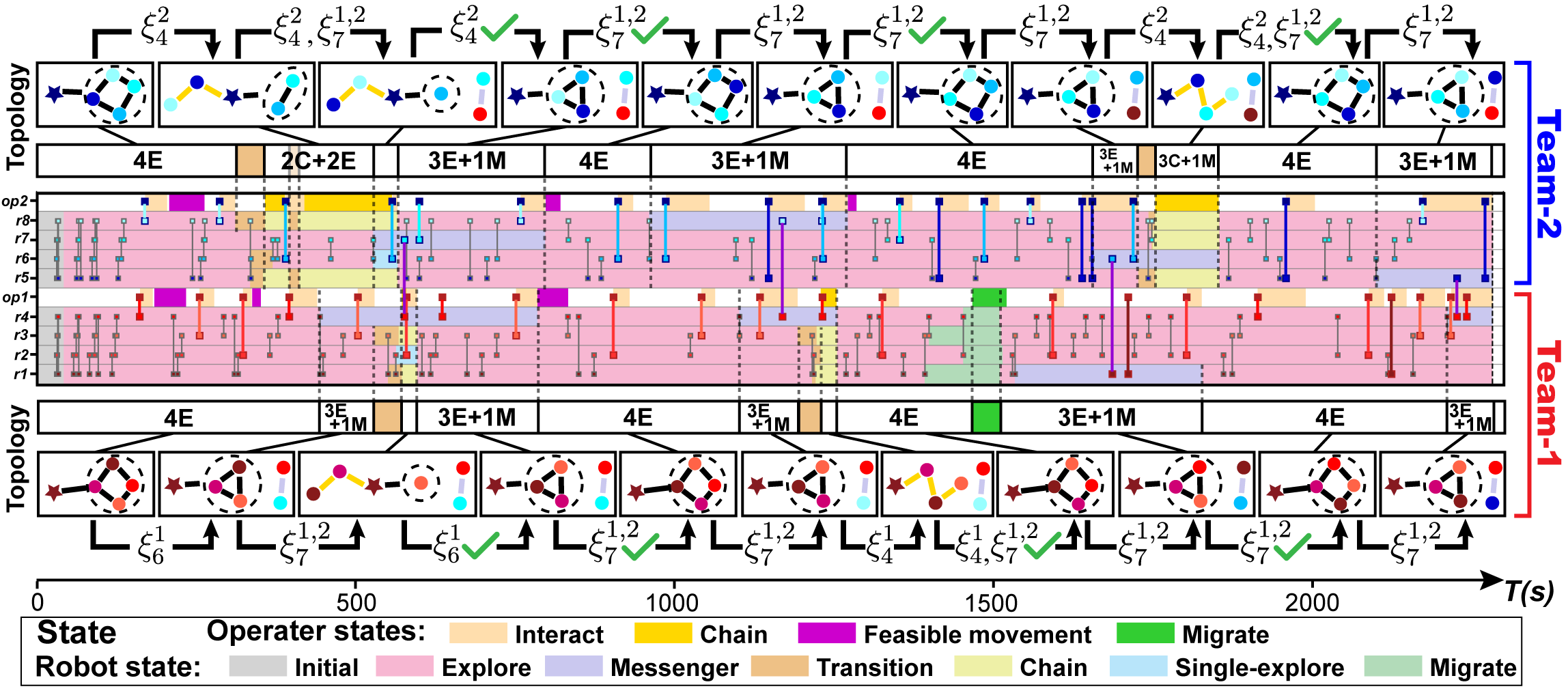}
  \vspace{-2mm}\caption{Simulation results of two teams (2 operators and 8 robots)
    fulfilling 5 contingent requests via graph transitions,
    while exploring a $90m\times 70m$ office environment.
    \textbf{Top} and \textbf{Bottom}:
    the complete evolution of the operation state and
    the associated communication graphs of Team-2 and Team-1
    after each request is activated and fulfilled, respectively;
    \textbf{Middle}:
    the intermittent intra-team and inter-team
    communication events between the corresponding robots or operators
    (dumbbell-shaped vertical lines).
    The communication graph is specified by the number of robots
    and their operation mode,
    where `E', `C' and `M' represent the exploration, chain,
    and messenger state respectively,
    i.e., ``2E+2C'' means 2 robots in the exploration state and 2 robots
    in the chain state.
  }\label{fig:exp-trans}
\end{figure*}

\subsubsection{Single-team in Scenario-1 under Distinct Operator Motion Patterns}
To validate the intra-team latency constraint, the proposed coordination scheme is evaluated for a single team in Scenario-1 with two operators exhibiting distinct motion behaviors (Fig.~\ref{fig:exp_explore_1}).
In both cases, the maximum latency remains below the bound $T^k_\texttt{h}=120s$, as each R-edge is consistently shorter than its corresponding L-edge, and full map coverage is achieved within a few hundred seconds.
For the first operator, the grid cell inducing the largest latency is initially explored by robot~$0$ and delivered at the first return event after $118s$; the operator then moves toward the environment center via three consecutive motions at $123s$, $282s$, and $475s$,
all within the feasible region.
The exploration completes at $585s$, with $23$ inter-robot meeting events and $7$ return events.
For the second operator, moving downward first and then toward the top-left region yields a shorter completion time of $494s$ while keeping the maximum latency at $115s$.
This speedup mainly results from mitigating late-stage long-range transmissions from corner regions and reducing the frequency of planned return events,
as the operator remains closer to planned robot meeting points, leading to more spontaneous returns.
Consequently, robots spend more time exploring rather than relaying data, despite a longer operator travel distance of $79.7m$ compared with $50.0m$ in the first case.

\subsubsection{Two Teams in Scenario-1 under Different~$T_{\texttt{c}}$}
In Scenario-1,  the inter-team coordination is evaluated with two teams initially deployed in close proximity and set $T_{\texttt{c}}=300s$, where the first external exchange is triggered after operator separation and subsequent requests follow~\eqref{eq:interteam_req},
and the results are shown in Fig.~\ref{fig:exp_explore_2}.
As the two operators move in different directions, the teams explore largely complementary regions, and the first external communication allows messengers (robot~$1$ and robot~$3$) to merge their local maps at $382s$, substantially reducing redundant exploration.
Consequently, operator-1 and operator-2 obtain the complete map at $413s$ and $494s$, respectively, with more than $60\%$ of the area explored by only one team; moreover, operator-2's travel distance drops to around $40m$ compared with $80m$ in the single-team case.
Throughout the run, latency guarantees are preserved, with maximum intra-team latencies of $76s$ and $85s$ and maximum inter-team latency of $284s$ (below $T_{\texttt{c}}=300s$), indicating reliable inter-team information delivery under the combined inter-/intra-team bounds.
As shown in Table~\ref{table:exp1-data}, further experiments varying $T_{\texttt{c}}$ reveal a consistent efficiency gain over the single-team baseline but also a clear trade-off: overly small $T_{\texttt{c}}$ induces premature exchanges that fuse maps before each team finishes exploiting nearby frontiers,
increasing spatial overlap and coordination overhead, whereas overly large $T_{\texttt{c}}$ delays information fusion and allows redundancy to re-emerge.
Therefore, $T_{\texttt{c}}$ plays a key role in balancing early synchronization cost against late sharing inefficiency.

\subsubsection{Two Teams in Scenario-2 with Different Operator Requests}
Scenario-2 evaluates inter-team latency constraints in a large subterranean environment with two teams, each consisting of one operator and four robots, and $T_{\texttt{c}}=300s$ in Figs.~\ref{fig:exp-indoor} and~\ref{fig:exp-indoor-compare}.
With \texttt{migration} and inter-team coordination enabled, both operators obtain the complete map at $1872s$ while achieving complementary branch coverage: team-1 and team-2 explore $59.2\%$ and $67.4\%$ of the environment with only $26.6\%$ overlap, and operators travel $139.4m$ and $145.0m$, respectively. These results highlight two factors for scalability in this branched topology. First, \texttt{migration} is necessary when feasible-region motion alone cannot reach new frontiers; without it, exploration stalls early and the final exploration rate drops to $32.6\%$ for one team of eight robots and $58.5\%$ for two teams of four robots. Second, explicit inter-team coordination under latency constraints is critical for efficiency rather than mere completion: removing it still allows completion under \texttt{migration}, but overlap rises from $26.6\%$ to $84.0\%$ and completion slows from $1872s$ to $2433s$. Moreover, although \texttt{migration} enables reliable completion across configurations, a single team of eight robots remains slower and finishes at $2648s$, underscoring the benefit of multi-team parallelism for covering multiple branches simultaneously.

\subsection{Multi-Modal Requests and Topology Adaptation}\label{subsec:exp-requests}
In this part, the proposed framework is evaluated regarding various multi-model online
requests that require the fleet to transit among different communication topologies.

\subsubsection{Simulated Scenario}
Two teams are deployed to explore an office environment of size $90m\times 70m$,
where each team is composed of 4 robots and one operator.
In addition to exploration, \emph{in total 5 tasks} are scattered in the environment,
the completion of which require the proposed CEC chain.
As summarized in Table~\ref{tab:task_summary}, these tasks include the inspection of a target area,
the direct assistance to a robot, and to control a robot remotely to open a door.
During the experiment, each operator can specify any type of requests,
including $\xi_2, \xi_3, \xi_4$ and the migration command;
when a robot observes a task, it can send requests of type $\xi_5$ or $\xi_6$
to the operator in order to fulfill the task.
Initially, both teams are located at the top end of a corridor,
and the exploration starts after the first round of communication.
The latency constraints $T_\texttt{h}$ and $T_\texttt{c}$ are set to~$120s$ and $600s$,
respectively.

\begin{table}[t!]
  \centering
  \vspace{-2mm}\caption{Summary of the Contingent Tasks, along with the
    Associated Online Requests.}
  \label{tab:task_summary}
  \small
  \setlength{\tabcolsep}{2pt}
  \renewcommand\arraystretch{1.2}
  \begin{threeparttable}
  \begin{tabular}{@{}ccccccc@{}}
  \toprule
  \textbf{Task} & \textbf{Type} & \textbf{Request} & \textbf{Robots} & $T_\texttt{tra}$[s]\tnote{1}  & $T_\texttt{sta}$[s]\tnote{2}  & $T_\texttt{dur}$[s]\tnote{3}  \\
  \midrule
  Task-1 & Inspection & $\xi^2_4$ & r5, r8 & 44 & 356 & 39 \\
  Task-2 & Inspection & $\xi^2_4$ & r5, r8 & 16 & 411 & 152 \\
  Task-3 & Assistance & $\xi^1_6$ & r1, r3 & 43 & 572 & 24 \\
  Task-4 & Push door  & $\xi^1_4$ & r1, r2, r3 & 35 & 1231 & 25 \\
  Task-5 & Inspection & $\xi^2_4$ & r5, r7, r8 & 28 & 1755 & 99 \\
  \bottomrule
  \end{tabular}

  \begin{tablenotes}[para]
    \footnotesize
    \item[1] Transition time from the joint wheel to chain graph.
    \item[2] Starting time of the task execution.
    \item[3] Duration of the task.
  \end{tablenotes}
  \end{threeparttable}

  \vspace{2mm}
\end{table}

\subsubsection{Results}
Figure~\ref{fig:scenario}, \ref{fig:exp-trans} and Table~\ref{tab:task_summary} show that
the framework can execute heterogeneous online requests with frequent graph reconfiguration while keeping the overall process efficient.
All five tasks are successfully fulfilled by $1854s$, including four operator-issued requests via $\xi_4$ and one robot-issued request $\xi^{1}_{6}$,
indicating that both human-initiated and robot-initiated tasking are handled without compromising completion.
Crucially, the computational overhead of topology selection is negligible relative to physical execution:
constructing the task-specific target graph set (CEC chain) takes $0.3s$ on average and matching process takes $0.1s$, whereas the task execution itself spans $25s$--$152s$,
with inspection tasks requiring $39s$--$152s$ and assistance or door-opening taking around $25s$.
This separation in timescales confirms that performance is bottlenecked by task actions rather than planning or graph matching.
Despite $11$ graph transitions per team, reconfiguration remains tractable, with wheel-to-chain transitions completed within $34s$--$42s$ for chain lengths $3$ and $2$, respectively,
and with the most efficient case achieving a direct task-to-task transition of only $16s$.
Compared with the average wheel--chain transition time of $38s$,
this highlights that the framework can exploit topological similarity across consecutive tasks to reduce reconfiguration latency.
Moreover, the system maintains bounded inter-team information exchange under repeated reconfiguration,
evidenced by only four external communication events at $574s$, $1167s$, $1683s$, and $2224s$, while both operators still obtain complete maps at $2242s$ and $2290s$.
These results substantiate that topology adaptation scales to frequent online tasking, with planning costs at the sub-second level and transition costs that remain small compared with task durations.

\subsection{Comparisons}\label{subsec:comparisons}
The proposed framework, denoted as \textbf{MOROCO}, is compared against representative baselines in both single-team and multi-team settings.
Since existing methods cannot jointly support collaborative exploration and consistent tasks,
the comparison focuses on collaborative exploration under bounded-latency operator updates.

\subsubsection{Single-team Experiments}
A single team consisting of one operator and four robots is evaluated first in three environments under the same request $\xi^k_1$ with $T_\texttt{h}=160s$, repeating each method over five runs.
As most baselines do not explicitly enforce operator-update latency, they are minimally adapted to satisfy $T_\texttt{h}$ by inserting return events while preserving their original exploration policies.
Baselines include \textbf{CMRE}~\cite{burgard2005coordinated}, \textbf{M-TARE}~\cite{cao2023tare}, \textbf{JSSP}~\cite{Ani2024iros}, and \textbf{MOROCO-NI} that disables online human-robot interaction.
Details of the baselines and the adaptation are described in the Supplementary file.

\begin{figure}[t!]
  \centering
  \includegraphics[width=0.95\linewidth]{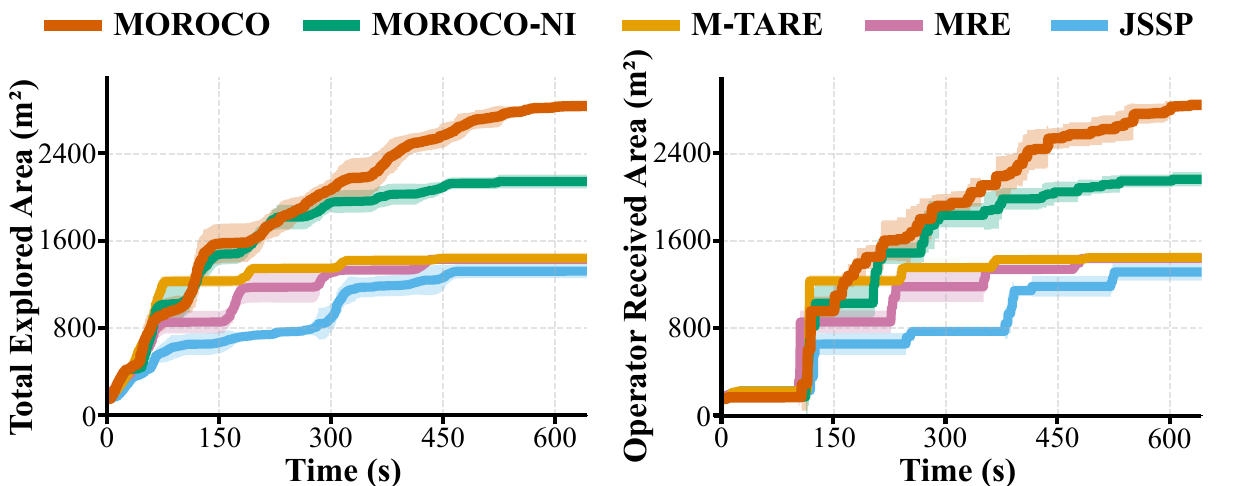}
  \vspace{-2mm}\caption{Evolution of the explored area (\textbf{Left})
    and the operator-received area (\textbf{Right}),
    via the proposed method and baseline methods in the building complex scenarios,
    averaged over 5 runs.
  }\label{fig:single-compare}
\end{figure}

Table~\ref{tab:single_team_compact} and Fig.~\ref{fig:single-compare} reveal a consistent trend across all environments.
CMRE and M-TARE typically expand rapidly at early stages but become effectively range-limited once periodic returns are imposed,
causing map updates to stagnate after roughly $400s$ and resulting in incomplete coverage, e.g., around $50\%$ in the building environment.
JSSP further degrades under the same constraint due to suboptimal rendezvous and waiting behaviors, yielding even lower coverage.
MOROCO-NI substantially improves coverage and reduces the return rate, indicating more efficient information propagation under the latency constraint;
however, it still fails to reliably complete large environments without online operator guidance.
In contrast, {MOROCO} achieves \emph{$100\%$ coverage} in all three environments while maintaining the \emph{lowest return rate} and timely operator updates,
showing that bounded-latency exploration requires not only compliant return scheduling
but also tight coupling between exploration and communication together with online human-in-the-loop intervention.

\begin{table}[t]
  \centering
  \vspace{-2mm}\caption{Single-team results in three scenarios (averaged over $5$ runs).}
  \label{tab:single_team_compact}
  \vspace{-0.06in}
  \small
  \setlength{\tabcolsep}{2.2pt}
  \renewcommand{\arraystretch}{0.92}
  \begin{threeparttable}
    \begin{tabular*}{\columnwidth}{@{\extracolsep{\fill}}ll|cccccc@{}}
      \toprule[1pt]
      \textbf{Environment} & \textbf{Method} &
      \textbf{CA}\tnote{1} & \textbf{ER}\tnote{2} & \textbf{Eff}\tnote{3} &
      \textbf{LU}\tnote{4} & \textbf{RR}\tnote{5} & \textbf{Int}\tnote{6} \\
      \midrule
      \multirow{5}{*}{\makecell{Building\\$60m\times50m$}}
        & CMRE       & 1433 & 50.5 & 2.4 & 432.8 & 3.8 & no \\
        & M-TARE     & 1442 & 50.8 & 2.4 & 445.5 & 3.7 & no \\
        & JSSP       & 1323 & 46.6 & 2.2 & 425.5 & 2.9 & no \\
        & MOROCO-NI  & 2146 & 75.6 & 3.6 & 449.6 & 2.1 & no \\
        & MOROCO     & \textbf{2838} & \textbf{100} & \textbf{4.7} & \textbf{573.5} & \textbf{1.4} & \textbf{yes} \\
      \midrule
      \multirow{5}{*}{\makecell{Cave\\$65m\times57m$}}
        & CMRE       & 1233 & 63.1 & 1.1 & 433.6 & 3.8 & no \\
        & M-TARE     & 1397 & 71.5 & 1.3 & 556.0 & 4.0 & no \\
        & JSSP       & 813  & 28.0 & 0.7 & 500.8 & 2.5 & no \\
        & MOROCO-NI  & 1572 & 80.5 & 1.4 & 875.2 & 1.75 & no \\
        & MOROCO     & \textbf{1953} & \textbf{100} & \textbf{1.8} & \textbf{1027.1} & \textbf{1.1} & \textbf{yes} \\
      \midrule
      \multirow{5}{*}{\makecell{Indoor\\$120m\times90m$}}
        & CMRE       & 562  & 19.3 & 0.2 & 543.2 & 3.6 & no \\
        & M-TARE     & 564  & 19.4 & 0.2 & 484.5 & 3.5 & no \\
        & JSSP       & 580  & 19.9 & 0.2 & 495.6 & 2.9 & no \\
        & MOROCO-NI  & 966  & 33.2 & 0.3 & 473.1 & 2.1 & no \\
        & MOROCO     & \textbf{2908} & \textbf{100} & \textbf{1.0} & \textbf{2818.0} & \textbf{1.4} & \textbf{yes} \\
      \bottomrule[1pt]
    \end{tabular*}

    \begin{tablenotes}[para]
      \footnotesize
      \item[1] CA: explored area in m$^2$.
      \item[2] ER: exploration rate in \%.
      \item[3] Eff: exploration efficiency in m$^2$/s.
      \item[4] LU: last update time in s.
      \item[5] RR: return rate in \#/$T_\texttt{h}$.
      \item[6] Int: whether operator--robot interaction is supported.
    \end{tablenotes}
  \end{threeparttable}
  \vspace{-2mm}
\end{table}

\begin{table}[t]
  \centering
  \vspace{-2mm}\caption{Two-team results in three scenarios (averaged over $5$ runs).}
  \label{tab:multi_team}
  \vspace{-0.06in}
  \small
  \setlength{\tabcolsep}{2.2pt}
  \renewcommand{\arraystretch}{0.92}
  \begin{threeparttable}
    \begin{tabular}{@{}ll|ccccc@{}}
      \toprule[1pt]
      \textbf{Environment} & \textbf{Method} &
      \textbf{EA(m$^2$)}\tnote{1} & \textbf{MT(s)}\tnote{2} & \textbf{ET(s)}\tnote{3} & \textbf{IL(s)}\tnote{4} & \textbf{PO(\%)}\tnote{5} \\
      \midrule
      \multirow{5}{*}{\makecell{Building\\$60m\times50m$}}
        & M-CMRE & 1453.2 & --     & --     & 0.0    & 82.4 \\
        & MO-NM  & 2837.0 & 493.5  & 420.0  & 321.4  & 48.5 \\
        & MO-NP  & 2819.7 & 588.7  & 570.7  & \textbf{202.1} & 76.6 \\
        & MO-NE  & 2833.6 & 582.7  & 546.4  & 359.0  & 63.2 \\
        & MOROCO & \textbf{2838.1} & \textbf{478.3} & \textbf{413.5} & 280.8 & \textbf{47.0} \\
      \midrule
      \multirow{5}{*}{\makecell{Cave\\$65m\times57m$}}
        & M-CMRE & 1426.1 & --     & --     & 0.0    & 83.1 \\
        & MO-NM  & 1930.3 & 1023.0 & 952.8  & 287.2  & 63.0 \\
        & MO-NP  & 1955.7 & 1047.2 & 995.6  & \textbf{250.6} & 79.2 \\
        & MO-NE  & 1936.7 & 952.7  & 856.0  & 482.7  & 71.4 \\
        & MOROCO & \textbf{1958.4} & \textbf{806.1} & \textbf{724.4}  & 305.4 & \textbf{57.3} \\
      \midrule
      \multirow{5}{*}{\makecell{Indoor\\$120m\times90m$}}
        & M-CMRE & 572.5  & --     & --     & 0.0    & 73.0 \\
        & MO-NM  & 1873.7 & --     & --     & 328.9  & 64.7 \\
        & MO-NP  & 2748.6 & 2036.1 & \textbf{1830.8} & 346.7  & 70.7 \\
        & MO-NE  & 2712.0 & 2008.9 & 2005.5 & 1030.3 & 72.6 \\
        & MOROCO & \textbf{2779.0} & \textbf{1935.3} & 1865.2 & \textbf{319.4} & \textbf{56.2} \\
      \bottomrule[1pt]
    \end{tabular}
    \begin{tablenotes}[para]
      \footnotesize
      \item[1] Total explored area.
      \item[2] Mission time.
      \item[3] Exploration termination time.
      \item[4] Maximum inter-team latency.
      \item[5] Percentage of overlapping area.
    \end{tablenotes}
  \end{threeparttable}
  \vspace{-2mm}
\end{table}

\subsubsection{Multi-team Experiments}
Because no existing baseline simultaneously addresses intra-team and inter-team latency constraints, MOROCO is compared against three controlled ablations:
\textbf{MO-NM} without migrate, \textbf{MO-NP} without prioritized-regions, and \textbf{MO-NE} without external events and inter-team latency enforcement.
Besides, a naive extension of the single-team baseline CMRE to multi-teams is also tested for comparison, denoted as \textbf{M-CMRE}.
All methods are evaluated using two teams with two robots and one operator per team under $T_\texttt{h}=160s$ and $T_\texttt{c}=360s$, over five runs.

\begin{figure}[t!]
  \centering
  \includegraphics[width=0.95\linewidth]{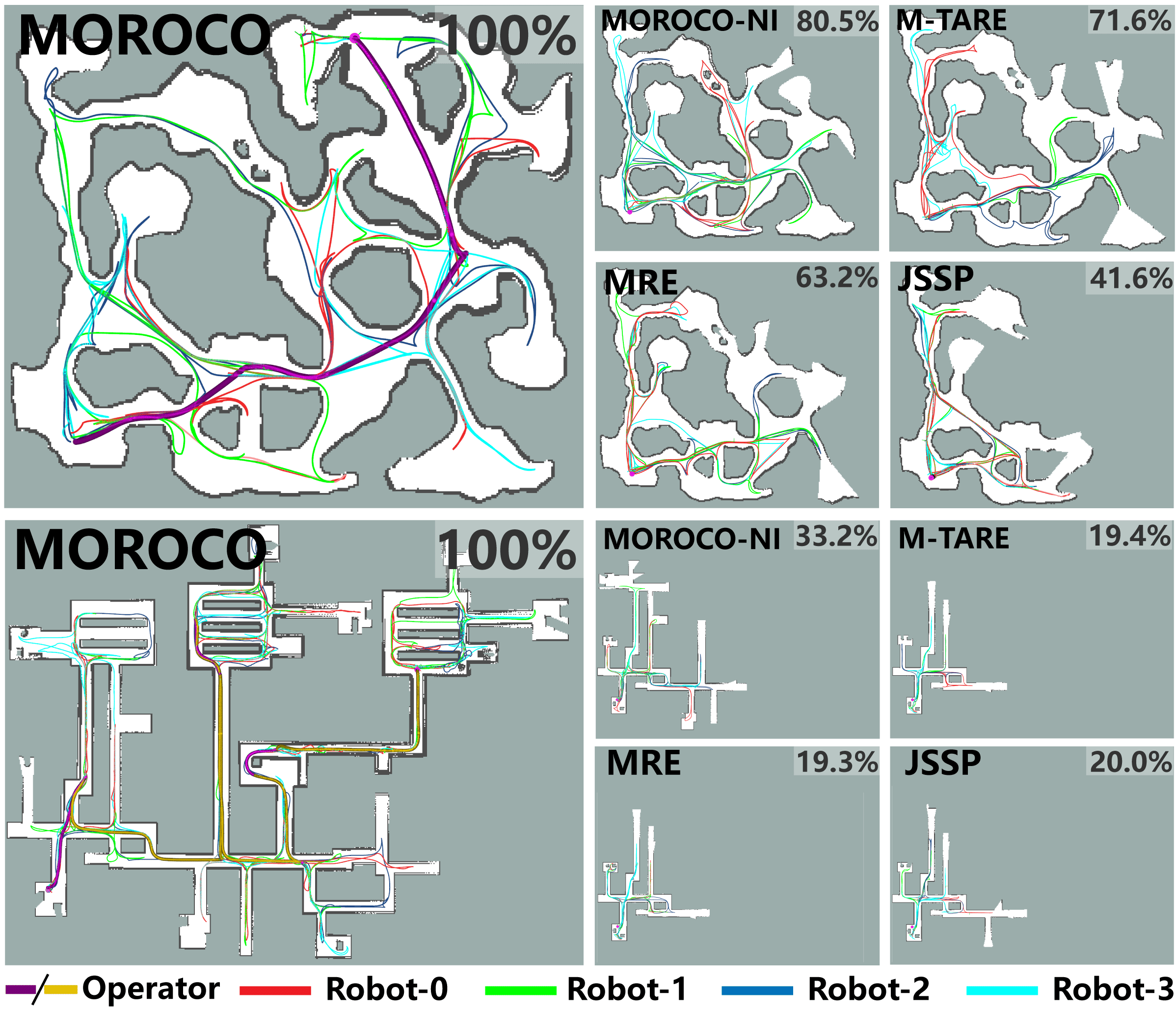}
  \vspace{-2mm}\caption{Final maps obtained by the operator (\textbf{Left}),
    and the trajectories of
    the operator and all robots (\textbf{Right}) in the cave (\textbf{Top})
    and indoor scenarios (\textbf{Bottom}),
    under the proposed method and four baselines.
  }\label{fig:cave-trajectory}
\end{figure}

\begin{figure*}[t!]
  \centering
  \includegraphics[width=0.95\linewidth]{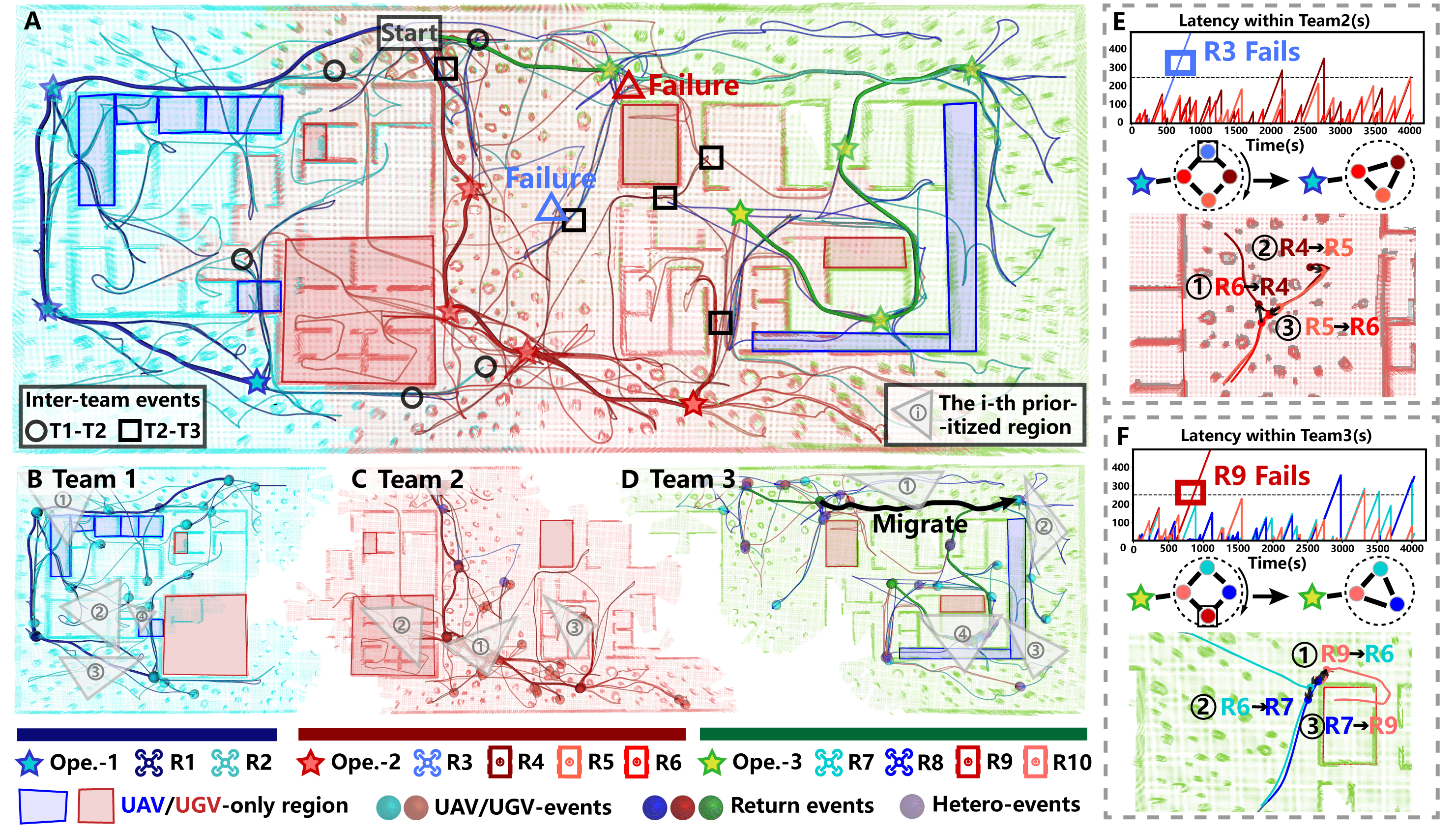}
  \vspace{-2mm}\caption{
    Simulation results of \textbf{three teams} composed of heterogeneous robots collaboratively exploring a large and complex environment
    and meanwhile dealing \emph{with potential robot failures}.
    \textbf{(A)}: the complete final map colored with contributions from each team,
    the trajectories of operators (thicker lines) and robots (thinner lines),
    the positions of external events between neighboring teams,
    and the regions reachable by only one type of robots;
    \textbf{(B)-(D)}: the trajectories of each team and positions of communication events colored by its type,
    as well as the prioritized regions ordered by time.
    \textbf{(E)-(F)}: the failure detection and recovery process for Team-2 and Team-3.
  }\label{fig:large-scale}
\end{figure*}

As summarized in Table~\ref{tab:multi_team}, MOROCO consistently delivers the strongest overall performance,
combining full coverage with shorter mission times, reduced inter-team overlap, and inter-team latency strictly within $T_\texttt{c}$.
The naive M-CMRE presents the lowest coverage due to static operator, and highest overlap as it lacks any mechanism for inter-team coordination.
The ablations isolate the functional necessity of each component.
Without migrate, MO-NM may terminate prematurely in constrained topologies such as the indoor environment, preventing completion.
Without prioritized regions, MO-NP induces substantial redundancy, with overlap exceeding $70\%$ in smaller environments where the teams would otherwise interfere.
Without enforced external exchanges, MO-NE exhibits rapidly increasing inter-team latency as the environment scales, exceeding $T_\texttt{c}=360s$ and degrading operator update timeliness and exploration efficiency.
Collectively, these results indicate that migrate, operator-directed spatial decoupling, and latency-constrained inter-team coordination are jointly required to sustain efficient multi-team exploration under bounded-latency updates.

\begin{figure*}[t!]
  \centering
  \includegraphics[width=0.95\linewidth]{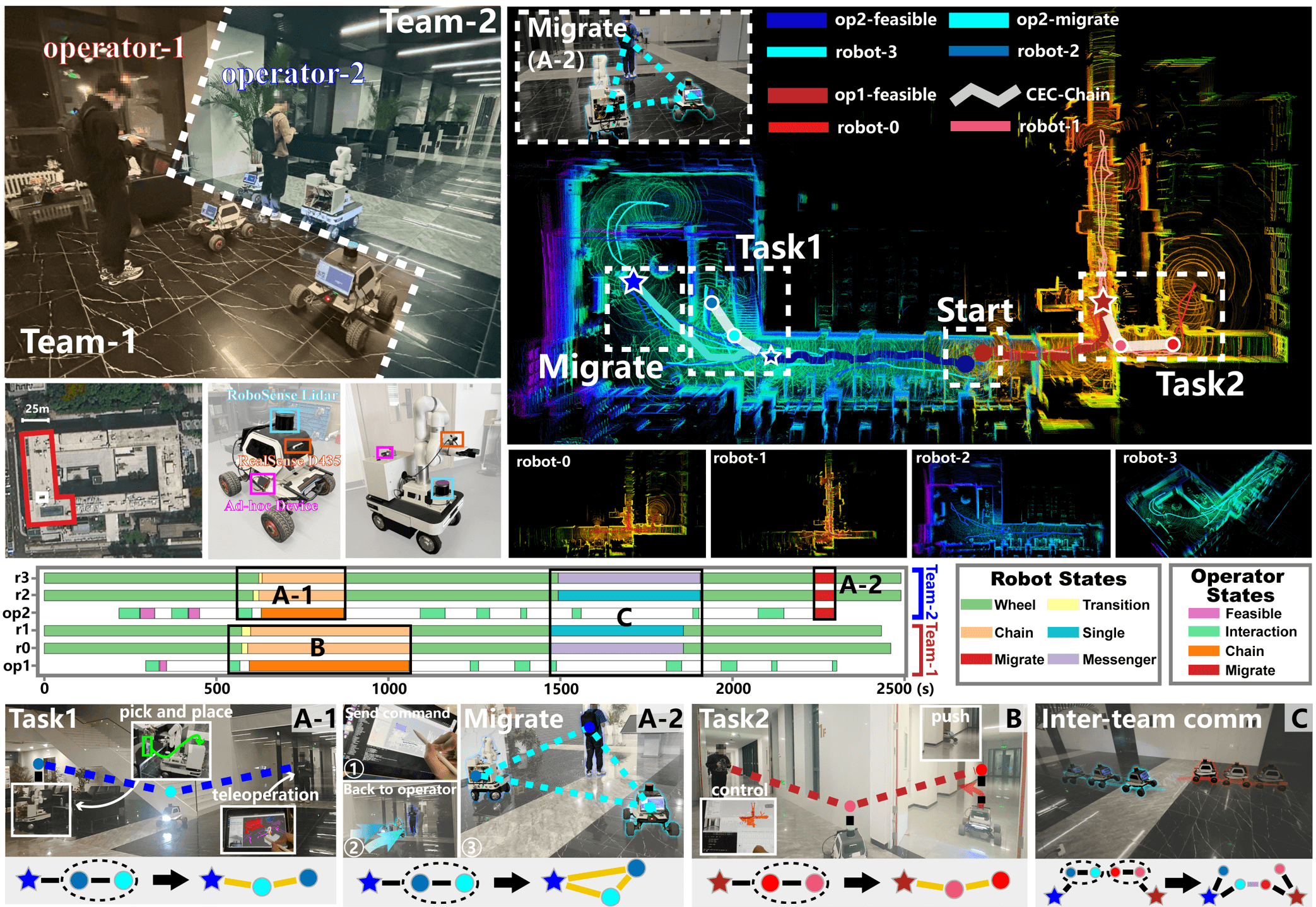}
  \vspace{-2mm}\caption{
\textbf{Top-left}:
  the experimental environment and the heterogeneous robot teams.
  Team-1 consists of 2 ground robots, while team-2 consists of 1 ground robot and 1 mobile manipulator.
  Local communication is enabled via the FT-Dlink ad-hoc device.
  \textbf{Top-right}:
  the final merged map at the end of the mission,
  and the local maps explored by each robot.
  Trajectories of the operators are indicated by the thick lines,
  starting from the initial location
  and ending at the solid stars,
  while the thin lines represent the trajectories of the robots.
  \textbf{Middle}:
  The evolution of communication graphs in both teams during the whole mission.
  \textbf{Bottom}: Snapshots of key moments and graph transitions during the mission.
  }\label{fig:real_exp_fig1}
\end{figure*}

\subsection{Generalizations}\label{subsec:generalizations}
This experiment validates key generalizations of the proposed framework in a single setting, including adaptation to heterogeneous fleets, online recovery from robot failures, and scalability to multiple teams.

\subsubsection{Simulated Scenario}
As shown in Fig.~\ref{fig:large-scale}, three teams with $10$ robots and $3$ operators collaboratively explore a large environment containing both UAV-only and UGV-only accessible regions.
UAVs and UGVs have different mobility limits, with maximum speeds of $1.2\,m/s$ and $0.8\,m/s$, respectively.
Inter-team communication follows a sparse line graph in which team-1 and team-3 communicate through team-2.
The latency bounds are set to $T_\texttt{h}=250s$ and $T_\texttt{c}=500s$, and unexpected robot failures may occur during execution, with waiting-time bounds $T^+_\texttt{max}=60s$ and $T^-_\texttt{max}=30s$.

\subsubsection{Results}
The results in Fig.~\ref{fig:large-scale} confirm that the framework generalizes to heterogeneous multi-team exploration while preserving bounded-latency information propagation under failures.
First, heterogeneous constraints are respected at execution time: planned and realized coordination events occur only at locations simultaneously reachable by the involved robot types, and the final exploration pattern demonstrates capability-aware allocation, where UAV-heavy teams naturally expand into UAV-only regions while UGV resources focus on height-constrained areas, reducing unnecessary detours in mixed-access structures.
Second, the system remains resilient to robot dropouts and continues exploration with limited disruption.
Robot-$2$ in team-2 fails at $356s$ and is detected after the $30s$ waiting bound, after which coordination proceeds with an updated $3$-robot intra-team topology; robot-$8$ in team-3 fails at $606s$, is detected after the $60s$ waiting bound, and is fully recovered after $3$ subsequent meeting events through message propagation and topology update.
Despite these failures, exploration still completes at around $3600s$ and the workload remains balanced across teams, with explored-area ratios of $45\%$, $51\%$, and $56\%$.
Most importantly, latency guarantees are preserved throughout, with maximum inter-team latencies of $485s$ and $472s$ below $T_\texttt{c}=500s$, and maximum intra-team latencies of $244s$, $221s$, and $237s$ below $T_\texttt{h}=250s$ for all non-failed robots and non-messenger robots.
It can be seen that the proposed scheme scales to multiple teams with heterogeneous capabilities and maintains bounded-latency coordination even under online topology reconfiguration induced by failures.

\subsection{Hardware Experiments}\label{subsec:hard-experiments}
To further validate the proposed method in real-world conditions, hardware experiments are conducted with a heterogeneous fleet and multiple operators.
More experimental details can be found in the supplementary video.

\subsubsection{System setup}\label{subsubsec:setup}
The experiments are performed in a standard office environment of size $100m\times 30m$ with rooms and corridors,
with no pre-installed communication infrastructure.
The fleet consists of three Scout-mini ground robots and one Cobot-kit mobile manipulator, deployed as two teams, each with two robots and one operator.
Each Scout-mini is equipped with a 16-line lidar, an IMU, and an onboard Nvidia Xavier computer, running fast-lio2~\cite{xu2022fastlio2} at $10$Hz to build a 3D point-cloud map that is converted to a 2D occupancy grid map via \texttt{octomap}~\cite{hornung13auro} for planning.
The mobile manipulator runs \texttt{hdl\_graph\_slam}~\cite{Kenji2019hdl} for lidar-based SLAM, and all robots carry an Intel RealSense D435 camera for inspection and tele-operation.
Robot navigation uses \texttt{move\_base} with a maximum speed of $0.5m/s$, and communication relies on range-limited FT-Dlink ad-hoc devices.
Each operator carries a laptop for computation and a lightweight Android tablet for visualization and issuing online requests through the GUI.
Teams are initialized at the same corner of the environment with a shared coordinate system; the intra-team latency is set to $T_\texttt{h}=300s$ and the inter-team latency is set to $T_\texttt{c}=1800s$.

\subsubsection{Results}\label{subsubsec:results}
As shown in Fig~\ref{fig:real_exp_fig1}, two teams successfully explore the office environment
and the operators obtain a complete map after about $2500s$, covering $1072m^2$.
Timely operator situational awareness is maintained through return events, with 9 returns per team and maximum update latencies of $287s$ and $263s$ respectively,
both below the prescribed bound $T_\texttt{h}=300s$.
Inter-team coordination further improves efficiency: the inter-team communication event occurs at $1565s$, satisfying $T_\texttt{c}=1800s$,
and limits redundant exploration to $92m^2$, corresponding to $8.6\%$ overlap.
Beyond exploration, two operator-handled tasks further demonstrate the effectiveness of the proposed interaction mechanism under constrained connectivity.
First, the ``push door and inspect'' task is executed by Operator-1 in Team-1.
After Operator-1 issues the request $\xi^1_4$, the graph matching process takes~$3.1s$ to generate a feasible plan,
and it takes $27s$ for the two robots to reach the assigned locations.
Then, a smooth video stream from Robot-0 is transmitted to Operator-1 to support tele-operation for door pushing and inspection.
Second, the ``pick and place'' task is executed by Operator-2 in Team-2,
who teleoperates the Cobot-kit mobile manipulator (Robot-2) to pick and place a bottle on the table.
The graph matching takes $2.8s$ and transition takes $26s$,
which ensures that the mobile manipulator is assigned as the end robot in the chain.
Then, robots 2 and 3 form a connected line graph, enabling Operator-2 to perform the task using a customized tablet GUI that provides
(i) a video stream from the gripper-mounted camera, (ii) the local point cloud around the manipulator,
and (iii) a \texttt{MoveIt} interface for remote tele-operation.
With this support, Operator-2 completes this task in $240s$.
Lastly, Team-2 performs a team-wise migration as a star graph at $2243s$ for further exploration.
Overall, the latency-bounded updates via the intermittent communication have shown to be essential for the operators to make informed decisions,
while the graph decomposition enables online request fulfillment.
\section{Conclusion} \label{sec:conclusion}
This work presented {MoRoCo}, an online topology-adaptive
framework for multi-operator multi-robot coordination under restricted
communication. Built on a latency-bounded intermittent communication
backbone, MoRoCo enables collaborative exploration and online request
servicing through dynamic graph reconfiguration using only local
communication. Simulations and hardware experiments demonstrated its
effectiveness in communication-restricted environments.
Future work will address robustness to uncertain map merging,
richer semantic and multimodal interaction,
and topology-aware exploration strategies under restricted communication.


\bibliographystyle{IEEEtran}
\bibliography{contents/references}

@inproceedings{yamauchi1997frontier,
  title={A frontier-based approach for autonomous exploration},
  author={Yamauchi, Brian},
  booktitle={Proc. IEEE Int. Symp. Comput. Intell. Robot. Autom.},
  pages={146--151},
  year={1997}
}

@ARTICLE{zhou2023racer,
  author={Zhou, Boyu and Xu, Hao and Shen, Shaojie},
  journal={IEEE Trans. Robot.}, 
  title={RACER: Rapid Collaborative Exploration With a Decentralized Multi-UAV System}, 
  year={2023},
  volume={39},
  number={3},
  pages={1816-1835},
  doi={10.1109/TRO.2023.3236945}}

@article{yamauchi1999decentralized,
title = {Decentralized coordination for multirobot exploration},
journal = {Robot. Auton. Syst.},
volume = {29},
number = {2},
pages = {111-118},
year = {1999},
issn = {0921-8890},
doi = {https://doi.org/10.1016/S0921-8890(99)00046-9},
author = {{B. Yamauchi}}
}

@article{burgard2005coordinated,
  title={Coordinated multi-robot exploration},
  author={Burgard, Wolfram and Moors, Mark and Stachniss, Cyrill and Schneider, Frank E},
  journal={IEEE Trans. Robot.},
  volume={21},
  number={3},
  pages={376--386},
  year={2005},
}

@inproceedings{colares2016next,
  title={The next frontier: Combining information gain and distance cost for decentralized multi-robot exploration},
  author={Colares, Rafael Gon{\c{c}}alves and Chaimowicz, Luiz},
  booktitle={Proc. ACM Symp. Appl. Comput.},
  pages={268--274},
  year={2016}
}

@inproceedings{hussein2014multi,
  title={Multi-robot task allocation for search and rescue missions},
  author={Hussein, Ahmed and Adel, Mohamed and Bakr, Mohamed and Shehata, Omar M and Khamis, Alaa},
  booktitle={J. Phys.: Conf. Ser.},
  volume={570},
  number={5},
  pages={052006},
  year={2014},
}

@inproceedings{klaesson2020planning,
  author    = {Filip Klaesson and Petter Nilsson and Tiago Stegun Vaquero and Scott Tepsuporn and Aaron D. Ames and Richard M. Murray},
  title     = {Planning and Optimization for Multi-Robot Planetary Cave Exploration under Intermittent Connectivity Constraints},
  booktitle = {Proc. ICAPS Workshop Planning Robot.},
  year      = {2020}
}

@inproceedings{vaquero2018approach,
  title={An approach for autonomous multi-rover collaboration for mars cave exploration: Preliminary results},
  author={Vaquero, Tiago and Troesch, Martina and Chien, Steve},
  booktitle={Proc. Int. Symp. Artif. Intell., Robot. Autom. Space},
  year={2018}
}

@article{couceiro2017overview,
  title={An overview of swarm robotics for search and rescue applications},
  author={Couceiro, Micael Santos},
  journal={Artif. Intell.: Concepts, Methodol., Tools, Appl.},
  pages={1522--1561},
  year={2017},
  publisher={IGI Global}
}

@inproceedings{marchukov2019fast,
  title={Fast and scalable multi-robot deployment planning under connectivity constraints},
  author={Marchukov, Yaroslav and Montano, Luis},
  booktitle={Proc. IEEE Int. Conf. Auton. Robot Syst. Compet.},
  pages={1--7},
  year={2019},
}

@INPROCEEDINGS{patil2023graph,
  author={Tolstaya, Ekaterina and Paulos, James and Kumar, Vijay and Ribeiro, Alejandro},
  booktitle={Proc. IEEE/RSJ Int. Conf. Intell. Robots Syst.}, 
  title={Multi-Robot Coverage and Exploration using Spatial Graph Neural Networks}, 
  year={2021},
  volume={},
  number={},
  pages={8944-8950},
  doi={10.1109/IROS51168.2021.9636675}
}

@article{saboia2022achord,
  title={ACHORD: Communication-aware multi-robot coordination with intermittent connectivity},
  author={Saboia, Maira and Clark, Lillian and Thangavelu, Vivek and Edlund, Jeffrey A and Otsu, Kyohei and Correa, Gustavo J and Varadharajan, Vivek Shankar and Santamaria-Navarro, Angel and Touma, Thomas and Bouman, Amanda and others},
  journal={IEEE Robot. Autom. Lett.},
  volume={7},
  number={4},
  pages={10184--10191},
  year={2022},
  publisher={IEEE}
}

@inproceedings{cesare2015multi,
  title={Multi-UAV exploration with limited communication and battery},
  author={Cesare, Kyle and Skeele, Ryan and Yoo, Soo-Hyun and Zhang, Yawei and Hollinger, Geoffrey},
  booktitle={Proc. IEEE Int. Conf. Robot. Autom.},
  pages={2230--2235},
  year={2015},
}

@inproceedings{gao2022meeting,
  title={Meeting-merging-mission: A multi-robot coordinate framework for large-scale communication-limited exploration},
  author={Gao, Yuman and Wang, Yingjian and Zhong, Xingguang and Yang, Tiankai and Wang, Mingyang and Xu, Zhixiong and Wang, Yongchao and Lin, Yi and Xu, Chao and Gao, Fei},
  booktitle={Proc. IEEE/RSJ Int. Conf. Intell. Robots Syst.},
  pages={13700--13707},
  year={2022},
}

@article{kantaros2019temporal,
  title={Temporal logic task planning and intermittent connectivity control of mobile robot networks},
  author={Kantaros, Yiannis and Guo, Meng and Zavlanos, Michael M},
  journal={IEEE Trans. Autom. Control},
  volume={64},
  number={10},
  pages={4105--4120},
  year={2019},
  publisher={IEEE}
}

@inproceedings{moravec1985high,
  title={High resolution maps from wide angle sonar},
  author={Moravec, Hans and Elfes, Alberto},
  booktitle={Proc. IEEE Int. Conf. Robot. Autom.},
  volume={2},
  pages={116--121},
  year={1985},
}

@article{thrun2002probabilistic,
  title={Probabilistic robotics},
  author={Thrun, Sebastian},
  journal={Commun. ACM},
  volume={45},
  number={3},
  pages={52--57},
  year={2002},
  publisher={ACM New York, NY, USA}
}

@article{tian2022kimera,
  title={Kimera-multi: Robust, distributed, dense metric-semantic slam for multi-robot systems},
  author={Tian, Yulun and Chang, Yun and Arias, Fernando Herrera and Nieto-Granda, Carlos and How, Jonathan P and Carlone, Luca},
  journal={IEEE Trans. Robot.},
  volume={38},
  number={4},
  year={2022},
  publisher={IEEE}
}

@book{burkard2012assignment,
author = {Burkard, Rainer and Dell'Amico, Mauro and Martello, Silvano},
title = {Assignment Problems},
publisher = {Society for Industrial and Applied Mathematics},
year = {2012},
doi = {10.1137/1.9781611972238},
address = {},
edition   = {},
}

@misc{gor,
author={Google OR-Tools},
howpublished={\texttt{\url{https://github.com/google/or-tools}}},
year={2022}
}

@inproceedings{esposito2006maintaining,
  title={Maintaining wireless connectivity constraints for swarms in the presence of obstacles},
  author={Esposito, Joel M and Dunbar, Thomas W},
  booktitle={Proc. IEEE Int. Conf. Robot. Autom.},
  pages={946--951},
  year={2006},
}

@article{zavlanos2011graph,
  title={Graph-theoretic connectivity control of mobile robot networks},
  author={Zavlanos, Michael M and Egerstedt, Magnus B and Pappas, George J},
  journal={Proc. IEEE},
  volume={99},
  number={9},
  pages={1525--1540},
  year={2011},
  publisher={IEEE}
}

@article{pei2013connectivity,
  title={Connectivity and bandwidth-aware real-time exploration in mobile robot networks},
  author={Pei, Yuanteng and Mutka, Matt W and Xi, Ning},
  journal={Wireless Commun. Mob. Comput.},
  volume={13},
  number={9},
  pages={847--863},
  year={2013},
  publisher={Wiley Online Library}
}

@article{zhang2022mr,
  title={MR-TopoMap: Multi-robot exploration based on topological map in communication restricted environment},
  author={Zhang, Zhaoliang and Yu, Jincheng and Tang, Jiahao and Xu, Yuanfan and Wang, Yu},
  journal={IEEE Robot. Autom. Lett.},
  volume={7},
  number={4},
  pages={10794--10801},
  year={2022},
  publisher={IEEE}
}

@ARTICLE{schack2024sound,
  author={Schack, Matthew A. and Rogers, John G. and Dantam, Neil T.},
  journal={IEEE Robot. Autom. Lett.}, 
  title={The Sound of Silence: Exploiting Information From the Lack of Communication}, 
  year={2024},
  volume={9},
  number={7},
  pages={6736-6743},
  doi={10.1109/LRA.2024.3410158}
}

@article{marcotte2020optimizing,
  title={Optimizing multi-robot communication under bandwidth constraints},
  author={Marcotte, Ryan J and Wang, Xipeng and Mehta, Dhanvin and Olson, Edwin},
  journal={Auton. Robots.},
  volume={44},
  number={1},
  pages={43--55},
  year={2020},
  publisher={Springer}
}

@inproceedings{tian2024ihero,
  title={{iHERO}: Interactive Human-oriented Exploration and Supervision Under Scarce Communication},
  author={ Tian, Zhuoli and Zhang, Yuyang and Wei, Jinsheng and Guo, Meng},
  booktitle={Proc. Robot.: Sci. Syst.},
  year={2024}
}

@article{dahiya2023survey,
  title={A survey of multi-agent Human--Robot Interaction systems},
  author={Dahiya, Abhinav and Aroyo, Alexander M and Dautenhahn, Kerstin and Smith, Stephen L},
  journal={Robot. Auton. Syst.},
  volume={161},
  pages={104335},
  year={2023},
  publisher={Elsevier}
}

@article{villani2020humans,
  title={Humans interacting with multi-robot systems: a natural affect-based approach},
  author={Villani, Valeria and Capelli, Beatrice and Secchi, Cristian and Fantuzzi, Cesare and Sabattini, Lorenzo},
  journal={Auton. Robots.},
  volume={44},
  pages={601--616},
  year={2020},
  publisher={Springer}
}

@article{cao2023tare,
author = {C. Cao  and H. Zhu  and Z. Ren  and H. Choset  and J. Zhang },
title = {Representation granularity enables time-efficient autonomous exploration in large, complex worlds},
journal = {Sci. Robot.},
volume = {8},
number = {80},
pages = {eadf0970},
year = {2023},
}

@INPROCEEDINGS{Ani2024iros,
  author={Da Silva, Alysson Ribeiro and Chaimowicz, Luiz and Silva, Thales C. and Hsieh, M. Ani},
  booktitle={Proc. IEEE/RSJ Int. Conf. Intell. Robots Syst.}, 
  title={Communication-Constrained Multi-Robot Exploration with Intermittent Rendezvous}, 
  year={2024},
  volume={},
  number={},
  pages={3490-3497},
  }

@ARTICLE{liu2024slideslam,
  author={Liu, Xu and Lei, Jiuzhou and Prabhu, Ankit and Tao, Yuezhan and Spasojevic, Igor and Chaudhari, Pratik and Atanasov, Nikolay and Kumar, Vijay},
  journal={IEEE Trans. Robot.}, 
  title={SlideSLAM: Sparse, Lightweight, Decentralized Metric-Semantic SLAM for Multirobot Navigation}, 
  year={2025},
  volume={41},
  number={},
  pages={6529-6548},
  doi={10.1109/TRO.2025.3629786}
}

@ARTICLE{tan20254cnetdiffusionapproachmap,
  author={Tan, Aaron Hao and Narasimhan, Siddarth and Nejat, Goldie},
  journal={IEEE Trans. Robot.}, 
  title={4CNet: A Diffusion Approach to Map Prediction for Decentralized Multi-Robot Exploration}, 
  year={2026},
  volume={},
  number={},
  pages={1-17},
  doi={10.1109/TRO.2026.3666133}
}

@ARTICLE{xu2022fastlio2,
  author={Xu, Wei and Cai, Yixi and He, Dongjiao and Lin, Jiarong and Zhang, Fu},
  journal={IEEE Trans. Robot.}, 
  title={FAST-LIO2: Fast Direct LiDAR-Inertial Odometry}, 
  year={2022},
  volume={38},
  number={4},
  pages={2053-2073},
}

@ARTICLE{hornung13auro,
  author = {Armin Hornung and Kai M. Wurm and Maren Bennewitz and Cyrill
  Stachniss and Wolfram Burgard},
  title = {{OctoMap}: An Efficient Probabilistic {3D} Mapping Framework Based
  on Octrees},
  journal = {Auton. Robots.},
  year = 2013,
  volume  = {34},
  number  = {3},
  pages   = {189--206},
  doi = {10.1007/s10514-012-9321-0},
}

@inproceedings{sygkounas2022multi,
  title={Multi-agent exploration with reinforcement learning},
  author={Sygkounas, Alkis and Tsipianitis, Dimitris and Nikolakopoulos, George and Bechlioulis, Charalampos P},
  booktitle={Proc. IEEE Mediterranean Conf. Control Autom. (MED)},
  pages={630--635},
  year={2022}
}

@inproceedings{calzolari2024reinforcement,
  author    = {Gabriele Calzolari and Vidya Sumathy and Christoforos Kanellakis and George Nikolakopoulos},
  title     = {Reinforcement Learning Driven Multi-Robot Exploration via Explicit Communication and Density-Based Frontier Search},
  booktitle = {Proc. IEEE Int. Conf. Robot. Autom.},
  pages     = {11406--11412},
  year      = {2025},
  doi       = {10.1109/ICRA55743.2025.11128566}
}

@inproceedings{baek2025pipe,
  author={Baek, Seungjae and Moon, Brady and Kim, Seungchan and Cao, Muqing and Ho, Cherie and Scherer, Sebastian and Jeon, Jeong Hwan},
  booktitle={Proc. IEEE/RSJ Int. Conf. Intell. Robots Syst.}, 
  title={PIPE Planner: Pathwise Information Gain with Map Predictions for Indoor Robot Exploration}, 
  year={2025},
  pages={7684-7691},
  doi={10.1109/IROS60139.2025.11246190}
}

@article{al2015enhanced,
  title={Enhanced frontier-based exploration for indoor environment with multiple robots},
  author={Al Khawaldah, Mohammad and N{\"u}chter, Andreas},
  journal={Adv. Robot.},
  volume={29},
  number={10},
  pages={657--669},
  year={2015},
  publisher={Taylor \& Francis}
}

@article{ohradzansky2022multi,
  title={Multi-agent autonomy: Advancements and challenges in subterranean exploration},
  author={Ohradzansky, Michael T and Rush, Eugene R and Riley, Danny G and Mills, Andrew B and Ahmad, Shakeeb and McGuire, Steve and Biggie, Harel and Harlow, Kyle and Miles, Michael J and Frew, Eric W and others},
  journal={Field Robot.},
  volume={2},
  pages={1068--1104},
  year={2022},
  publisher={FRPS}
}

@ARTICLE{dahlquist2025deployment,
  author={Dahlquist, Niklas and Nordström, Samuel and Stathoulopoulos, Nikolaos and Lindqvist, Björn and Saradagi, Akshit and Nikolakopoulos, George},
  journal={IEEE Trans. Field Robot.}, 
  title={Deployment of an Aerial Multiagent System for Automated Task Execution in Large-Scale Underground Mining Environments}, 
  year={2025},
  volume={2},
  number={},
  pages={465-484},
  doi={10.1109/TFR.2025.3580068}
}

@ARTICLE{asgharivaskasi2025riemannian,
  author={Asgharivaskasi, Arash and Girke, Fritz and Atanasov, Nikolay},
  journal={IEEE Trans. Robot.}, 
  title={Riemannian Optimization for Active Mapping With Robot Teams}, 
  year={2025},
  volume={41},
  number={},
  pages={1077-1097},
  doi={10.1109/TRO.2025.3526295}
}

@inproceedings{chen20243d,
  title={A 3d mixed reality interface for human-robot teaming},
  author={Chen, Jiaqi and Sun, Boyang and Pollefeys, Marc and Blum, Hermann},
  booktitle={Proc. IEEE Int. Conf. Robot. Autom.},
  pages={11327--11333},
  year={2024}
}

@inproceedings{yu2021smmr,
  title={Smmr-explore: Submap-based multi-robot exploration system with multi-robot multi-target potential field exploration method},
  author={Yu, Jincheng and Tong, Jianming and Xu, Yuanfan and Xu, Zhilin and Dong, Haolin and Yang, Tianxiang and Wang, Yu},
  booktitle={Proc. IEEE Int. Conf. Robot. Autom.},
  pages={8779--8785},
  year={2021},
}

@ARTICLE{bi2023cure,
  author={Bi, Qingchen and Zhang, Xuebo and Wen, Jian and Pan, Zhangchao and Zhang, Shiyong and Wang, Runhua and Yuan, Jing},
  journal={IEEE Trans. Autom. Sci. Eng.}, 
  title={CURE: A Hierarchical Framework for Multi-Robot Autonomous Exploration Inspired by Centroids of Unknown Regions}, 
  year={2024},
  volume={21},
  number={3},
  pages={3773-3786},
  doi={10.1109/TASE.2023.3285300}
}

@masterthesis{horner2016map,
  author = {Jiří Hörner},
  title = {Map-merging for multi-robot system},
  address = {Prague},
  year = {2016},
  school = {Charles University in Prague, Faculty of Mathematics and Physics},
  type = {Bachelor's thesis}
}

@article{yu2020review,
author = {Yu, Shuien and Fu, Chunyun and K Gostar, Amirali and Hu, Minghui},
year = {2020},
month = {12},
pages = {6988},
title = {A Review on Map-Merging Methods for Typical Map Types in Multiple-Ground-Robot SLAM Solutions},
volume = {20},
journal = {Sensors},
doi = {10.3390/s20236988}
}

@article{Kenji2019hdl,
  author  = {K. Koide and J. Miura and E. Menegatti},
  title   = {A portable three-dimensional {LIDAR}-based system for long-term and wide-area people behavior measurement},
  journal = {Int. J. Adv. Robot. Syst.},
  volume  = {16},
  number  = {2},
  year    = {2019},
  doi     = {10.1177/1729881419841532}
}

@ARTICLE{xia2023relink,
  author={Xia, Lijia and Deng, Beiming and Pan, Jie and Zhang, Xiaoxun and Duan, Peiming and Zhou, Boyu and Cheng, Hui},
  journal={IEEE Robot. Autom. Lett.}, 
  title={RELINK: Real-Time Line-of-Sight-Based Deployment Framework of Multi-Robot for Maintaining a Communication Network}, 
  year={2023},
  volume={8},
  number={12},
  pages={8152-8159},
  }

@article{riley2023fielded,
  title={Fielded Human-Robot Interaction for a Heterogeneous Team in the DARPA Subterranean Challenge},
  author={Danny G. Riley and Eric W. Frew},
  journal={ACM Trans. Human-Robot Interact.},
  year={2023},
  volume={12},
  pages={1 - 24},
}

@article{wang2025multirobot,
      title={Multi-Robot System for Cooperative Exploration in Unknown Environments: A Survey}, 
      author={Chuqi Wang and Chao Yu and Xin Xu and Yuman Gao and Xinyi Yang and Wenhao Tang and Shu'ang Yu and Yinuo Chen and Feng Gao and ZhuoZhu Jian and Xinlei Chen and Fei Gao and Boyu Zhou and Yu Wang},
      year={2025},
      eprint={2503.07278},
      archivePrefix={arXiv},
      primaryClass={cs.RO},
      journal={arXiv preprint arXiv:2503.07278}
}

@ARTICLE{8633953,
  author={Corah, Micah and O'Meadhra, Cormac and Goel, Kshitij and Michael, Nathan},
  journal={IEEE Robot. Autom. Lett.}, 
  title={Communication-Efficient Planning and Mapping for Multi-Robot Exploration in Large Environments}, 
  year={2019},
  volume={4},
  number={2},
  pages={1715-1721},
  doi={10.1109/LRA.2019.2897368}
}

@INPROCEEDINGS{11128211,
  author={Bai, Ruofei and Yuan, Shenghai and Li, Kun and Guo, Hongliang and Yau, Wei-Yun and Xie, Lihua},
  booktitle={Proc. IEEE Int. Conf. Robot. Autom.}, 
  title={Realm: Real-Time Line-of-Sight Maintenance in Multi-Robot Navigation with Unknown Obstacles}, 
  year={2025},
  volume={},
  number={},
  pages={7363-7369},
  doi={10.1109/ICRA55743.2025.11128211}}

\appendix

\subsection{{Proof of Lemmas and Theorems}}\label{app:proof}
\begin{proof}
\textbf{of Lemma~\ref{prop:latency}}.
Denote by $t^-_{\texttt{h}}(n_i)$ and $t^+_{\texttt{h}}(n_i)$ the time
instants immediately before and after the operator $\texttt{h}$
incorporates the received data at the return event
$\mathbf{c}_{h,n_i}$. Then, for each
$\mathbf{c}_{h,n_i}\in \Gamma_{\texttt{h}}$, it holds that
$
  t^{j}_{\texttt{h}}\!\left(t^+_{\texttt{h}}(n_i)\right)=
  \mathbf{max}\Big\{t^{j}_{\texttt{h}}\!\left(t^-_{\texttt{h}}(n_i)\right),\,
  t^{j}(n_i)\Big\},\;
  \forall j\in \mathcal{N},
$
where $t^{j}_{\texttt{h}}(t)$ denotes the most recent update time at
the operator $\texttt{h}$ for data originating from robot $j$, and
$t^{j}(n_i)$ is the time-stamp of robot $j$ carried by $n_i$ at the
return event. Consequently,
$\delta_{\texttt{h}}\!\left(t^+_{\texttt{h}}(n_i)\right)
=t_{\texttt{h}}(n_i)-\mathbf{min}_{j\in \mathcal{N}}\big\{t^{j}(n_i)\big\}
\leq t_{\texttt{h}}(n_i)-\chi_{n_i}$,
where
$\chi_{n_i}\triangleq \mathbf{min}_{j\in \mathcal{N}}\{t^{j}(n_i)\}$.
If conditions (I) and (II) hold, then
$\chi_{n_i}>\chi_{n_{i-1}}$ implies
$
t_{\texttt{h}}(n_i)-\chi_{n_i}
< t_{\texttt{h}}(n_i)-\chi_{n_{i-1}}
\leq T_{\texttt{h}},
$
which yields
$\delta_{\texttt{h}}(t^+_{\texttt{h}}(n_i))<T_{\texttt{h}}$.
Moreover, by the monotonicity of $\delta_{\texttt{h}}(t)$ between
consecutive return events, it holds that
\begin{align*}
  \delta_{\texttt{h}}\!\left(t^-_{\texttt{h}}(n_i)\right)
  &=\delta_{\texttt{h}}\!\left(t^+_{\texttt{h}}(n_{i-1})\right)
    +t_{\texttt{h}}(n_i)-t_{\texttt{h}}(n_{i-1})\\
  &\leq t_{\texttt{h}}(n_i)-\chi_{n_{i-1}}
  \leq T_{\texttt{h}},
\end{align*}
which implies
$\delta_{\texttt{h}}(t^-_{\texttt{h}}(n_i))\leq T_{\texttt{h}}$.
Therefore, the latency constraint holds at each return event.
Furthermore, for any
$t\in (t_{\texttt{h}}(n_{i-1}),\,t_{\texttt{h}}(n_i))$, it follows
that
$\delta_{\texttt{h}}(t)\leq
\delta_{\texttt{h}}(t^-_{\texttt{h}}(n_i))\leq T_{\texttt{h}}$.
This completes the proof.
\end{proof}

\begin{proof}
\textbf{of Lemma~\ref{prop:exist-solution}}.
Consider one execution of $\texttt{PairCoord}(\cdot)$ between robots
$i$ and $j$. The while-loop removes exactly one task $f^\star$ from
the unsigned set $\mathcal{F}^-$ at each iteration, hence
$|\mathcal{F}^-|$ strictly decreases and the loop terminates in at most
$|\mathcal{F}^-|$ iterations, proving finite termination.
It remains to show that the returned event $\mathbf{c}_{ij}$ is
feasible, i.e., satisfies~\eqref{eq:pairing_cond}. After the
return-event stage, either condition~\eqref{eq:return_cond} holds and
no return is appended, in which case the current stamp $\chi^\star$
already provides a valid upper bound for the next pairing; or a return
event is appended to one robot $n\in\{i,j\}$, which updates the stamp
to $\chi^\star$ and yields a conservative feasible upper bound of
$T_\texttt{h}+\chi^\star$ for subsequent coordination.
In the task-sequence stage, $\mathbf{p}_{ij}$ is recomputed from
$\texttt{TSP\text{-}Path}(p^{L_i}_i,\mathcal{F}^-,p^{L_j}_j)$ and
$\mathbf{c}_{ij}$ is selected on $\mathbf{p}_{ij}$ by
$\texttt{SelComm}(\cdot)$. If the loop stops at some iteration, then
\eqref{eq:pairing_cond} is satisfied by construction. Otherwise, tasks
are removed until $\mathcal{F}^-=\emptyset$, in which case
$\mathbf{p}_{ij}$ reduces to a direct path connecting $p^{L_i}_i$ and
$p^{L_j}_j$, and $\texttt{SelComm}(\cdot)$ chooses a meeting event
$\mathbf{c}_{ij}$ on this path that minimizes the resulting
coordination time. Since the return-event stage guarantees a feasible
schedule under the bound $T_\texttt{h}+\chi^\star$ for this worst-case
direct coordination, $\mathbf{c}_{ij}$ must satisfy
\eqref{eq:pairing_cond}. Therefore, $\texttt{PairCoord}(\cdot)$ always
returns a feasible plan and a feasible next pairing event.
\end{proof}

\begin{proof}
\textbf{of Theorem~\ref{prop:Ts-constraint}}.
Lemma~\ref{prop:exist-solution} ensures that each execution of
pairwise coordination returns a feasible solution. In other words,
every communication event satisfies constraint~\eqref{eq:pairing_cond}.
Therefore, by Lemma~\ref{prop:latency}, the base request
$\{\xi^k_1\}$ is fulfilled at all times for each team
$\mathcal{T}_k\in\mathcal{T}$.
\end{proof}

\begin{proof}
\textbf{of Lemma~\ref{thm:graphmatch-exec-short}}.
The candidate family $\widehat{\mathcal{G}}_k^{\xi}$ is finite, and
Alg.~\ref{alg:graphmatch} evaluates one candidate per iteration. The
heuristic in~\eqref{eq:graphmatch_next_candidate} determines only the
order of evaluation, so all candidates are eventually examined and the
request-instantiation procedure terminates in finite time.
For each candidate for which~\eqref{eq:graphmatch_trs} returns
$(T_{\texttt{trs}},\,\widehat{\pi}_k^{\xi})$, the corresponding
detach--execute--rejoin transition is executable by construction.
Then \texttt{LBAP} in~\eqref{eq:role_assignment} returns the
bottleneck-minimizing assignment $\Pi_k^{\xi}$ for that detach
schedule. Hence, each such candidate admits a well-defined exact value
of~\eqref{eq:match_objective}. Since Alg.~\ref{alg:graphmatch}
updates the incumbent using this exact objective value and exhausts the
candidate family, the returned solution is executable under the detach
mechanism and minimizes~\eqref{eq:match_objective} over all candidates
for which \eqref{eq:graphmatch_trs} returns a timed detach schedule.
\end{proof}

\begin{proof}
\textbf{of Theorem~\ref{thm:moroco-complete-short}}.
Since the arrival sequence of request bundles is finite, only finitely
many requests can be introduced during execution.
Consider first a request-activation event. Whenever a request becomes
admissible under the sequential compatibility logic of MoRoCo,
Alg.~\ref{alg:moroco} invokes the request-instantiation procedure
$\texttt{GraphMatch}(\cdot)$ as in~\eqref{eq:combinatorial}. By
assumption, this returns an executable detach schedule for the
corresponding robot set, together with an instantiated request subgraph
and an updated remaining wheel backbone. Hence, after instantiation, the
embedded graph still admits the decomposition in~\eqref{eq:decomposition},
namely a remaining wheel component plus the collection of active request
subgraphs. If a request is not admissible, it is deferred and the
embedded graph remains unchanged. Therefore, after each request-activation
event, the embedded graph remains feasible for subsequent coordination.

Next consider a request-completion event. By assumption, every
instantiated request subgraph completes its assigned request and rejoins
the wheel backbone in finite time. Then, according to
Alg.~\ref{alg:moroco}, the associated subgraph is removed and its robots
are merged back into the wheel component. Thus, after each completion
event, the embedded graph again satisfies~\eqref{eq:decomposition}. This
proves Item~I.
Since only finitely many requests arrive, and each instantiated request
is completed and removed in finite time, every admissible request that
is instantiated by Alg.~\ref{alg:moroco} is serviced in finite time,
which proves Item~II. The special operator-relocation request
$\{\xi_3^k\}$ is deferred while the fleet is split, and becomes
admissible once all active request subgraphs have rejoined and the
complete wheel backbone is restored; it is then handled by the same
argument.
Finally, because only finitely many requests can be instantiated and
each active request subgraph disappears after finite-time completion,
the number of active request subgraphs decreases to zero after finitely
many completion events. Consequently, there exists a finite time $t_f$
such that
$\mathcal{G}_k(t)=\mathcal{G}_k^{\texttt{w}}$ for all $t\ge t_f$.
This proves Item~III.
\end{proof}

\end{document}